\documentclass[10pt,twocolumn,letterpaper]{article}

\usepackage[pagenumbers]{cvpr} 

\usepackage{cuted} 

\usepackage{currfile} 

\usepackage{caption} 
\usepackage{microtype}

\makeatletter
\renewcommand\paragraph{\@startsection{paragraph}{4}{\z@}{.5em \@plus .1em}{-0.8em}{\normalfont\normalsize\bfseries}}
\makeatother

\setlist*[itemize]{itemsep=3pt,topsep=4pt}
\setlist*[enumerate]{itemsep=3pt,topsep=4pt}

\usepackage{multirow}
\usepackage{makecell}
\usepackage{colortbl}
\usepackage{array}
\usepackage{pifont}
\usepackage{bm}
\usepackage[most]{tcolorbox}

\definecolor{secgray}{HTML}{E8E8E8}
\definecolor{oursrow}{HTML}{EEF4FA}
\definecolor{bestgreen}{HTML}{C6F6C6}
\definecolor{worstred}{HTML}{F8C4C4}
\definecolor{dgreen}{HTML}{137333}
\definecolor{dgray}{HTML}{5F6368}
\definecolor{dred}{HTML}{8F1E18}
\definecolor{promptframe}{RGB}{62,124,148}
\definecolor{promptback}{RGB}{239,249,254}

\newcommand{\upgood}[1]{\,{\tiny\textcolor{dgreen}{$\uparrow$#1}}}
\newcommand{\downgood}[1]{\,{\tiny\textcolor{dgreen}{$\downarrow$#1}}}
\newcommand{\upbad}[1]{\,{\tiny\textcolor{dgray}{$\uparrow$#1}}}
\newcommand{\downbad}[1]{\,{\tiny\textcolor{dgray}{$\downarrow$#1}}}

\newcommand{\cmark}{\textcolor{dgreen}{\ding{51}}}
\newcommand{\xmark}{\textcolor{dred}{\ding{55}}}

\tcbset{promptbox/.style={breakable, colback=promptback, colframe=promptframe,
  coltitle=white, fonttitle=\bfseries\large, boxrule=0.5mm}}

\newcommand{\TODO}[1]{\textbf{\color{red}[TODO: #1]}}
\renewcommand{\TODO}[1]{}

\definecolor{cvprblue}{rgb}{0.21,0.49,0.74}
\usepackage[breaklinks,colorlinks,allcolors=cvprblue]{hyperref}

\def\confName{CVPR}
\def\confYear{2027}

\makeatletter
\patchcmd{\@maketitle}{\vskip .375in}{\vskip .10in}{}{}
\patchcmd{\@maketitle}{\vspace*{24pt}}{\vspace*{10pt}}{}{}
\patchcmd{\@maketitle}{\vspace*{12pt}}{\vspace*{0pt}}{}{}
\makeatother
\title{EcoVid: Efficient Code to Video with a Token-Saving and\\Geometry-Aware Agent Harness}

\author{%
Zexuan Yan$^{1}$ \quad
Zhikai Wang$^{2}$ \quad
Zhenkun Gao$^{3}$ \quad
Yizhuo Jia$^{4}$ \quad
Juncheng Zhou$^{5}$ \quad
Xinyu Wang$^{6}$ \\[1pt]
Yuzhou Wu$^{1}$ \quad
Jinghong Liu$^{7}$ \quad
Yuanhuiyi Lyu$^{8}$ \quad
Yue Ma$^{8}$ \quad
Linfeng Zhang$^{1}$ \\[4pt]
\small
$^{1}$Shanghai Jiao Tong University \quad
$^{2}$Southern University of Science and Technology \quad
$^{3}$East China Normal University \\[-1pt]
\small
$^{4}$Fudan University \quad
$^{5}$Nankai University \quad
$^{6}$Tsinghua University \quad
$^{7}$Renmin University of China \\[-1pt]
\small
$^{8}$The Hong Kong University of Science and Technology \\[3pt]
\footnotesize
\textbf{Code:} \href{https://github.com/yuriYanZeXuan/EcoVid}{\texttt{https://github.com/yuriYanZeXuan/EcoVid}}
}

\begin{document}
\maketitle
\begin{strip}
\centering
\includegraphics[width=0.92\linewidth]{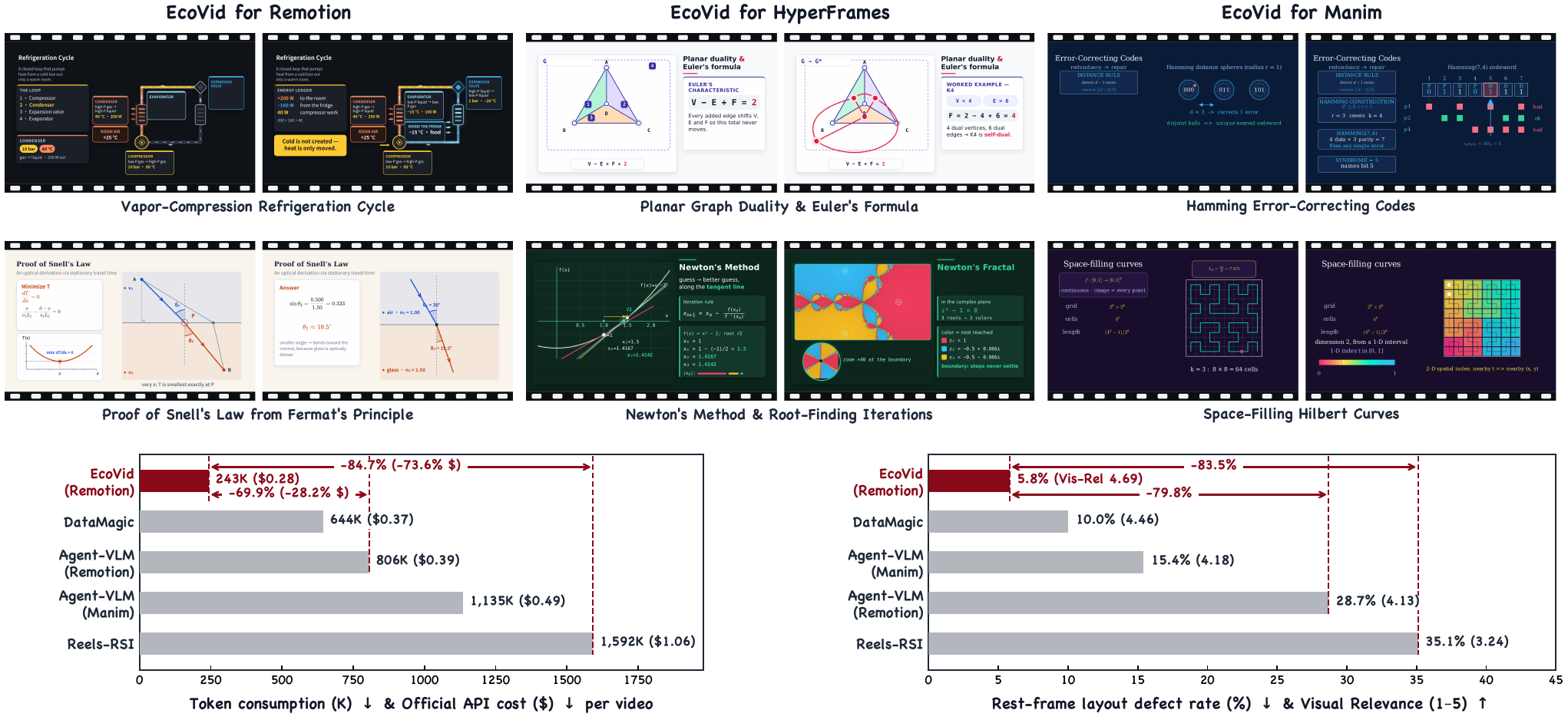}
\captionsetup{hypcap=false}
\captionof{figure}{\textbf{EcoVid: Let the LLM animate, let the harness keep the books.} \textbf{Top:} Representative keyframes across six topics on Remotion, HyperFrames, and Manim showcase diverse publication-grade visual aesthetics with zero layout collisions. \textbf{Bottom:} Offloading timing alignment, geometry verification, and context compaction to our harness substantially reduces token usage, API cost, and rest-frame layout defects while achieving the highest visual relevance across all three rendering carriers.}
\captionsetup{hypcap=true}
\label{fig:\currfilebase}%
\end{strip}

\begin{abstract}
Generating narrated explainer videos by writing executable animation code yields sharp typography, crisp vector diagrams, and exact mathematical visuals, yet existing declarative pipelines and native VLM agents remain slow, costly, and prone to spatial layout collisions and audio-visual drift.
Analyzing multi-turn agent trajectories reveals that over two-thirds of model turns and token expenditure are consumed by mechanical bookkeeping---manual frame-offset arithmetic, pixel-level screenshot inspection, and uncached history replay---rather than high-level visual design.
We propose \textbf{EcoVid} (\textbf{E}fficient \textbf{co}de-to-\textbf{Vid}eo), a carrier-agnostic, token-saving, and geometry-aware agent harness that offloads this mechanical bookkeeping to three deterministic runtime mechanisms.
First, \emph{Symbolic TTS Timing Binding} schedules visual transitions against symbolic sentence anchors in code and binds them to measured speech intervals at runtime, eliminating cumulative timing drift across variable speaking rates.
Second, \emph{Geometry-Aware Feedback} inspects the structured scene layout to deterministically flag bounding-box overlaps, shape occlusions, canvas overflows, and low contrast at zero vision-token cost, routing spatial residuals into a fast text-only repair loop before rendering video pixels.
Third, \emph{Cache-Aware Context Management} compacts multi-turn repair history only at discrete epoch boundaries, keeping the leading prompt prefix byte-identical across consecutive turns to maximize prefix KV-cache reuse while bounding context growth.
Evaluated across a 60-topic bilingual benchmark spanning Remotion, HyperFrames, and Manim, \textbf{EcoVid} cuts token usage by $\mathbf{84\%}$ and API cost by $\mathbf{57\%}$, reduces layout defects to $\mathbf{2.8\%}$, and keeps animations tightly synchronized with spoken narration.
\end{abstract}

\begin{figure}[t]
  \centering
  \includegraphics[width=0.84\linewidth]{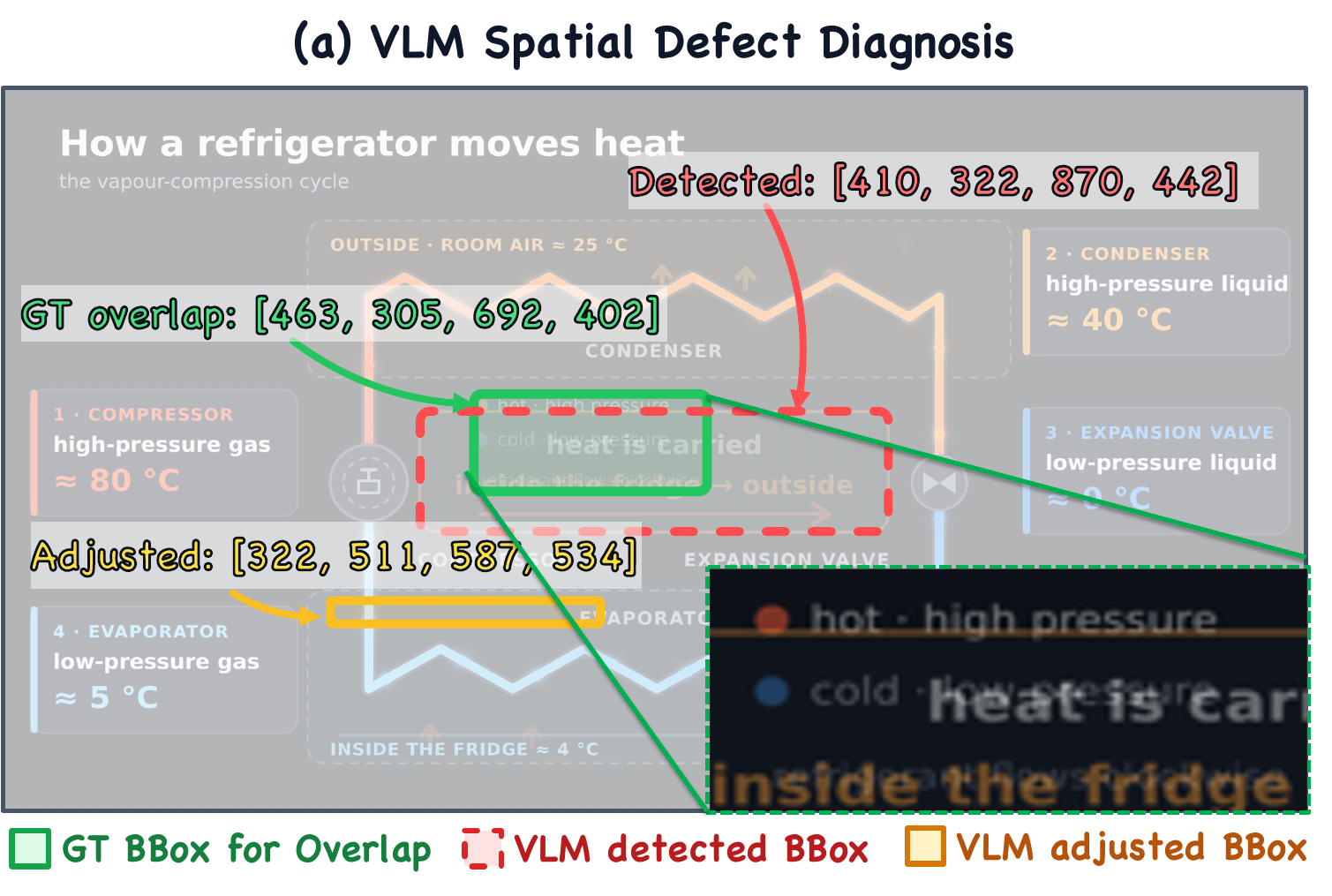}\\[4pt]
  \includegraphics[width=\linewidth]{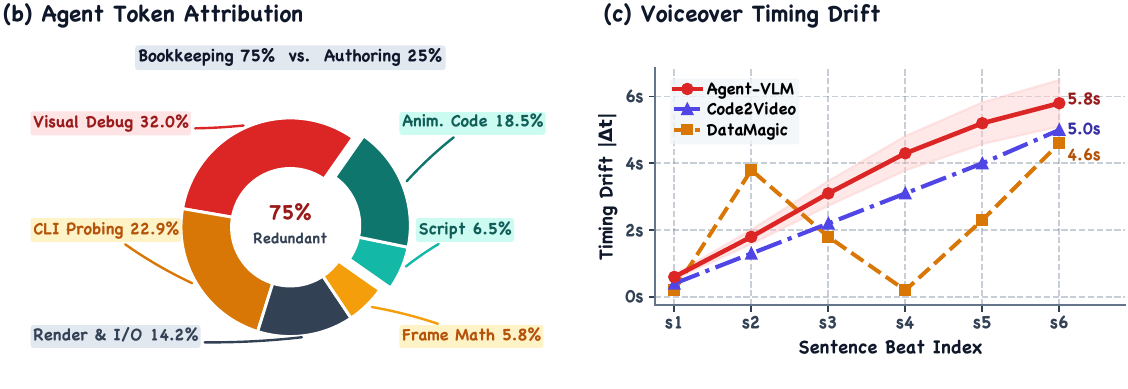}
  \caption{\textbf{Core failure modes of existing code-to-video paradigms.} \textbf{(a)} Pixel-level VLM critics misestimate bounding-box coordinates on dense canvases and fail to resolve layout overlaps. \textbf{(b)} Native VLM agents expend $75\%$ of their token budget on mechanical bookkeeping. \textbf{(c)} Existing code-to-video paradigms suffer voiceover timing drift across spoken sentences.}
  \label{fig:defect_drift}
\end{figure}

\section{Introduction}
\label{sec:intro}

Existing video diffusion models~\cite{kong2024hunyuanvideo,wan2025wan} have achieved remarkable success in video generation, particularly in rendering everyday scenes and simple motion clips.
However, they still struggle with educational explainer videos due to poor controllability, weak structural consistency, and limited text rendering.
To address these issues, synthesizing videos via code-to-video offers clear video quality~\cite{chen2026code2video,ku2025theoremexplainagent,wang2026teachmaster,yan2026beyond}.
Yet turning a topic prompt into a narrated animation exposes difficulties in visual quality, audio-video synchronization, and generation cost~\cite{chen2026code2video,heygen2026code2video,liu2025cachesurvey,yan2025eedit}.
Open-ended agentic loops~\cite{yan2026glyphbanana} consume hundreds of thousands to millions of tokens per video while still leaving overlapping labels, occluded formulas, and desynchronized voiceovers.

We attribute the high cost and fragility of iterative code-to-video generation to two distinct factors. Specifically, when an LLM agent generates an explainer video in animation pipelines~\cite{remotion,hyperframes,manim}, it must handle two distinct tasks.
The first is high-level visual planning, which designs explanatory diagrams and step-by-step animations.
The second is low-level mechanical bookkeeping, which converts speech timing into frame indices, checks for collisions, and extracts context from the repair history.
Tracing native VLM agent trajectories reveals that $75.4\%$ of model calls and $75.0\%$ of tokens are consumed by mechanical bookkeeping, such as feeding screenshots into a vision model to hunt for layout bugs, probing frames, and parsing speech timestamps.
In contrast, only $15.3\%$ of calls and $18.5\%$ of tokens directly write or refine animation code, as shown in \cref{fig:defect_drift}b.
Furthermore, as shown in \cref{fig:defect_drift}a, both declarative scene-level pipelines~\cite{chen2026code2video} and native VLM agents~\cite{lin2026reelsrsi} struggle with three core bookkeeping bottlenecks:
First, pixel-level VLM critique mislocalizes spatial collisions and misses transient motion. Vision models cannot locate precise bounding-box coordinates on dense canvases, so their proposed shifts fail to resolve overlaps. Meanwhile, sparse screenshot inspection misses most static and transient mid-animation collisions.
Second, coarse scene binding and manual frame arithmetic for voiceover timing binding suffer severe voiceover drift. The former is represented by DataMagic~\cite{xie2026datamagic}, which relies on audio-only alignment at multi-sentence scene boundaries and splits internal beats evenly; as shown in \cref{fig:defect_drift}c, this causes intra-scene drift across variable-length sentences. The latter is represented by Agent-VLM, which manually converts speech timestamps into frame indices. Either it accumulates timing errors in the dialogue, or it ends early, which creates a noticeable audio-video sync issue such as trailing silence.
Finally, uncompacted logs and naive pruning break prefix KV-cache economics, as retaining full repair histories inflates context quadratically across turns. Long context windows not only drive up costs, but also hurt retrieval.

To achieve both high visual quality and low generation cost, the agent architecture must offload mechanical bookkeeping to an execution-grounded harness without restricting the LLM's expressive freedom.
Inspired by the above insights, we introduce \textbf{EcoVid} (\textbf{E}fficient \textbf{co}de-to-\textbf{Vid}eo), a token-saving, geometry-aware agent harness (\cref{fig:teaser}) built around three novel mechanisms.
First, \emph{Symbolic TTS Timing Binding} lets the LLM mark each sentence in the narration script with a symbolic anchor, so animations are scheduled by anchor symbols in code. At runtime, the harness binds these anchors to measured word-boundary timestamps, eliminating cumulative drift.
Second, \emph{Geometry-Aware Feedback} replaces VLM-grounded screenshot critique by querying the structured scene layout across the timeline to deterministically detect text overlaps, occlusions, overflows, and low contrast without vision-token consumption, routing compact text diagnostics back for batched code repairs before rendering.
Third, \emph{Cache-Aware Context Management} compacts working memory at discrete multi-turn epoch boundaries while preserving recent turns verbatim. Consequently, the compacted prefix stays byte-identical within each epoch, which ensures high cache-hit rates and reduces costs.

Our approach can be applied to various code-to-video frameworks and has been tested across a wide range of topic domains and languages.
Overall, our main contributions are as follows:
\begin{itemize}
  \item \textbf{The EcoVid agent harness:} We propose \textbf{EcoVid}, a novel code-to-video generation framework that offloads mechanical bookkeeping from the LLM through three core mechanisms: \emph{Symbolic TTS Timing Binding}, \emph{Geometry-Aware Feedback}, and \emph{Cache-Aware Context Management}.
  \item \textbf{Cross-framework efficiency and quality advantage:} \textbf{EcoVid} generalizes seamlessly across various frameworks, cutting token consumption by up to $\mathbf{84\%}$ and API cost by up to $\mathbf{57\%}$ per video while reducing layout defects to $\mathbf{2.8\%}$ to achieve state-of-the-art generation quality.
  \item \textbf{Comprehensive benchmark and ablation analysis:} We construct \textbf{\textsc{Bench60}}, a 60-topic bilingual benchmark, and conduct extensive comparative and ablation studies that systematically validate the effectiveness of our methods.
\end{itemize}

\section{Related Work}
\label{sec:related}

\begin{table}[t]
  \centering
  \caption{\textbf{Comparison of code-to-video systems and agent harnesses across rendering carriers and core capabilities.}}
  \label{tab:related_comp}
  \scriptsize
  \setlength{\tabcolsep}{3.2pt}
  \resizebox{\columnwidth}{!}{%
  \begin{tabular}{l|c|ccc}
    \toprule
    \textbf{\textit{System / Harness}} & \textbf{\textit{Carrier}} & \textbf{\textit{Audio Sync}} & \textbf{\textit{Geometry Check}} & \textbf{\textit{Context Compaction}} \\
    \midrule
    TheoremExplainAgent~\cite{ku2025theoremexplainagent} & Manim & \cmark & \xmark & \xmark \\
    Code2Video~\cite{chen2026code2video} & Manim & \xmark & \xmark & \xmark \\
    SGA~\cite{lopez2026sga} / PRISM~\cite{zhang2026prism} & Manim & \xmark & \cmark & \xmark \\
    DataMagic~\cite{xie2026datamagic} & Remotion & \cmark & \xmark & \xmark \\
    Reels-RSI~\cite{lin2026reelsrsi} & HyperFrames & \cmark & \cmark & \xmark \\
    anything2explainer~\cite{wei2026anything2explainer} & Remotion & \cmark & \xmark & \xmark \\
    video-shotcraft~\cite{wei2026shotcraft} & Remotion & \cmark & \xmark & \xmark \\
    Agent-VLM & Remotion / HF / Manim & \cmark & \xmark & \xmark \\
    \midrule
    \rowcolor{oursrow}\textbf{EcoVid (Ours)} & \textbf{Remotion / HF / Manim} & \cmark & \cmark & \cmark \\
    \bottomrule
  \end{tabular}%
  }
\end{table}

\paragraph{Educational and code-driven video generation.}
Generating instructional videos by synthesizing executable graphics programs has emerged as a reliable alternative to pixel-based diffusion models~\cite{kong2024hunyuanvideo,wan2025wan}, offering sharp typography, exact mathematical notation, and deterministic editability.
TheoremExplainAgent~\cite{ku2025theoremexplainagent} and Code2Video~\cite{chen2026code2video} orchestrate multi-stage planner, coder, and vision-critic pipelines to produce Manim~\cite{manim} animations. Building on this, subsequent works explore pedagogical structuring and lecture generation~\cite{lin2026findings,ma2026from,wang2026teachmaster,yan2026beyond,zhu2025paper2video}, tool-verified execution traces~\cite{fang2026code,li2026videococo,liao2026algogen}, sparse Manim scene-graph checks~\cite{li2026see,lopez2026sga,oli2026manibench,zhang2026prism}, and agentic reinforcement learning~\cite{silva2026training}.
In parallel, web and vector animation stacks such as Remotion~\cite{remotion} and HyperFrames~\cite{hyperframes,heygen2026code2video} power declarative data-video compilers such as DataMagic~\cite{xie2026datamagic} alongside SVG and Lottie motion generators~\cite{chen2026lottiegpt,ji2026animation2code,liu2024logomotion,ma2025mover,shirakawa2025mggen,xiao2026figact,yang2026omnilottie}.
However, existing declarative pipelines remain restricted to rigid single-carrier templates, coarse scene-level audio alignment, and stateless calls that lack dense geometric verification.

\paragraph{Code-driven agent harnesses.}
Beyond fixed pipelines, recent works wrap coding agents in iterative execution-and-critique harnesses across 3D graphics, embodied simulation, and visual design.
In 3D modeling, SceneCraft~\cite{hu2024scenecraft} translates spatial relation graphs into numerical constraints and Blender scripts refined via vision foundation models, BlenderAlchemy~\cite{huang2024blenderalchemy} pairs an edit generator with a visual state evaluator to search material and geometry edits, and 3D-GPT~\cite{sun20233dgpt} dispatches procedural modeling across conceptualization and coding agents~\cite{tang2026bvb}.
For web-based 3D physics, VisPhyWorld~\cite{liang2026visphyworld} and WorldCoder-Bench~\cite{lu2026worldcoder} prompt agents to author executable Three.js and WebGL physical simulations.
In embodied AI, Holodeck~\cite{yang2024holodeck} optimizes 3D asset layouts over LLM-generated spatial constraints and RoboGen~\cite{wang2024robogen} automates task and scene synthesis in a propose-generate-learn loop, while SceneSmith~\cite{pfaff2026scenesmith} coordinates designer, critic, and orchestrator agents with physics checks across layout, furniture, and clutter stages; similarly, Design2Code~\cite{si2025design2code} and GlyphBanana~\cite{yan2026glyphbanana} compile DOM or vector code to refine 2D layouts~\cite{yang2025ui2coden,sun2026januscoder}.
While these harnesses effectively couple code execution with spatial feedback in their domains, they do not address spoken audio-visual timing alignment or KV-cache-preserving context compaction across multi-turn repair trajectories, which \textbf{EcoVid} introduces for narrated video generation.

\begin{figure*}[t]
  \centering
  \includegraphics[width=0.98\textwidth]{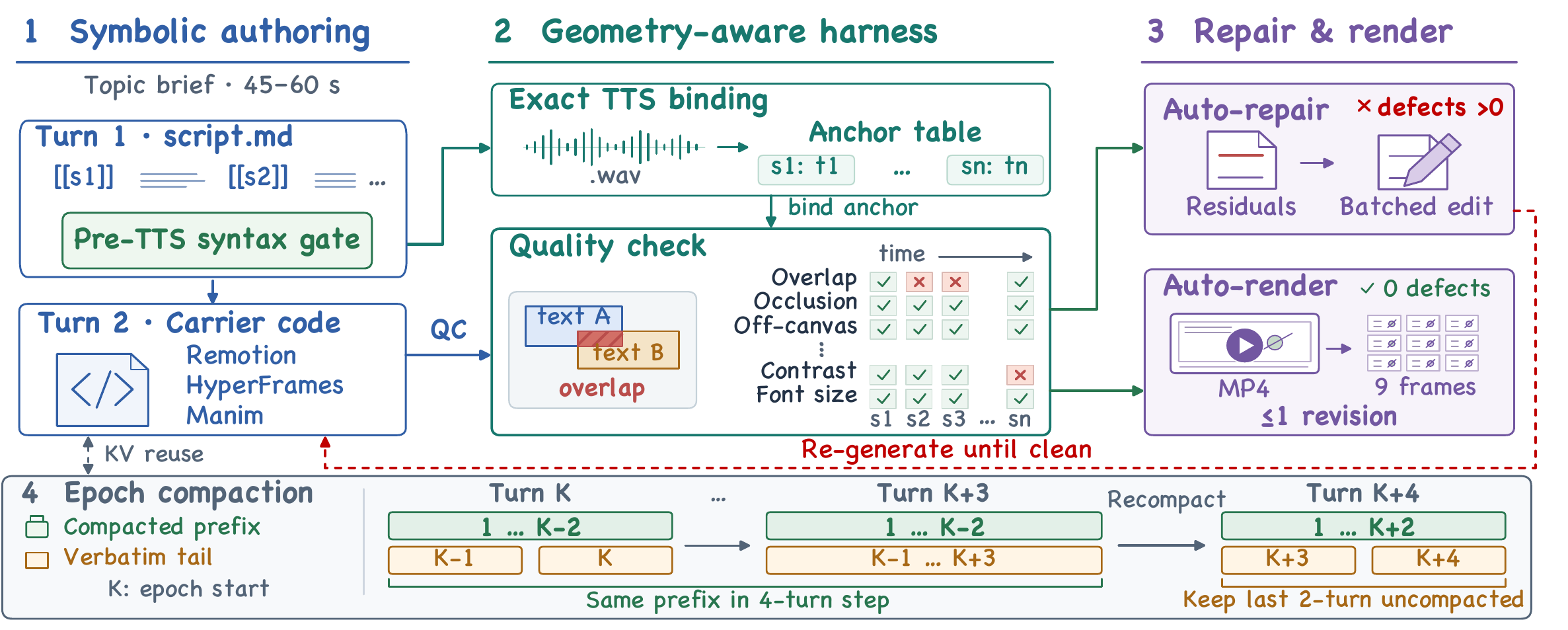}
  \caption{\textbf{Overview of the EcoVid agent harness.} Our framework decouples high-level visual storytelling from mechanical bookkeeping via \textbf{(1)} symbolic TTS timing binding that maps sentence anchors to measured speech intervals at runtime, \textbf{(2)} geometry-aware feedback that checks element bounding-box overlaps and canvas overflows directly on the structured scene layout, and \textbf{(3)} cache-aware context management that prunes repair history at discrete epoch boundaries to preserve prefix KV-cache reuse.}
  \label{fig:pipeline}
\end{figure*}

\section{Preliminaries}
\label{sec:problem}

\paragraph{Task and Scene-Layout Notation.}
Given a user topic prompt, a narrated code-to-video system generates two coupled artifacts: a narration script $S = \{s_1, \ldots, s_M\}$ of $M$ spoken sentences with synthesized text-to-speech (TTS) audio intervals $\{[t^{\text{start}}_m, t^{\text{end}}_m]\}_{m=1}^{M}$ spanning total duration $T = t^{\text{end}}_M$, and an executable animation program $P$.
Rather than synthesizing raw video pixels directly, $P$ specifies visual elements and their temporal animations in a declarative or programmatic graphics engine---such as React component trees in Remotion, timeline-driven HTML/SVG DOMs in HyperFrames, or mathematical scene objects in Manim.
At any timestamp $t \in [0, T]$, executing $P$ produces a structured scene layout $G_t = \{v_1(t), \ldots, v_N(t)\}$, where each visible element $v_i(t)$ is characterized by its axis-aligned bounding box $B_i(t) \subset \Omega$ on the video canvas $\Omega = [0, W] \times [0, H]$, its $z$-index stacking order, and its foreground and background styling attributes.

\paragraph{General Iterative Code-to-Video Pipeline.}
Across existing code-to-video frameworks, turning a topic prompt into a rendered explainer video follows a three-stage iterative workflow over agent turns $k = 1, \ldots, K$:
In the \emph{scripting and speech synthesis stage}, the system generates the narration script $S$ and invokes a TTS engine to obtain the spoken audio waveform and measured sentence intervals $\{[t^{\text{start}}_m, t^{\text{end}}_m]\}_{m=1}^{M}$.
Next, in the \emph{code generation stage}, the LLM coding agent writes or updates the animation program $P_k$ to visually illustrate each spoken sentence $s_m$ across the timeline $[0, T]$.
Finally, in the \emph{verification and repair stage}, the harness evaluates the visual output of $P_k$, produces diagnostic feedback $f_k$, and appends the turn record $h_k = (P_k, f_k)$ to the interaction history $H_k = (h_1, \ldots, h_k)$. This loop repeats until verification passes, when the final MP4 video is rendered.

During multi-turn inference, modern LLM providers accelerate prefill and reduce token billing via prefix key-value (KV) caching.
Specifically, at turn $k$, the provider compares the serialized prompt token sequence $H_k$ against the preceding turn's prompt $H_{k-1}$ starting from the very first token, and reuses cached KV states along their \emph{longest common prefix (LCP)}, denoted $L_k = \mathrm{LCP}(H_k, H_{k-1})$.
Intuitively, $\mathrm{LCP}(H_k, H_{k-1})$ is the maximal contiguous leading token sequence that remains byte-identical between $H_{k-1}$ and $H_k$, up to the first modified, deleted, or inserted token. Because causal attention depends on all preceding tokens, any token after the first discrepancy suffers a cache miss.
Since cached prefix tokens are billed at a fraction of the uncached rate ($c_{\text{cache}} \ll c_{\text{uncached}}$), the input token cost at turn $k$ is:
\begin{equation}
  \mathrm{Cost}(H_k \mid H_{k-1}) = c_{\text{cache}}\,|L_k| + c_{\text{uncached}}\,\bigl(|H_k| - |L_k|\bigr),
  \label{eq:kv_cost}
\end{equation}
where $|L_k|$ is the matched prefix length and $|H_k| - |L_k|$ is the uncached suffix length.
Preserving a long, invariant prefix $|L_k|$ while bounding total context length $|H_k|$ is thus central to the token economics of multi-turn response.

\section{Method}
\label{sec:method}

We propose \textbf{EcoVid}, a carrier-agnostic agent harness that decouples high-level visual design from low-level mechanical bookkeeping across four coordinated stages as shown in \cref{fig:pipeline}.
Our harness isolates creative authoring from deterministic runtime verification, allowing the LLM to focus on high-level visual design while offloading mechanical bookkeeping to the harness.
In Part 1, where the workflow begins, the coding agent translates a topic brief into a narration script $S$ with symbolic sentence anchors in Turn 1. After passing a pre-TTS syntax gate, it authors the executable carrier program $P$ in Turn 2.
In Part 2, the harness assigns the TTS binding plan to the anchor table.
The corresponding anchor table then integrates at the matching symbolic location within the animation code to serve as the voiceover schedule for the actual animation.
During this stage, we run geometry-aware quality feedback to check for flaws and issues among the various elements within and between frames across the animation scenes.
It will help the agent iterate and fix the issues based on that feedback.
In Part 3, the harness detects a defect score.
If any defects are detected, it enters the auto repair stage.
It will repeatedly regenerate until it is clean, and then start modifying again from the carrier code in Part 1.
Otherwise, it moves on to auto render, which will generate the video and, during this stage, produce a 9-frame overview of the entire video.
Only up to one revision will be performed during this phase.
In Part 4, the harness periodically compacts the prefix via epoch compaction to maximize KV-cache reuse across repair turns.

\subsection{Symbolic TTS Timing Binding}
\label{sec:binding}

To eliminate both open-loop duration guessing and manual frame arithmetic, Part 1 and Part 2 decouple symbolic event scheduling in code from deterministic audio timestamps.
Instead of specifying $P$ by raw seconds or frame indices, the LLM writes the narration script $S = \{s_1, \ldots, s_M\}$ in Turn 1 with a symbolic anchor attached to each sentence $s_m$, and schedules every animation entrance, emphasis, and exit in $P$ in Turn 2 using only the sentence symbol $s_m$ and a normalized local progress $\phi \in [0, 1]$.

Before speech synthesis, the pre-TTS syntax gate in Part~1 enforces every TTS-anchor binds one sentence on $S$, rejecting drafts with missing, duplicate, or out-of-order anchors before any audio synthesis or code generation occurs.
Once the voiceover is synthesized in Part 2, the harness extracts the measured audio interval $[t^{\text{start}}_m, t^{\text{end}}_m]$ for each sentence $s_m$ to populate the runtime anchor table $A$ and maps symbolic progress $(s_m, \phi)$ linearly to actual time $t$:
\begin{equation}
  t(s_m, \phi) = t^{\text{start}}_m + \phi \cdot \bigl(t^{\text{end}}_m - t^{\text{start}}_m\bigr), \qquad \phi \in [0, 1].
  \label{eq:temporal_morphism}
\end{equation}
At runtime, the rendering engine resolves the symbolic schedule in $P$ against the anchor table $A$ through \cref{eq:temporal_morphism}, so each transition fires at exactly the acoustic start $t(s_m, 0) = t^{\text{start}}_m$ of its sentence and finishes at exactly $t(s_m, 1) = t^{\text{end}}_m$.
Within each sentence, the agent can freely schedule arbitrary sub-animations over any sub-interval $[\phi_a, \phi_b] \subseteq [0, 1]$ or attach optional word-level sub-anchors to specific spoken terms, so intra-sentence motions scale naturally with the spoken pace.
Consequently, timing drift across all $M$ sentences is eliminated, regardless of speaking rate, language, or frame rate. Moreover, when a sentence in $S$ is edited, its updated duration propagates to all downstream timestamps in $A$ without editing the code.

\subsection{Geometry-Aware Feedback}
\label{sec:gaf}

Instead of debugging layouts from rendered pixels, which localizes spatial errors imprecisely and consumes vision tokens, Part 2 and Part 3 verify layout geometry directly on the structured scene layout $G_t = \{v_1(t), \ldots, v_N(t)\}$ prior to rendering video frames.
Specifically, the harness probes $G_t$ across the sentence timeline $s_1, \ldots, s_M$ at uniform time steps over $[0, T]$ and extracts the exact bounding box $B_i(t)$, stacking order, and styling attributes of each visible element $v_i(t)$ on the canvas $\Omega$.
Without requiring inference from downsampled screenshots, our harness reads these exact layout coordinates directly from the browser layout engine within seconds, capturing both static rest-frame overlaps and motional collisions across the full video timeline.
As illustrated in \cref{fig:pipeline}, the verifier probes every sentence anchor $s_m$ at each sampled timestamp $t$ with a multi-dimensional quality checker.
Specifically, for any two co-visible elements $v_i(t)$ and $v_j(t)$, it first computes their pairwise overlap ratio:
\begin{equation}
  \rho_{\text{ov}}(i, j, t) = \frac{|B_i(t) \cap B_j(t)|}{\min\bigl(|B_i(t)|, |B_j(t)|\bigr)}.
  \label{eq:geometry_feedback}
\end{equation}
Next, for each visible element $v_i(t)$, it measures the out-of-canvas overflow ratio beyond the boundary $\Omega$:
\begin{equation}
  \rho_{\text{out}}(i, t) = \frac{|B_i(t) \setminus \Omega|}{|B_i(t)|}.
  \label{eq:geometry_overflow}
\end{equation}
The verifier then flags a defect whenever $\rho_{\text{ov}}(i, j, t) > \tau_{\text{ov}}$ or $\rho_{\text{out}}(i, t) > 0$. It also performs foreground-to-background contrast and minimum font-size legibility checks.

Contiguous violations are merged into a single defect and mapped back to their enclosing sentence anchor $s_m$ via \cref{eq:temporal_morphism}. This yields a compact residual listing the colliding elements, offending anchors, and correction offsets without vision-token cost.
As shown in \cref{fig:pipeline}, this residual decides whether the agent repairs or renders next.
If defects remain, the agent revises $P$ in a single batch and rechecks the scene layout without rasterizing pixels.
When no defect is left, the harness renders the video and pulls a nine-frame contact sheet to guide at most one holistic revision, leaving vision tokens for overall aesthetics rather than coordinate fixes.

\subsection{Cache-Aware Context Management}
\label{sec:cache}

During multi-turn interactive generation over history $H_k = (h_1, \ldots, h_k)$, inference overhead depends on whether the prompt prefix hits or misses the KV cache, which is determined by the longest common prefix $L_k = \mathrm{LCP}(H_k, H_{k-1})$ in \cref{eq:kv_cost}.
To bound context length while keeping the prompt prefix invariant across consecutive turns, EcoVid updates the pruning boundary $b_k$ only at discrete $s$-turn epoch boundaries while preserving at least $w$ recent turns:
\begin{equation}
  b_k = \Big\lfloor \frac{\max(0, k - w)}{s} \Big\rfloor \cdot s.
  \label{eq:epoch_boundary}
\end{equation}
Given $b_k$, the harness constructs the compacted history $\widetilde{H}_k$ by pruning the prefix up to turn $b_k$ and appending the unchanged recent tail:
\begin{equation}
  \widetilde{H}_k = \mathrm{Prune}(H_{1:b_k}) \,\|\, H_{b_k+1:k},
  \label{eq:epoch_compaction}
\end{equation}
where $\mathrm{Prune}(H_{1:b_k})$ replaces superseded code drafts and resolved diagnostics in the prefix with compact one-line summaries while keeping the latest working code, and $\|$ denotes concatenation (\cref{fig:pipeline} traces turns $k$ to $k+s$).
Across the $s$ consecutive turns from $k$ to $k+s-1$, $b_k$ does not move, so the compacted prefix covering turns $1$ to $b_k$ stays byte-identical and its KV-cache is reused in full, while each new repair turn is appended unchanged to the tail.
At turn $k+s$ the trajectory reaches the next epoch boundary, and the harness recompacts: the turns from $b_k + 1$ to $b_{k+s}$ are folded into the locked prefix, and the unchanged tail is reset to turns $b_{k+s} + 1$ to $k+s$.
Prefix KV-cache hit rates therefore stay high across multi-turn repair while the total context length remains bounded ($s = 4, w = 2$ by default).

\section{Experiments}
\label{sec:exp}

\begin{table*}[t]
  \centering
  \caption{\textbf{Main quantitative comparison across Remotion, HyperFrames, and Manim.} All methods uniformly use Qwen3.8-Max-Medium, and green subscripts denote improvements relative to Agent-VLM.}
  \label{tab:bench60}
  \scriptsize
  \renewcommand{\arraystretch}{1.08}
  \setlength{\tabcolsep}{2.8pt}
  \begin{tabular}{ll|cccc|ccccc}
    \toprule
    \multirow{2}{*}{\textbf{\textit{Carrier}}} & \multirow{2}{*}{\textbf{\textit{Method}}} & \multicolumn{4}{c|}{\textbf{\textit{Efficiency \& Cost Evaluation}}} & \multicolumn{5}{c}{\textbf{\textit{Quality, Synchronization \& Pedagogical Evaluation}}} \\
    \cline{3-11}
    & & \textbf{\textit{Tokens$\downarrow$}} & \textbf{\textit{Cost ($\$$)$\downarrow$}} & \textbf{\textit{Calls$\downarrow$}} & \textbf{\textit{Time (m)$\downarrow$}} & \textbf{\textit{Def.$\downarrow$}} & \textbf{\textit{VS$\uparrow$}} & \textbf{\textit{Vis-Rel$\uparrow$}} & \textbf{\textit{M-KC$\uparrow$}} & \textbf{\textit{AES$\uparrow$}} \\
    \midrule
    \multirow{5}{*}{Remotion}
    & anything2explainer~\cite{wei2026anything2explainer} & 10,258K & \$3.37 & 114.7 & 55.2 & 21.3\% & \textbf{0.821} & 3.86 & 0.776 & 91.5 \\
    & video-shotcraft~\cite{wei2026shotcraft} & 12,674K & \$3.58 & 121.5 & 87.5 & 45.9\% & 0.748 & 3.40 & 0.611 & 80.7 \\
    & DataMagic~\cite{xie2026datamagic} & 644K & \$0.37 & 22.9 & 13.1 & 10.0\% & 0.626 & 4.46 & 0.793 & 95.8 \\
    & Agent-VLM & 806K & \$0.39 & 29.7 & 14.4 & 28.7\% & 0.735 & 4.13 & 0.668 & 88.4 \\
     & \cellcolor{oursrow}\textbf{EcoVid (Ours)} & \cellcolor{oursrow}\textbf{243K}\downgood{563K} & \cellcolor{oursrow}\textbf{\$0.28}\downgood{0.11} & \cellcolor{oursrow}\textbf{13.3}\downgood{16.4} & \cellcolor{oursrow}\textbf{11.6}\downgood{2.8} & \cellcolor{oursrow}\textbf{5.8\%}\downgood{22.9} & \cellcolor{oursrow}0.811\upgood{0.077} & \cellcolor{oursrow}\textbf{4.69}\upgood{0.56} & \cellcolor{oursrow}\textbf{0.820}\upgood{0.152} & \cellcolor{oursrow}\textbf{96.8}\upgood{8.4} \\
    \midrule
    \multirow{3}{*}{HyperFrames}
    & Reels-RSI~\cite{lin2026reelsrsi} & 1,592K & \$1.06 & 57.6 & 26.6 & 35.1\% & 0.689 & 3.24 & 0.696 & 85.5 \\
    & Agent-VLM & 1,146K & \$0.54 & 30.4 & 16.5 & 18.2\% & 0.746 & 4.20 & 0.723 & 91.8 \\
     & \cellcolor{oursrow}\textbf{EcoVid (Ours)} & \cellcolor{oursrow}\textbf{210K}\downgood{935K} & \cellcolor{oursrow}\textbf{\$0.25}\downgood{0.29} & \cellcolor{oursrow}\textbf{11.4}\downgood{19.1} & \cellcolor{oursrow}\textbf{10.8}\downgood{5.7} & \cellcolor{oursrow}\textbf{6.8\%}\downgood{11.0} & \cellcolor{oursrow}\textbf{0.769}\upgood{0.023} & \cellcolor{oursrow}\textbf{4.60}\upgood{0.40} & \cellcolor{oursrow}\textbf{0.811}\upgood{0.088} & \cellcolor{oursrow}\textbf{96.1}\upgood{4.3} \\
    \midrule
    \multirow{4}{*}{Manim}
    & TheoremExplainAgent~\cite{ku2025theoremexplainagent} & \textbf{126K} & \$0.41 & 14.5 & 15.1 & 9.1\% & 0.573 & 4.08 & 0.714 & 94.3 \\
    & Code2Video~\cite{chen2026code2video} & 133K & \$0.39 & 23.2 & 16.0 & 13.3\% & 0.575 & 3.40 & 0.701 & 93.3 \\
    & Agent-VLM & 1,135K & \$0.49 & 36.4 & 19.9 & 15.4\% & 0.746 & 4.18 & 0.704 & 92.6 \\
     & \cellcolor{oursrow}\textbf{EcoVid (Ours)} & \cellcolor{oursrow}183K\downgood{952K} & \cellcolor{oursrow}\textbf{\$0.21}\downgood{0.28} & \cellcolor{oursrow}\textbf{12.8}\downgood{23.5} & \cellcolor{oursrow}\textbf{10.3}\downgood{9.6} & \cellcolor{oursrow}\textbf{2.8\%}\downgood{12.6} & \cellcolor{oursrow}\textbf{0.840}\upgood{0.094} & \cellcolor{oursrow}\textbf{4.64}\upgood{0.47} & \cellcolor{oursrow}\textbf{0.829}\upgood{0.124} & \cellcolor{oursrow}\textbf{98.6}\upgood{6.0} \\
    \bottomrule
  \end{tabular}
\end{table*}

\begin{figure*}[t]
  \centering
  \includegraphics[width=0.99\linewidth]{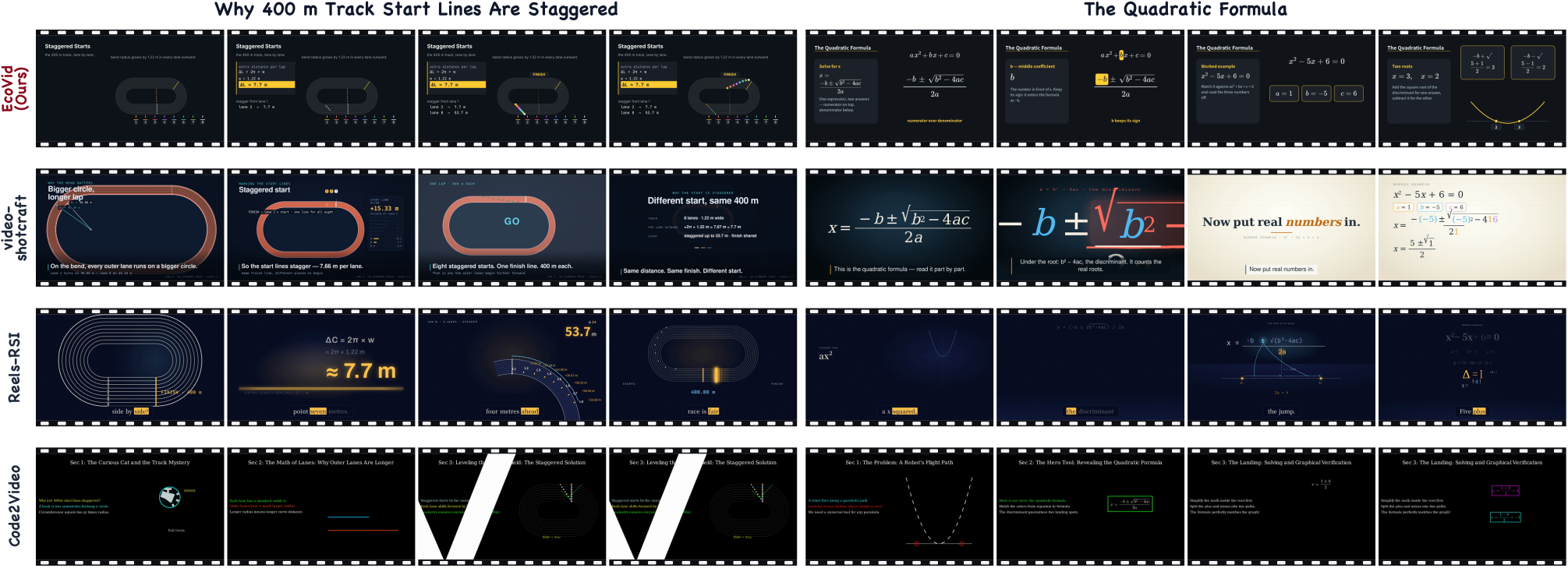}
  \caption{\textbf{Qualitative comparison across four pedagogical stages on \emph{Why 400\,m Track Start Lines Are Staggered} and \emph{The Quadratic Formula}.}
  \textbf{EcoVid} produces clean geometry and step-by-step synchronized derivations, whereas baseline methods exhibit colliding track labels, clipped equations, and unreasonable layout.}
  \label{fig:filmstrip_main}
\end{figure*}

\noindent\textbf{Benchmarks and Evaluation Protocol.}
To evaluate narrated code-to-video generation across diverse domains and languages, we propose \textbf{Bench60}, a new bilingual benchmark of 60 instructional topics spanning science, everyday life, astronomy, geography, health, and sports rules evaluated across Remotion, HyperFrames, and Manim.
Using Qwen3.8-Max-Medium uniformly across all methods, we compare \textbf{EcoVid} against three baseline families: native VLM agents (Agent-VLM) that rely on manual timing arithmetic and screenshot inspection across all three engines; declarative compilation pipelines including DataMagic~\cite{xie2026datamagic}, TheoremExplainAgent~\cite{ku2025theoremexplainagent}, and Code2Video~\cite{chen2026code2video}; and iterative agentic workflows including Reels-RSI~\cite{lin2026reelsrsi}, anything2explainer~\cite{wei2026anything2explainer}, and video-shotcraft~\cite{wei2026shotcraft}.
We measure generation efficiency via \textbf{Tokens}, \textbf{API Cost}, \textbf{Calls}, and \textbf{Time}, alongside five video quality metrics:
\textbf{Static Frame Defect Rate (Def.)}, measuring the share of stationary rest frames with text overlap, shape occlusion, or canvas overflow via an independent 15\,Hz layout probe;
\textbf{Visual Synchronization (VS)}, assessing audio and visual alignment by testing whether each video segment without subtitles visually matches its spoken sentence rather than lagging behind or leading into adjacent sentences;
\textbf{Visual Relevance (Vis-Rel)}, rating how clearly keyframes visually reinforce the spoken explanation~\cite{ku2025theoremexplainagent};
\textbf{Multimodal Knowledge Coverage (M-KC)}, jointly measuring comprehension question accuracy, layout legibility, and visual responsiveness; and
\textbf{Aesthetic Score (AES)}, evaluating overall visual aesthetics, diagrammatic richness, and progressive animation transitions~\cite{chen2026code2video,zhang2026prism}. All experiments were conducted under fair comparison using the average results of three random seeds. Complete benchmark specifications, formal metric definitions, and additional evaluation results are provided in the supplementary material.

\subsection{Main Results}

\begin{table*}[t]
  \centering
  \caption{\textbf{Component-wise ablation study of EcoVid across our three core harness mechanisms in \textsc{Bench60} on Remotion.}}
  \label{tab:abl_main}
  \scriptsize
  \renewcommand{\arraystretch}{1.08}
  \setlength{\tabcolsep}{2.6pt}
  \resizebox{\textwidth}{!}{%
  \begin{tabular}{ll|ccccc|ccccc}
    \toprule
    \multirow{2}{*}{\textbf{\textit{Harness Component}}} & \multirow{2}{*}{\textbf{\textit{Ablated Variant / Baseline}}} & \multicolumn{5}{c|}{\textbf{\textit{Efficiency, KV-Cache \& API Cost}}} & \multicolumn{5}{c}{\textbf{\textit{End-to-End Video Quality}}} \\
    \cline{3-12}
    & & \textbf{\textit{Calls$\downarrow$}} & \textbf{\textit{Tok.$\downarrow$}} & \textbf{\textit{Uncach.$\downarrow$}} & \textbf{\textit{Hit\%$\uparrow$}} & \textbf{\textit{Cost ($\$$)$\downarrow$}} & \textbf{\textit{Def.$\downarrow$}} & \textbf{\textit{VS$\uparrow$}} & \textbf{\textit{Vis-Rel$\uparrow$}} & \textbf{\textit{M-KC$\uparrow$}} & \textbf{\textit{AES$\uparrow$}} \\
    \midrule
    \multirow{2}{*}{1. TTS Timing Binding}
    & w/o Symbolic Timing Binding & 16.5 & 271K & 69K & 72.2\% & \textbf{\$0.28} & 9.1\% & 0.667 & 4.28 & 0.737 & 89.8 \\
     & \cellcolor{oursrow}\textbf{w/ Symbolic Timing Binding (Ours)} & \cellcolor{oursrow}\textbf{13.3} & \cellcolor{oursrow}\textbf{243K} & \cellcolor{oursrow}\textbf{56K} & \cellcolor{oursrow}\textbf{74.1\%} & \cellcolor{oursrow}\textbf{\$0.28} & \cellcolor{oursrow}\textbf{5.8\%} & \cellcolor{oursrow}\textbf{0.811} & \cellcolor{oursrow}\textbf{4.69} & \cellcolor{oursrow}\textbf{0.820} & \cellcolor{oursrow}\textbf{96.8} \\
    \midrule
    \multirow{2}{*}{2. Geometry-Aware Feedback}
    & w/o Geometry-Aware Feedback & 29.7 & 806K & 73K & \textbf{90.7\%} & \$0.39 & 28.7\% & 0.735 & 4.13 & 0.668 & 88.4 \\
     & \cellcolor{oursrow}\textbf{w/ Geometry-Aware Feedback (Ours)} & \cellcolor{oursrow}\textbf{13.3} & \cellcolor{oursrow}\textbf{243K} & \cellcolor{oursrow}\textbf{56K} & \cellcolor{oursrow}74.1\% & \cellcolor{oursrow}\textbf{\$0.28} & \cellcolor{oursrow}\textbf{5.8\%} & \cellcolor{oursrow}\textbf{0.811} & \cellcolor{oursrow}\textbf{4.69} & \cellcolor{oursrow}\textbf{0.820} & \cellcolor{oursrow}\textbf{96.8} \\
    \midrule
    \multirow{3}{*}{3. Context Management}
    & w/o Compaction & 15.6 & 342K & 62K & \textbf{80.3\%} & \$0.31 & 6.8\% & 0.796 & 4.61 & 0.811 & 95.9 \\
    & Compact every 1 step & 16.4 & \textbf{228K} & 86K & 57.5\% & \$0.32 & 6.9\% & 0.786 & 4.58 & 0.808 & 95.5 \\
     & \cellcolor{oursrow}\textbf{Compact every 4 steps (Ours)} & \cellcolor{oursrow}\textbf{13.3} & \cellcolor{oursrow}243K & \cellcolor{oursrow}\textbf{56K} & \cellcolor{oursrow}74.1\% & \cellcolor{oursrow}\textbf{\$0.28} & \cellcolor{oursrow}\textbf{5.8\%} & \cellcolor{oursrow}\textbf{0.811} & \cellcolor{oursrow}\textbf{4.69} & \cellcolor{oursrow}\textbf{0.820} & \cellcolor{oursrow}\textbf{96.8} \\
    \bottomrule
  \end{tabular}%
  }
\end{table*}

\Cref{tab:bench60} and \cref{fig:filmstrip_main} present our primary quantitative and qualitative evaluations on Bench60.
Across all three rendering engines, \textbf{EcoVid} consistently advances the Pareto frontier across visual quality, synchronization, and efficiency.

\paragraph{Reducing Layout Defects via Geometry-Aware Feedback.}
To verify whether structured scene layout verification improves visual quality without restricting visual expressiveness, we compare static-frame defect rates, visual relevance, and aesthetic scores in \Cref{tab:bench60}.
Native VLM agents and multi-stage workflows leave up to $45.9\%$ of rest frames with text overlaps, occlusions, or overflows because VLMs fail to recognize accurate coordinate offsets of visual defects based on concatenated image frames.
By contrast, \textbf{EcoVid} eliminates layout defects to $\mathbf{5.8\%}$ on Remotion, $\mathbf{6.8\%}$ on HyperFrames, and $\mathbf{2.8\%}$ on Manim, while achieving the highest visual relevance (up to $\mathbf{4.69}$) and aesthetic score (up to $\mathbf{98.6}$).
As shown in \cref{fig:filmstrip_main}, whereas baselines exhibit colliding track labels, clipped equations, and unreasonable layout, \textbf{EcoVid} produces clean animation frames and concise narrative structures.

\paragraph{Improving Audio-Visual Synchronization.}
To test whether symbolic timing binding eliminates both the intra-scene drift and the cumulative arithmetic errors, we examine visual synchronization and multimodal knowledge coverage in \Cref{tab:bench60}.
Declarative pipelines~\cite{xie2026datamagic,ku2025theoremexplainagent,chen2026code2video} bind audio only at coarse scene boundaries, while native VLM agents plateau at $0.735$ for visual synchronization due to cumulative frame-math drift and frozen tail frames.
Binding symbolic sentence anchors to measured speech intervals at runtime lifts visual synchronization to $0.811$ and multimodal knowledge coverage to $\mathbf{0.820}$.

\paragraph{Achieving Token and Cost Efficiency.}
Comparing token usage, API cost, calls, and latency in \Cref{tab:bench60}, open-source skills consume over $10\text{M}$ tokens pricing at $3.37\text{\$}$ per video and native VLM agents or self-improvement loops~\cite{lin2026reelsrsi} expend over 1M tokens across more than 50 turns; even stateless multi-agent pipelines~\cite{ku2025theoremexplainagent,chen2026code2video} cost nearly twice as much as \textbf{EcoVid} ($0.49\text{\$}$ vs.\ $0.21\text{\$}$) on Manim because stateless per-scene calls substantially miss KV-cache reuse and rewrite full files (\cref{eq:kv_cost}).
In contrast, \textbf{EcoVid} converges in less than 14 calls and 250K tokens, cutting token usage by up to $\mathbf{84\%}$ and API cost by up to $\mathbf{57\%}$.

\subsection{Ablation Studies}
We conducted ablation studies across various components to verify their effectiveness, as well as the generalization performance of our harness.

\begin{table}[t]
  \centering
  \caption{\textbf{Sensitivity analysis of geometry-aware feedback sampling frequency.}}
  \label{tab:abl_probing}
  \scriptsize
  \setlength{\tabcolsep}{2.2pt}
  \resizebox{\columnwidth}{!}{%
  \begin{tabular}{l|cccc|ccc}
    \toprule
    \textbf{\textit{Feedback Frequency}} & \textbf{\textit{Frames}} & \textbf{\textit{Check$\downarrow$}} & \textbf{\textit{Recall$\uparrow$}} & \textbf{\textit{Trans.$\uparrow$}} & \textbf{\textit{Def.$\downarrow$}} & \textbf{\textit{VS$\uparrow$}} & \textbf{\textit{M-KC$\uparrow$}} \\
    \midrule
    VLM Stills & 6.1 & 14.5s & 34.2\% & 11.5\% & 28.7\% & 0.735 & 0.668 \\
    0.2\,Hz (Beat-end) & 11 & \textbf{0.5s} & 46.0\% & 24.0\% & 9.1\% & 0.667 & 0.737 \\
    1\,Hz Geometry & 60 & 0.7s & 61.1\% & 23.0\% & 7.4\% & 0.743 & 0.774 \\
    2\,Hz Geometry & 119 & 0.9s & 74.8\% & 40.0\% & 6.8\% & 0.775 & 0.796 \\
    5\,Hz Geometry & 298 & 1.5s & 90.1\% & 76.5\% & 6.1\% & 0.802 & 0.812 \\
    \rowcolor{oursrow}\textbf{8\,Hz Geometry (Ours)} & 478 & 2.1s & 94.1\% & 86.0\% & 5.8\% & 0.811 & 0.820 \\
    15\,Hz Geometry & 896 & 3.5s & \textbf{100.0\%} & \textbf{100.0\%} & \textbf{5.6\%} & \textbf{0.812} & \textbf{0.821} \\
    \bottomrule
  \end{tabular}%
  }
\end{table}

\begin{figure*}[t]
  \centering
  \includegraphics[width=0.98\linewidth]{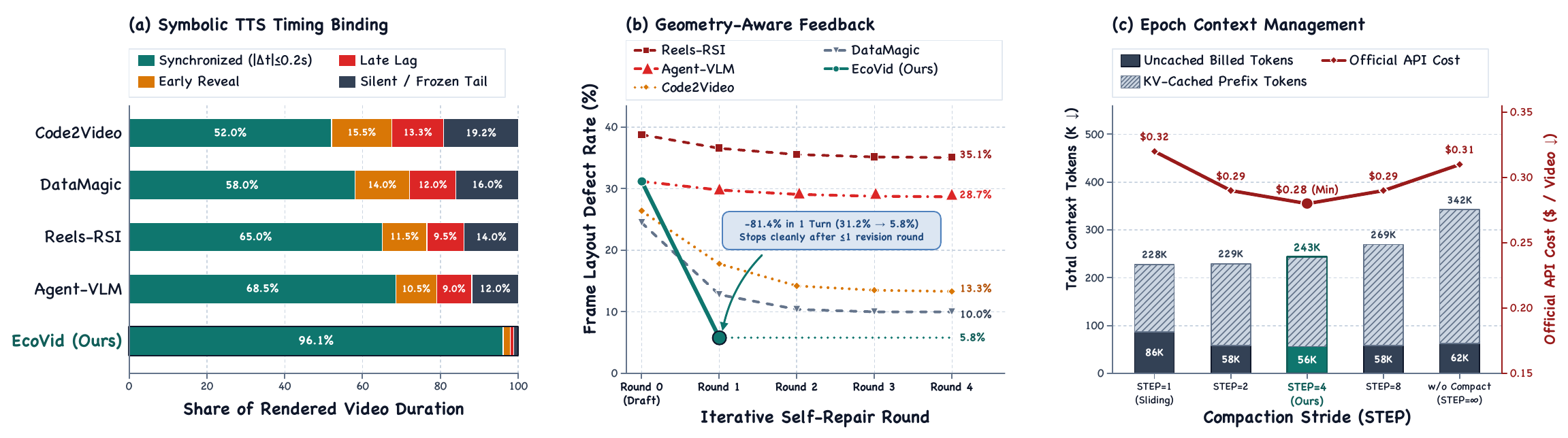}
  \caption{\textbf{Mechanism analysis across our three core harness components:} \textbf{(a)} rendered timing breakdown across TTS binding strategies, \textbf{(b)} rest-frame defect convergence across iterative repair rounds, and \textbf{(c)} prefix KV-cache reuse versus uncached billed tokens across compaction strides.}
  \label{fig:insights}
\end{figure*}

\paragraph{Effect of Symbolic TTS Timing Binding.}
To verify the effectiveness of Symbolic TTS timing binding, as shown in \Cref{tab:abl_main}, removing sentence-level binding drops VS from $0.811$ to $0.667$ and aesthetic score from $96.8$ to $89.8$.
Comparing \textbf{EcoVid} against other methods without sentence-level binding in \cref{fig:insights}a, it shows that baselines stay synchronized for only $52.0\%\text{--}68.5\%$ of the video duration, and with $12.0\%\text{--}19.2\%$ frozen tails. 
Symbolic binding lifts synchronized duration to $96.1\%$ and shrinks frozen tails to $1.2\%$.

\begin{figure}[t]
  \centering
  \includegraphics[width=\columnwidth]{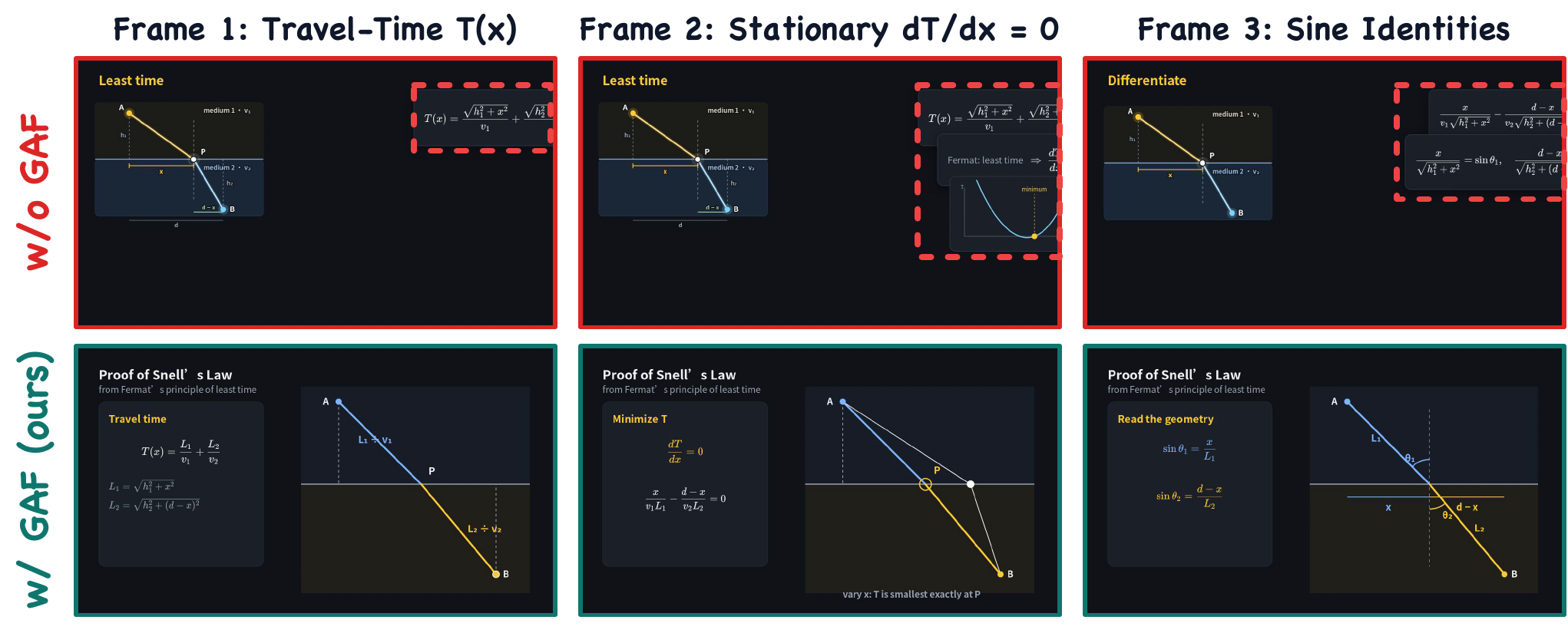}
  \caption{\textbf{Qualitative ablation of Geometry-Aware Feedback (GAF) on Remotion (\emph{Proof of Snell's Law}).} Red dashed boxes mark off-canvas formula clipping without GAF.}
  \label{fig:abl_gaf_qual}
\end{figure}

\paragraph{Effectiveness of Geometry-Aware Feedback.}
We used extensive quantitative and qualitative experiments to validate the effectiveness of the geometry-aware feedback (\textbf{GAF}). As shown in \Cref{tab:abl_main}, the method using GAF achieved significant improvements on AES, with scores rising from 88.4 to 96.8. 
This metric, along with M-KC, saw comprehensive gains across the board. At the same time, it substantially reduced API overhead and token consumption. 

GAF not only helps reduce compute costs, but also ensures the generated video has a reasonable layout, clean visuals, and fewer defects.
Furthermore, as seen in \Cref{tab:abl_probing}, compared to the traditional approach of using VLM for frame-sampling reading, GAF makes it easy to adjust the feedback frequency. As the sampling frequency increases, defect recall rises from $34.2\%$ (VLM stills) to $94.1\%$ at $8$\,Hz in $2.1$\,s, which we adopt for our final implementation.

Additionally, \cref{fig:insights}b illustrates GAF only requires a single round of inspection and repair to achieve a lower defect ratio compared to other baseline methods, quickly converging to an appropriate value and yielding a lower defect rate efficiently.
As also seen in \cref{fig:abl_gaf_qual}, GAF can significantly reduce the occurrence of defects caused by collisions between different formulas or elements in video keyframes.

\paragraph{Effectiveness of Cache-Aware Context Management.}
\begin{table}[t]
  \centering
  \caption{\textbf{Ablation of context compaction stride and verbatim tail length on KV-cache efficiency and video quality.}}
  \label{tab:abl_compaction}
  \scriptsize
  \setlength{\tabcolsep}{2.2pt}
  \resizebox{\columnwidth}{!}{%
  \begin{tabular}{ll|cccc|cc}
    \toprule
    \textbf{\textit{Policy}} & \textbf{\textit{$(s, w)$}} & \textbf{\textit{Total$\downarrow$}} & \textbf{\textit{Uncach.$\downarrow$}} & \textbf{\textit{Hit\%$\uparrow$}} & \textbf{\textit{Cost$\downarrow$}} & \textbf{\textit{Calls$\downarrow$}} & \textbf{\textit{VS$\uparrow$}} \\
    \midrule
    w/o Compaction & $\infty$ & 342K & 62K & \textbf{80.3\%} & \$0.31 & 15.6 & 0.796 \\
    Sliding Win. & $(1, 1)$ & \textbf{207K} & 64K & 64.4\% & \$0.29 & 16.8 & 0.768 \\
    Sliding Win. & $(1, 2)$ & 228K & 86K & 57.5\% & \$0.32 & 16.4 & 0.786 \\
    Epoch & $(2, 2)$ & 229K & 58K & 71.3\% & \$0.29 & 16.1 & 0.780 \\
    Epoch & $(4, 1)$ & 229K & 57K & 71.8\% & \$0.29 & 13.8 & 0.803 \\
    \rowcolor{oursrow}\textbf{Epoch (EcoVid)} & $\mathbf{(4, 2)}$ & 243K & \textbf{56K} & 74.1\% & \textbf{\$0.28} & \textbf{13.3} & \textbf{0.811} \\
    Epoch & $(8, 2)$ & 269K & 58K & 76.1\% & \$0.29 & 14.2 & 0.799 \\
    \bottomrule
  \end{tabular}%
  }
\end{table}

We compared the cost of uncompacted history, sliding-window pruning, and epoch compaction across strides $s \in \{2, 4, 8\}$ and tail windows $w \in \{1, 2\}$ in \Cref{tab:abl_main,tab:abl_compaction} and \cref{fig:insights}c.
It reveals a U-shaped cost feature: the highest token consumption and the most cache misses both lead to higher actual costs.
Without compacting history leads to the highest token consumption.
However, compacting history at every step results in the highest cache-miss rate.

We use the hyperparameters $s=4, w=2$ to balance the two, achieving the lowest per-video cost.
This setting ensures that context pruning runs every 4 turns, while the context for the most recent 2 turns remains untouched.
As a result, the average token consumption is 243K and the average API cost is $0.28\$$ per video.

\begin{table}[t]
  \centering
  \scriptsize
  \renewcommand{\arraystretch}{0.96}
  \setlength{\tabcolsep}{1.8pt}
  \caption{\textbf{Model-agnostic evaluation across diverse LLM backbones (Agent-VLM vs.\ \textbf{EcoVid}).} Complete metrics are in the supplementary material.}
  \label{tab:abl_backbones}
  \resizebox{\columnwidth}{!}{%
  \begin{tabular}{ll|cc|cccc}
    \toprule
    \textbf{\textit{Backbone}} & \textbf{\textit{Method}}
    & \textbf{\textit{Tok.$\downarrow$}} & \textbf{\textit{Unc.$\downarrow$}}
    & \textbf{\textit{Def.$\downarrow$}} & \textbf{\textit{VS$\uparrow$}} & \textbf{\textit{M-KC$\uparrow$}} & \textbf{\textit{AES$\uparrow$}} \\
    \midrule
    \multirow{2}{*}{GLM-5.3}
    & Agent-VLM & 266K & 51K & 30.4\% & 0.697 & 0.686 & 87.6 \\
    & \cellcolor{oursrow}\textbf{EcoVid} & \cellcolor{oursrow}\textbf{154K} & \cellcolor{oursrow}\textbf{39K} & \cellcolor{oursrow}\textbf{6.1\%} & \cellcolor{oursrow}\textbf{0.806} & \cellcolor{oursrow}\textbf{0.816} & \cellcolor{oursrow}\textbf{96.8} \\
    \midrule
    \multirow{2}{*}{DeepSeek-V4.1}
    & Agent-VLM & 1,629K & 84K & 26.8\% & 0.729 & 0.711 & 89.9 \\
    & \cellcolor{oursrow}\textbf{EcoVid} & \cellcolor{oursrow}\textbf{210K} & \cellcolor{oursrow}\textbf{43K} & \cellcolor{oursrow}\textbf{5.2\%} & \cellcolor{oursrow}\textbf{0.813} & \cellcolor{oursrow}\textbf{0.819} & \cellcolor{oursrow}\textbf{97.1} \\
    \midrule
    \multirow{2}{*}{Kimi-K3.1-Pro}
    & Agent-VLM & 305K & 109K & 48.2\% & 0.615 & 0.562 & 81.9 \\
    & \cellcolor{oursrow}\textbf{EcoVid} & \cellcolor{oursrow}\textbf{127K} & \cellcolor{oursrow}\textbf{64K} & \cellcolor{oursrow}\textbf{0.9\%} & \cellcolor{oursrow}\textbf{0.866} & \cellcolor{oursrow}\textbf{0.837} & \cellcolor{oursrow}\textbf{98.9} \\
    \midrule
    \multirow{2}{*}{Gemini-4-Argon}
    & Agent-VLM & 127K & 76K & 28.7\% & 0.736 & 0.707 & 89.4 \\
    & \cellcolor{oursrow}\textbf{EcoVid} & \cellcolor{oursrow}\textbf{58K} & \cellcolor{oursrow}\textbf{32K} & \cellcolor{oursrow}\textbf{5.2\%} & \cellcolor{oursrow}\textbf{0.812} & \cellcolor{oursrow}\textbf{0.820} & \cellcolor{oursrow}\textbf{97.1} \\
    \midrule
    \multirow{2}{*}{GPT-5.6}
    & Agent-VLM & 245K & 121K & 28.9\% & 0.686 & 0.677 & 87.1 \\
    & \cellcolor{oursrow}\textbf{EcoVid} & \cellcolor{oursrow}\textbf{82K} & \cellcolor{oursrow}\textbf{31K} & \cellcolor{oursrow}\textbf{6.3\%} & \cellcolor{oursrow}\textbf{0.815} & \cellcolor{oursrow}\textbf{0.818} & \cellcolor{oursrow}\textbf{96.2} \\
    \midrule
    \multirow{2}{*}{Claude-Opus-5.5}
    & Agent-VLM & 790K & 94K & 22.3\% & 0.688 & 0.662 & 88.2 \\
    & \cellcolor{oursrow}\textbf{EcoVid} & \cellcolor{oursrow}\textbf{107K} & \cellcolor{oursrow}\textbf{42K} & \cellcolor{oursrow}\textbf{0.2\%} & \cellcolor{oursrow}\textbf{0.788} & \cellcolor{oursrow}\textbf{0.841} & \cellcolor{oursrow}\textbf{99.2} \\
    \bottomrule
  \end{tabular}%
  }
\end{table}

\paragraph{Model-Agnostic Generalization.}
To verify that our harness is model-agnostic, \Cref{tab:abl_backbones} evaluates six open-weight and frontier backbones under matched medium thinking effort with and without \textbf{EcoVid}.
As shown in \Cref{tab:abl_backbones}, the performance of each backbone improves significantly after applying our method.
Both the token consumption and the number of cache misses drop notably, with tokens decreasing by over $50\%$.
In addition, the generated video quality metrics show comprehensive, stable, and consistent improvements.
These include a substantial reduction in layout defect rates, along with enhanced visual synchronization and improved multimodal knowledge coverage and aesthetic scores.
Together, these results demonstrate that our harness is model-agnostic and exhibits strong generalization capabilities.

\section{Conclusion}
\label{sec:conclusion}

We introduced \textbf{EcoVid} (\textbf{E}fficient \textbf{co}de-to-\textbf{Vid}eo), a token-saving, and geometry-aware agent harness for narrated code-to-video generation built around \emph{Symbolic TTS Timing Binding}, \emph{Geometry-Aware Feedback}, and \emph{Cache-Aware Context Management}.
Together, these methods reduce the context length and computational cost overhead and provide effective context management, achieving state-of-the-art results across various metrics.
\textbf{EcoVid} cuts token usage by $\mathbf{84\%}$, API cost by $\mathbf{57\%}$, and layout defects to $\mathbf{2.8\%}$, keeping animations synchronized with narration across Bench60 on Remotion, HyperFrames, and Manim.
In conclusion, offloading mechanical bookkeeping to our harness is more efficient and effective than scaling interaction frequency or substituting frontier models, as our harness is model-agnostic.

\clearpage
{
    \small
    \bibliographystyle{ieeenat_fullname}
    \bibliography{main}
}

\clearpage
\maketitlesupplementary
\appendix

\renewcommand{\topfraction}{0.95}
\renewcommand{\bottomfraction}{0.95}
\renewcommand{\textfraction}{0.05}
\renewcommand{\floatpagefraction}{0.90}
\renewcommand{\dbltopfraction}{0.95}
\renewcommand{\dblfloatpagefraction}{0.90}

\section*{Overview of Supplementary Material}
\begin{figure}[t]
  \centering
  \includegraphics[width=0.98\columnwidth,keepaspectratio]{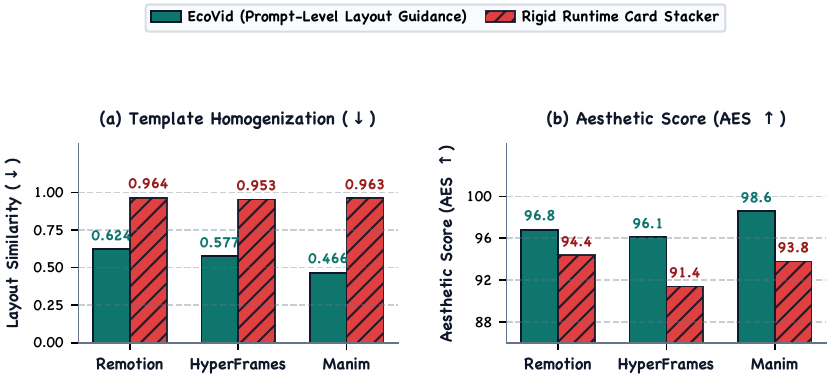}
  \caption{\textbf{Effect of layout guidance strategy across Remotion, HyperFrames, and Manim on \textsc{Bench60}.}
  \textbf{(a)} Enforcing a fixed runtime layout template produces nearly identical spatial occupancy across diverse topics, whereas prompt-level layout guidance combined with geometry-aware feedback preserves topic-specific visual organization.
  \textbf{(b)} Preserving flexible spatial composition consistently improves aesthetic scores across all three rendering engines.}
  \label{fig:supp_i5_bar}
\end{figure}

This supplementary document provides additional ablation studies, full-benchmark comparisons across \textsc{Bench60} and \textsc{MMMC-40}, token and prefix-caching analyses, cross-domain and bilingual evaluations, model-agnostic generalization experiments, qualitative visual comparisons, and formal metric definitions:
\begin{itemize}
  \item \textbf{Design Ablations and Baseline Comparisons (\Cref{sec:supp_evolution}):} Examines how authoring granularity, motion filtering, and layout guidance affect generation quality across Remotion, HyperFrames, and Manim in \Cref{tab:hf_ablations} and \cref{fig:supp_i5_bar}, followed by comparisons against declarative compilation pipelines and iterative agentic workflows on \textsc{Bench60} and \textsc{MMMC-40} in \Cref{tab:supp_external_academic} and \cref{fig:supp_pareto}.
  \item \textbf{Token Economics and Prefix Cache Analysis (\Cref{sec:supp_token_cost}):} Analyzes how prefix KV-cache reuse and incremental code edits reduce actual API cost under standard billing in \Cref{tab:supp_token_cost}, and attributes model calls and tokens between mechanical bookkeeping and high-level visual authoring in \Cref{tab:supp_trajectory}.
  \item \textbf{Domain Sensitivity and Bilingual Evaluation (\Cref{sec:supp_categories}):} Evaluates how different subject domains and bilingual scripts affect layout quality and audio-visual alignment in \Cref{tab:supp_categories,tab:supp_bilingual}.
  \item \textbf{Model-Agnostic Generalization (\Cref{sec:supp_small_model}):} Evaluates \textbf{EcoVid} across open-weight and closed frontier backbones on Remotion, HyperFrames, and Manim in \Cref{tab:supp_model_agnostic}.
  \item \textbf{Qualitative Comparisons (\Cref{sec:supp_filmstrips}):} Presents visual comparisons across geometry-aware feedback ablations, rendering engines, bilingual topics, mathematical derivations, layout strategies, and external baselines in \Cref{fig:suppl_gaf_manim,fig:suppl_agentvlm_comp,fig:supp_zh,fig:supp_sci,fig:supp_abl,fig:supp_ext}.
  \item \textbf{System Prompts and Evaluation Metrics (\Cref{sec:supp_impl}):} Details the system prompts for Remotion, HyperFrames, and Manim alongside the mathematical definitions of our five evaluation metrics.
  \item \textbf{Benchmark Design and Topic Catalog (\Cref{sec:supp_topics}):} Describes the benchmark protocol and lists all topics in \textsc{Bench60} and \textsc{MMMC-40} in \Cref{tab:supp_topics}.
\end{itemize}

\section{Design Ablations and Baseline Comparisons}
\label{sec:supp_evolution}

\begin{table}[t]
  \centering
  \scriptsize
  \renewcommand{\arraystretch}{1.04}
  \setlength{\tabcolsep}{1.6pt}
  \caption{\textbf{Ablation of authoring granularity, motion filtering, and layout guidance strategies across Remotion, HyperFrames, and Manim on \textsc{Bench60}.}}
  \label{tab:hf_ablations}
  \resizebox{\columnwidth}{!}{%
  \begin{tabular}{llccccccc}
    \toprule
    \textbf{\textit{Carrier}} & \textbf{\textit{Variant}} & \textbf{\textit{Tok.$\downarrow$}} & \textbf{\textit{Calls$\downarrow$}} & \textbf{\textit{M-KC$\uparrow$}} & \textbf{\textit{VS$\uparrow$}} & \textbf{\textit{Vis-Rel$\uparrow$}} & \textbf{\textit{AES$\uparrow$}} & \textbf{\textit{Sim.$\downarrow$}} \\
    \midrule
    \multirow{4}{*}{Remotion}
    & \cellcolor{oursrow}\textbf{EcoVid (Ours)} & \cellcolor{oursrow}243K & \cellcolor{oursrow}13.3 & \cellcolor{oursrow}0.820 & \cellcolor{oursrow}\textbf{0.811} & \cellcolor{oursrow}\textbf{4.69} & \cellcolor{oursrow}\textbf{96.8} & \cellcolor{oursrow}0.624 \\
    & w/ Post-Hoc Motion Filter & 328K\upbad{85K} & 19.6\upbad{6.3} & 0.738\downbad{0.082} & 0.710\downbad{0.101} & 4.38\downbad{0.31} & 93.2\downbad{3.6} & \textbf{0.488}\downgood{0.136} \\
    & Merged Single-Turn Script \& Code & \textbf{156K}\downgood{87K} & \textbf{9.9}\downgood{3.4} & 0.793\downbad{0.027} & 0.796\downbad{0.015} & 4.41\downbad{0.28} & 93.1\downbad{3.7} & 0.591\downgood{0.033} \\
    & Fixed Runtime Layout Template & 179K\downgood{64K} & 10.3\downgood{3.0} & \textbf{0.834}\upgood{0.014} & 0.807\downbad{0.004} & 4.58\downbad{0.11} & 94.4\downbad{2.4} & 0.964\upbad{0.340} \\
    \midrule
    \multirow{4}{*}{HyperFrames}
    & \cellcolor{oursrow}\textbf{EcoVid (Ours)} & \cellcolor{oursrow}210K & \cellcolor{oursrow}11.4 & \cellcolor{oursrow}0.811 & \cellcolor{oursrow}0.769 & \cellcolor{oursrow}\textbf{4.60} & \cellcolor{oursrow}\textbf{96.1} & \cellcolor{oursrow}0.577 \\
    & w/ Post-Hoc Motion Filter & 198K\downgood{12K} & \textbf{9.9}\downgood{1.5} & 0.802\downbad{0.009} & 0.740\downbad{0.029} & 4.54\downbad{0.06} & 95.3\downbad{0.8} & \textbf{0.555}\downgood{0.022} \\
    & Merged Single-Turn Script \& Code & 211K\upbad{1K} & 10.5\downgood{0.9} & 0.779\downbad{0.032} & \textbf{0.778}\upgood{0.009} & 4.46\downbad{0.14} & 94.2\downbad{1.9} & 0.567\downgood{0.010} \\
    & Fixed Runtime Layout Template & \textbf{186K}\downgood{24K} & 11.1\downgood{0.3} & \textbf{0.812}\upgood{0.001} & 0.734\downbad{0.035} & 4.48\downbad{0.12} & 91.4\downbad{4.7} & 0.953\upbad{0.376} \\
    \midrule
    \multirow{4}{*}{Manim}
    & \cellcolor{oursrow}\textbf{EcoVid (Ours)} & \cellcolor{oursrow}183K & \cellcolor{oursrow}12.8 & \cellcolor{oursrow}\textbf{0.829} & \cellcolor{oursrow}\textbf{0.840} & \cellcolor{oursrow}\textbf{4.64} & \cellcolor{oursrow}\textbf{98.6} & \cellcolor{oursrow}0.466 \\
    & w/ Post-Hoc Motion Filter & 204K\upbad{21K} & 14.1\upbad{1.3} & 0.806\downbad{0.023} & 0.812\downbad{0.028} & 4.52\downbad{0.12} & 97.1\downbad{1.5} & \textbf{0.451}\downgood{0.015} \\
    & Merged Single-Turn Script \& Code & 176K\downgood{7K} & \textbf{11.2}\downgood{1.6} & 0.794\downbad{0.035} & 0.825\downbad{0.015} & 4.48\downbad{0.16} & 96.2\downbad{2.4} & 0.478\upbad{0.012} \\
    & Fixed Runtime Layout Template & \textbf{164K}\downgood{19K} & 11.6\downgood{1.2} & 0.825\downbad{0.004} & 0.804\downbad{0.036} & 4.51\downbad{0.13} & 93.8\downbad{4.8} & 0.963\upbad{0.497} \\
    \bottomrule
  \end{tabular}%
  }
\end{table}

\subsection{Analysis of Harness Design Choices}
\label{sec:supp_knowhow}

While offloading mechanical timing and spatial bookkeeping from the LLM is essential, a key design question is how strictly the harness should constrain the agent's authoring workflow. The controlled ablations across Remotion, HyperFrames, and Manim in \Cref{tab:hf_ablations} and \cref{fig:supp_i5_bar} highlight four practical design principles:
\begin{enumerate}
  \item \textbf{Pre-TTS Anchor Validation vs.\ Post-Hoc Motion Filtering:}
  Without sentence-level anchor enforcement, the agent often places a single timing marker at the start of a multi-sentence paragraph, leaving later sentences visually static. Validating that every sentence begins with its own symbolic anchor prior to speech synthesis resolves this issue directly at the scripting stage.
  By contrast, enforcing additional motion after code generation through a post-hoc motion filter, which rejects any sentence interval that lacks continuous element movement, consistently degrades both visual synchronization and visual relevance across all three rendering engines, as shown in \Cref{tab:hf_ablations}. Once each sentence already has a valid timing anchor, forcing extra intra-sentence movement causes the agent to insert unnecessary jitter or split coherent diagrams into fragmented steps.

  \item \textbf{Decoupled Two-Turn Authoring vs.\ Single-Turn Generation:}
  To reduce the number of interaction turns, a natural alternative is to merge the first two turns so the model generates both the narration script and the animation code in a single response.
  As shown in \Cref{tab:hf_ablations}, merging these stages consistently reduces multimodal knowledge coverage, visual relevance, and aesthetic scores across all three rendering engines. When required to produce both narration and hundreds of lines of animation code within a single output window, the model tends to shorten the narration script and omit intermediate derivation steps or concrete examples. Moreover, the initial token savings are largely offset by additional repair turns needed to fix rougher initial code. Separating script generation in Turn~1 from code authoring in Turn~2 allows the model to establish a complete pedagogical structure before writing animation code.

  \item \textbf{Prompt-Level Layout Guidance vs.\ Fixed Runtime Templates:}
  Because text overlap is a common spatial defect, one engineering shortcut is to provide a fixed runtime layout template that locks text cards into a predefined column by construction.
  Although this fixed template slightly reduces repair tokens in \Cref{tab:hf_ablations}, it leads to severe layout homogenization across topics. Rather than designing visual layouts suited to each subject, the coding agent places every explanation into the same vertical card stack. As shown in \cref{fig:supp_i5_bar}a, the fixed runtime template drives pairwise spatial occupancy similarity above $0.95$ across all three rendering engines while reducing aesthetic scores in \cref{fig:supp_i5_bar}b. Providing layout conventions as prompt-level guidance, backed by geometry-aware feedback to resolve actual collisions, preserves diverse, topic-specific visual designs while maintaining low defect rates.

  \item \textbf{Bounded Post-Render Visual Review:}
  Once structured scene layout checks pass, allowing at most one post-render review pass on a nine-frame contact sheet helps the model address global visual balance or color grouping issues that bounding-box checks alone do not capture. Removing this single-pass limit leads the agent to spend extra turns making minor cosmetic adjustments, increasing token usage and API cost without further reducing layout defects or improving pedagogical quality.
\end{enumerate}

\begin{figure*}[t]
  \centering
  \includegraphics[width=0.98\linewidth,keepaspectratio]{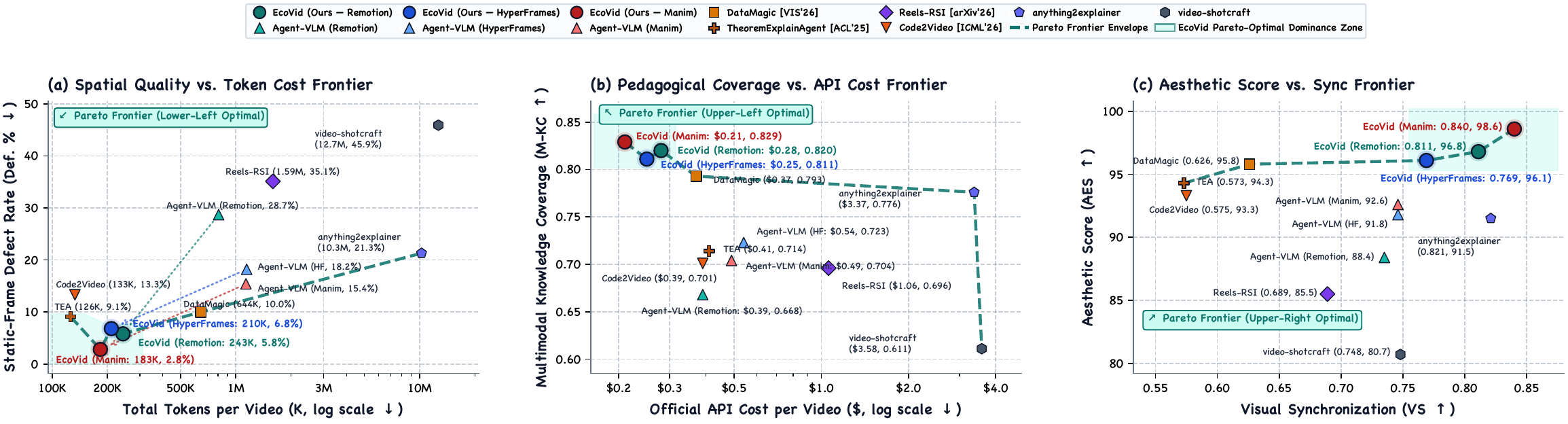}
  \caption{\textbf{Cross-engine efficiency and quality trade-offs on \textsc{Bench60}:} \textbf{(a)} static-frame defect rate versus total token consumption, \textbf{(b)} multimodal knowledge coverage versus official API cost, and \textbf{(c)} aesthetic score versus visual synchronization.}
  \label{fig:supp_pareto}
  \label{fig:supp_external_pareto}
  \label{fig:supp_new_pareto}
  \label{fig:supp_new_metrics}
\end{figure*}

\begin{table*}[t]
  \centering
  \scriptsize
  \renewcommand{\arraystretch}{1.04}
  \setlength{\tabcolsep}{3.1pt}
  \caption{\textbf{Comparison with open-source academic pipelines and community skills on \textsc{Bench60} and \textsc{MMMC-40}.} All systems use Qwen3.8-Max-Medium and identical TTS settings.}
  \label{tab:supp_external_academic}
  \begin{tabular}{llccccccccccccc}
    \toprule
    \multirow{2}{*}{\textbf{\textit{Carrier}}}
    & \multirow{2}{*}{\textbf{\textit{System}}}
    & \multicolumn{5}{c}{\textbf{\textit{Efficiency \& API Cost (Qwen3.8-Max)}}}
    & \multicolumn{5}{c}{\textbf{\textit{Quality, Synchronization \& Pedagogy}}}
    & \multicolumn{3}{c}{\textbf{\textit{Sync Diagnostics}}} \\
    \cmidrule(lr){3-7} \cmidrule(lr){8-12} \cmidrule(lr){13-15}
    & & \textbf{\textit{Tokens$\downarrow$}} & \textbf{\textit{Uncach.$\downarrow$}} & \textbf{\textit{Cost$\downarrow$}} & \textbf{\textit{Calls$\downarrow$}} & \textbf{\textit{Time$\downarrow$}}
    & \textbf{\textit{Def.$\downarrow$}} & \textbf{\textit{VS$\uparrow$}} & \textbf{\textit{Vis-Rel$\uparrow$}} & \textbf{\textit{M-KC$\uparrow$}} & \textbf{\textit{AES$\uparrow$}}
    & \textbf{\textit{Silent$\downarrow$}} & \textbf{\textit{Lag$\downarrow$}} & \textbf{\textit{Lead$\downarrow$}} \\
    \midrule
    \rowcolor{secgray}\multicolumn{15}{c}{\textit{\textbf{Part I: Bench60 (30 English and 30 Chinese Topics)}}} \\
    \midrule
    \multirow{4}{*}{Remotion}
    & anything2explainer~\cite{wei2026anything2explainer} & 10,258K & 574K & \$3.37 & 114.7 & 55.2 & 21.3\% & \textbf{0.821} & 3.86 & 0.776 & 91.5 & 1.5\% & \textbf{7.4\%} & \textbf{5.6\%} \\
    & video-shotcraft~\cite{wei2026shotcraft} & 12,674K & 501K & \$3.58 & 121.5 & 87.5 & 45.9\% & 0.748 & 3.40 & 0.611 & 80.7 & 3.0\% & 9.3\% & 9.1\% \\
    & DataMagic~\cite{xie2026datamagic} & 644K & 73K & \$0.37 & 22.9 & 13.1 & 10.0\% & 0.626 & 4.46 & 0.793 & 95.8 & 2.1\% & 16.0\% & 12.0\% \\
    & \cellcolor{oursrow}\textbf{EcoVid (Ours)} & \cellcolor{oursrow}\textbf{243K} & \cellcolor{oursrow}\textbf{56K} & \cellcolor{oursrow}\textbf{\$0.28} & \cellcolor{oursrow}\textbf{13.3} & \cellcolor{oursrow}\textbf{11.6} & \cellcolor{oursrow}\textbf{5.8\%} & \cellcolor{oursrow}0.811 & \cellcolor{oursrow}\textbf{4.69} & \cellcolor{oursrow}\textbf{0.820} & \cellcolor{oursrow}\textbf{96.8} & \cellcolor{oursrow}\textbf{1.2\%} & \cellcolor{oursrow}8.8\% & \cellcolor{oursrow}7.1\% \\
    \midrule
    \multirow{2}{*}{HyperFrames}
    & Reels-RSI~\cite{lin2026reelsrsi} & 1,592K & 122K & \$1.06 & 57.6 & 26.6 & 35.1\% & 0.689 & 3.24 & 0.696 & 85.5 & 1.9\% & 16.1\% & 7.3\% \\
    & \cellcolor{oursrow}\textbf{EcoVid (Ours)} & \cellcolor{oursrow}\textbf{210K} & \cellcolor{oursrow}\textbf{50K} & \cellcolor{oursrow}\textbf{\$0.25} & \cellcolor{oursrow}\textbf{11.4} & \cellcolor{oursrow}\textbf{10.8} & \cellcolor{oursrow}\textbf{6.8\%} & \cellcolor{oursrow}\textbf{0.769} & \cellcolor{oursrow}\textbf{4.60} & \cellcolor{oursrow}\textbf{0.811} & \cellcolor{oursrow}\textbf{96.1} & \cellcolor{oursrow}\textbf{1.1\%} & \cellcolor{oursrow}\textbf{11.2\%} & \cellcolor{oursrow}\textbf{6.8\%} \\
    \midrule
    \multirow{3}{*}{Manim}
    & TheoremExplainAgent~\cite{ku2025theoremexplainagent} & \textbf{126K} & 76K & \$0.41 & 14.5 & 15.1 & 9.1\% & 0.573 & 4.08 & 0.714 & 94.3 & 4.1\% & 17.7\% & 11.8\% \\
    & Code2Video~\cite{chen2026code2video} & 133K & 96K & \$0.39 & 23.2 & 16.0 & 13.3\% & 0.575 & 3.40 & 0.701 & 93.3 & 1.9\% & 13.1\% & 19.2\% \\
    & \cellcolor{oursrow}\textbf{EcoVid (Ours)} & \cellcolor{oursrow}183K & \cellcolor{oursrow}\textbf{39K} & \cellcolor{oursrow}\textbf{\$0.21} & \cellcolor{oursrow}\textbf{12.8} & \cellcolor{oursrow}\textbf{10.3} & \cellcolor{oursrow}\textbf{2.8\%} & \cellcolor{oursrow}\textbf{0.840} & \cellcolor{oursrow}\textbf{4.64} & \cellcolor{oursrow}\textbf{0.829} & \cellcolor{oursrow}\textbf{98.6} & \cellcolor{oursrow}\textbf{0.8\%} & \cellcolor{oursrow}\textbf{8.1\%} & \cellcolor{oursrow}\textbf{5.3\%} \\
    \midrule
    \rowcolor{secgray}\multicolumn{15}{c}{\textit{\textbf{Part II: MMMC-40 (40 Mathematical Topics)}}} \\
    \midrule
    \multirow{4}{*}{Remotion}
    & anything2explainer~\cite{wei2026anything2explainer} & 10,590K & 588K & \$3.46 & 118.4 & 56.8 & 24.6\% & 0.842 & 3.92 & 0.784 & 92.4 & 1.4\% & 6.8\% & 5.4\% \\
    & video-shotcraft~\cite{wei2026shotcraft} & 13,136K & 512K & \$3.70 & 124.2 & 89.4 & 42.8\% & 0.764 & 3.48 & 0.638 & 82.1 & 2.8\% & 8.6\% & 8.4\% \\
    & DataMagic~\cite{xie2026datamagic} & 578K & \textbf{71K} & \$0.34 & 21.1 & \textbf{12.0} & 19.6\% & 0.621 & 4.38 & 0.739 & 92.5 & 1.6\% & 16.1\% & 14.9\% \\
    & \cellcolor{oursrow}\textbf{EcoVid (Ours)} & \cellcolor{oursrow}\textbf{302K} & \cellcolor{oursrow}\textbf{71K} & \cellcolor{oursrow}\textbf{\$0.33} & \cellcolor{oursrow}\textbf{15.8} & \cellcolor{oursrow}13.3 & \cellcolor{oursrow}\textbf{1.8\%} & \cellcolor{oursrow}\textbf{0.888} & \cellcolor{oursrow}\textbf{4.75} & \cellcolor{oursrow}\textbf{0.865} & \cellcolor{oursrow}\textbf{99.3} & \cellcolor{oursrow}\textbf{0.4\%} & \cellcolor{oursrow}\textbf{5.6\%} & \cellcolor{oursrow}\textbf{4.2\%} \\
    \midrule
    \multirow{2}{*}{HyperFrames}
    & Reels-RSI~\cite{lin2026reelsrsi} & 1,807K & 113K & \$1.01 & 50.1 & 30.4 & 25.6\% & 0.680 & 3.43 & 0.721 & 88.5 & 1.7\% & 18.0\% & 7.5\% \\
    & \cellcolor{oursrow}\textbf{EcoVid (Ours)} & \cellcolor{oursrow}\textbf{310K} & \cellcolor{oursrow}\textbf{75K} & \cellcolor{oursrow}\textbf{\$0.35} & \cellcolor{oursrow}\textbf{15.6} & \cellcolor{oursrow}\textbf{15.8} & \cellcolor{oursrow}\textbf{1.0\%} & \cellcolor{oursrow}\textbf{0.779} & \cellcolor{oursrow}\textbf{4.78} & \cellcolor{oursrow}\textbf{0.883} & \cellcolor{oursrow}\textbf{99.6} & \cellcolor{oursrow}\textbf{1.2\%} & \cellcolor{oursrow}\textbf{11.5\%} & \cellcolor{oursrow}\textbf{6.6\%} \\
    \midrule
    \multirow{3}{*}{Manim}
    & TheoremExplainAgent~\cite{ku2025theoremexplainagent} & 147K & 91K & \$0.47 & 14.5 & 16.9 & 7.4\% & 0.549 & 4.21 & 0.750 & 95.0 & 3.2\% & 20.0\% & 10.9\% \\
    & Code2Video~\cite{chen2026code2video} & \textbf{146K} & 106K & \$0.42 & 26.9 & 17.3 & 11.7\% & 0.553 & 3.76 & 0.670 & 93.7 & 1.9\% & 13.9\% & 21.5\% \\
    & \cellcolor{oursrow}\textbf{EcoVid (Ours)} & \cellcolor{oursrow}209K & \cellcolor{oursrow}\textbf{44K} & \cellcolor{oursrow}\textbf{\$0.23} & \cellcolor{oursrow}\textbf{14.2} & \cellcolor{oursrow}\textbf{11.2} & \cellcolor{oursrow}\textbf{2.4\%} & \cellcolor{oursrow}\textbf{0.902} & \cellcolor{oursrow}\textbf{4.55} & \cellcolor{oursrow}\textbf{0.855} & \cellcolor{oursrow}\textbf{98.9} & \cellcolor{oursrow}\textbf{0.3\%} & \cellcolor{oursrow}\textbf{6.1\%} & \cellcolor{oursrow}\textbf{2.8\%} \\
    \bottomrule
  \end{tabular}
\end{table*}

\subsection{Comparison with Academic Pipelines and Community Skills}
\Cref{tab:supp_external_academic} and \cref{fig:supp_pareto} compare \textbf{EcoVid} against open-source community skills and academic code-to-video pipelines on \textsc{Bench60} and \textsc{MMMC-40}. Examining the quality and efficiency trade-offs in \cref{fig:supp_pareto} clarifies the core bottlenecks of existing approaches:
\begin{itemize}
  \item \textbf{High Token Overhead in Open-Source Community Skills (\cref{fig:supp_pareto}a):}
  Multi-stage community skills rely on long chains of planning, styling, and screenshot-based critique prompts. While fine-grained beat scripting allows these workflows to achieve reasonable audio-visual alignment, the lack of a structured scene layout verifier leads to over a hundred model calls and millions of tokens per video. Furthermore, because rendered screenshots do not provide exact element coordinates, a substantial fraction of stationary frames still exhibit text overlaps or canvas overflows.

  \item \textbf{Timing Drift in Declarative Compilation Pipelines (\cref{fig:supp_pareto}b,c):}
  Declarative compilation pipelines generate animations through structured intermediate specifications or scene-level planner-coder stages. This modular design produces clean static geometry and moderate defect rates, but introduces noticeable audio-visual timing drift. Because scene durations and internal animation steps are allocated at coarse scene boundaries rather than bound to individual sentence timestamps, visual transitions frequently lag behind or lead ahead of the voiceover.

  \item \textbf{Layout Collisions in Iterative VLM-Guided Workflows (\cref{fig:supp_pareto}a,c):}
  Iterative workflows that pair frame generation workers with a VLM judge can produce expressive motion transitions, but struggle to resolve spatial collisions on dense diagrams. When formulas or callout labels overlap with animated shapes, the VLM judge cannot reliably estimate the required coordinate adjustments from rendered images alone, leaving high rest-frame defect rates despite substantial token usage.

  \item \textbf{Balanced Quality and Efficiency with EcoVid:}
  By combining symbolic TTS timing binding with geometry-aware feedback on the structured scene layout, \textbf{EcoVid} consistently achieves low layout defect rates, strong visual synchronization, and high multimodal knowledge coverage across both \textsc{Bench60} and \textsc{MMMC-40} while requiring substantially fewer tokens and lower API cost.
\end{itemize}

\section{Token Economics and Prefix Cache Analysis}
\label{sec:supp_token_cost}

\paragraph{Official API Billing Model.}
Evaluating multi-turn agents solely by total token count ($T_{\text{tot}} = T_{\text{unc}} + T_{\text{cache}} + T_{\text{out}}$) does not reflect actual monetary cost, because commercial LLM APIs price uncached input tokens, cached prefix tokens, and generated output tokens at very different rates. Under standard API pricing, uncached input tokens ($T_{\text{unc}}$) cost $\$1.60 / 10^6$, cached prefix reads ($T_{\text{cache}}$) cost $\$0.16 / 10^6$, and output tokens ($T_{\text{out}}$) cost $\$6.40 / 10^6$. \Cref{tab:supp_token_cost} therefore reports both raw token components and \textbf{Equivalent Billed Input Tokens} ($T_{\text{eq}}$):
\begin{equation}
  \begin{aligned}
    T_{\text{eq}} &= T_{\text{unc}} + 0.1\,T_{\text{cache}} + 4.0\,T_{\text{out}}, \\
    \text{Cost ($\$$)} &= 1.60 \times 10^{-6} \, T_{\text{eq}}.
  \end{aligned}
  \label{eq:api_cost}
\end{equation}

\begin{table*}[t]
  \centering
  \footnotesize
  \renewcommand{\arraystretch}{1.04}
  \setlength{\tabcolsep}{2.9pt}
  \caption{\textbf{Token component breakdown, prefix KV-cache reuse, and official API monetary cost across all rendering carriers.}}
  \label{tab:supp_token_cost}
  \begin{tabular}{llccccccccc}
    \toprule
    \multirow{2}{*}{\textbf{\textit{Carrier}}}
    & \multirow{2}{*}{\textbf{\textit{Method}}}
    & \multicolumn{5}{c}{\textbf{\textit{Token Component \& KV-Cache Breakdown}}}
    & \multicolumn{2}{c}{\textbf{\textit{Official API Billing}}}
    & \multicolumn{2}{c}{\textbf{\textit{Execution Latency}}} \\
    \cmidrule(lr){3-7} \cmidrule(lr){8-9} \cmidrule(lr){10-11}
    & & \textbf{\textit{Total$\downarrow$}} & \textbf{\textit{Uncach.$\downarrow$}} & \textbf{\textit{Cached}} & \textbf{\textit{Cache Hit$\uparrow$}} & \textbf{\textit{Output$\downarrow$}} & \textbf{\textit{Eq.\ In.$\downarrow$}} & \textbf{\textit{Cost ($\$$)$\downarrow$}} & \textbf{\textit{Calls$\downarrow$}} & \textbf{\textit{Time (m)$\downarrow$}} \\
    \midrule
    \rowcolor{secgray}\multicolumn{11}{c}{\textit{\textbf{Part I: Bench60 (30 English and 30 Chinese Topics)}}} \\
    \midrule
    \multirow{5}{*}{Remotion}
    & anything2explainer~\cite{wei2026anything2explainer} & 10,258K & 574K & 9,540K & 94.3\% & 143.8K & 2,104K & \$3.37 & 114.7 & 55.2 \\
    & video-shotcraft~\cite{wei2026shotcraft} & 12,674K & 501K & 12,040K & \textbf{96.0\%} & 132.3K & 2,235K & \$3.58 & 121.5 & 87.5 \\
    & DataMagic~\cite{xie2026datamagic} & 644K & 73K & 545K & 88.1\% & 25.9K & 232K & \$0.37 & 22.9 & 13.1 \\
    & Agent-VLM & 806K & 73K & 709K & 90.7\% & \textbf{24.4K} & 242K & \$0.39 & 29.7 & 14.4 \\
    & \cellcolor{oursrow}\textbf{EcoVid (Ours)} & \cellcolor{oursrow}\textbf{243K}\downgood{563K} & \cellcolor{oursrow}\textbf{56K}\downgood{17K} & \cellcolor{oursrow}161K & \cellcolor{oursrow}74.1\% & \cellcolor{oursrow}25.9K\upbad{1.5K} & \cellcolor{oursrow}\textbf{176K}\downgood{66K} & \cellcolor{oursrow}\textbf{\$0.28}\downgood{0.11} & \cellcolor{oursrow}\textbf{13.3}\downgood{16.4} & \cellcolor{oursrow}\textbf{11.6}\downgood{2.8} \\
    \midrule
    \multirow{3}{*}{HyperFrames}
    & Reels-RSI~\cite{lin2026reelsrsi} & 1,592K & 122K & 1,369K & \textbf{91.8\%} & 100.4K & 661K & \$1.06 & 57.6 & 26.6 \\
    & Agent-VLM & 1,146K & 98K & 1,013K & 91.2\% & 34.3K & 337K & \$0.54 & 30.4 & 16.5 \\
    & \cellcolor{oursrow}\textbf{EcoVid (Ours)} & \cellcolor{oursrow}\textbf{210K}\downgood{936K} & \cellcolor{oursrow}\textbf{50K}\downgood{48K} & \cellcolor{oursrow}137K & \cellcolor{oursrow}73.3\% & \cellcolor{oursrow}\textbf{23.6K}\downgood{10.7K} & \cellcolor{oursrow}\textbf{158K}\downgood{179K} & \cellcolor{oursrow}\textbf{\$0.25}\downgood{0.29} & \cellcolor{oursrow}\textbf{11.4}\downgood{19.0} & \cellcolor{oursrow}\textbf{10.8}\downgood{5.7} \\
    \midrule
    \multirow{4}{*}{Manim}
    & TheoremExplainAgent~\cite{ku2025theoremexplainagent} & \textbf{126K} & 76K & 5K & 6.2\% & 45.1K & 256K & \$0.41 & 14.5 & 15.1 \\
    & Code2Video~\cite{chen2026code2video} & 133K & 96K & 0K & 0.0\% & 36.9K & 243K & \$0.39 & 23.2 & 16.0 \\
    & Agent-VLM & 1,135K & 83K & 1,021K & \textbf{92.5\%} & 30.2K & 306K & \$0.49 & 36.4 & 19.9 \\
    & \cellcolor{oursrow}\textbf{EcoVid (Ours)} & \cellcolor{oursrow}183K\downgood{952K} & \cellcolor{oursrow}\textbf{39K}\downgood{44K} & \cellcolor{oursrow}124K & \cellcolor{oursrow}76.0\% & \cellcolor{oursrow}\textbf{19.7K}\downgood{10.5K} & \cellcolor{oursrow}\textbf{130K}\downgood{176K} & \cellcolor{oursrow}\textbf{\$0.21}\downgood{0.28} & \cellcolor{oursrow}\textbf{12.8}\downgood{23.6} & \cellcolor{oursrow}\textbf{10.3}\downgood{9.6} \\
    \midrule
    \rowcolor{secgray}\multicolumn{11}{c}{\textit{\textbf{Part II: MMMC-40 (40 Mathematical Topics)}}} \\
    \midrule
    \multirow{5}{*}{Remotion}
    & anything2explainer~\cite{wei2026anything2explainer} & 10,590K & 588K & 9,854K & 94.4\% & 148.0K & 2,165K & \$3.46 & 118.4 & 56.8 \\
    & video-shotcraft~\cite{wei2026shotcraft} & 13,136K & 512K & 12,486K & \textbf{96.1\%} & 138.0K & 2,313K & \$3.70 & 124.2 & 89.4 \\
    & DataMagic~\cite{xie2026datamagic} & 578K & \textbf{71K} & 483K & 87.2\% & \textbf{23.5K} & 213K & \$0.34 & 21.1 & \textbf{12.0} \\
    & Agent-VLM & 1,101K & 82K & 991K & 92.4\% & 28.5K & 295K & \$0.47 & 35.6 & 16.4 \\
    & \cellcolor{oursrow}\textbf{EcoVid (Ours)} & \cellcolor{oursrow}\textbf{302K}\downgood{799K} & \cellcolor{oursrow}\textbf{71K}\downgood{11K} & \cellcolor{oursrow}202K & \cellcolor{oursrow}73.9\% & \cellcolor{oursrow}28.9K\upbad{0.4K} & \cellcolor{oursrow}\textbf{207K}\downgood{88K} & \cellcolor{oursrow}\textbf{\$0.33}\downgood{0.14} & \cellcolor{oursrow}\textbf{15.8}\downgood{19.8} & \cellcolor{oursrow}13.3\downgood{3.1} \\
    \midrule
    \multirow{3}{*}{HyperFrames}
    & Reels-RSI~\cite{lin2026reelsrsi} & 1,807K & 113K & 1,604K & \textbf{93.4\%} & 90.1K & 634K & \$1.01 & 50.1 & 30.4 \\
    & Agent-VLM & 1,210K & 103K & 1,075K & 91.3\% & 32.0K & 338K & \$0.54 & 31.7 & 17.5 \\
    & \cellcolor{oursrow}\textbf{EcoVid (Ours)} & \cellcolor{oursrow}\textbf{310K}\downgood{900K} & \cellcolor{oursrow}\textbf{75K}\downgood{28K} & \cellcolor{oursrow}205K & \cellcolor{oursrow}73.1\% & \cellcolor{oursrow}\textbf{30.2K}\downgood{1.8K} & \cellcolor{oursrow}\textbf{217K}\downgood{121K} & \cellcolor{oursrow}\textbf{\$0.35}\downgood{0.19} & \cellcolor{oursrow}\textbf{15.6}\downgood{16.1} & \cellcolor{oursrow}\textbf{15.8}\downgood{1.7} \\
    \midrule
    \multirow{4}{*}{Manim}
    & TheoremExplainAgent~\cite{ku2025theoremexplainagent} & 147K & 91K & 5K & 5.5\% & 50.4K & 293K & \$0.47 & 14.5 & 16.9 \\
    & Code2Video~\cite{chen2026code2video} & \textbf{146K} & 106K & 0K & 0.0\% & 39.5K & 264K & \$0.42 & 26.9 & 17.3 \\
    & Agent-VLM & 998K & 75K & 898K & \textbf{92.3\%} & 25.5K & 267K & \$0.43 & 35.4 & 16.7 \\
    & \cellcolor{oursrow}\textbf{EcoVid (Ours)} & \cellcolor{oursrow}209K\downgood{789K} & \cellcolor{oursrow}\textbf{44K}\downgood{31K} & \cellcolor{oursrow}144K & \cellcolor{oursrow}76.5\% & \cellcolor{oursrow}\textbf{20.7K}\downgood{4.8K} & \cellcolor{oursrow}\textbf{141K}\downgood{126K} & \cellcolor{oursrow}\textbf{\$0.23}\downgood{0.20} & \cellcolor{oursrow}\textbf{14.2}\downgood{21.2} & \cellcolor{oursrow}\textbf{11.2}\downgood{5.5} \\
    \bottomrule
  \end{tabular}
\end{table*}

\paragraph{Impact of Prefix Caching and Incremental Edits on API Cost.}
The breakdown in \Cref{tab:supp_token_cost} highlights two key factors that govern real monetary cost across different agent architectures:
\begin{enumerate}
  \item \textbf{Prefix Cache Misses and Full-Script Rewrites in Stateless Pipelines:}
  On Manim, stateless multi-agent pipelines consume slightly fewer nominal tokens than \textbf{EcoVid}, yet their actual API cost is nearly twice as high, as shown in \Cref{tab:supp_token_cost}. This reversal stems from two design differences.
  First, stateless pipelines issue separate planner, coder, and critic calls for each scene. Because the prompt prefix changes on every invocation, their prefix KV-cache hit rate drops to nearly zero, so almost all input tokens are billed at the full uncached price. By contrast, \textbf{EcoVid} maintains a single multi-turn trajectory with cache-aware epoch compaction, serving over three-quarters of its input tokens from the discounted prefix cache.
  Second, when a scene fails verification in a stateless pipeline, the coder regenerates the entire scene file from scratch, substantially increasing expensive output tokens compared to the compact incremental edits used by \textbf{EcoVid}.

  \item \textbf{Visual Token Overhead from Repeated Screenshot Inspection:}
  Native VLM agents and iterative VLM-guided workflows achieve high prefix cache hit rates because they append turns to a single conversation history, but they repeatedly insert multi-frame screenshot grids into the context during visual debugging. As these image tokens accumulate across turns, both uncached token costs and inference latency grow rapidly. Replacing screenshot inspection with compact text residuals from geometry-aware feedback eliminates this visual token overhead and significantly shortens generation time.
\end{enumerate}

\begin{table*}[t]
  \centering
  \footnotesize
  \renewcommand{\arraystretch}{1.08}
  \setlength{\tabcolsep}{3pt}
  \caption{\textbf{Trajectory call and token attribution on Remotion comparing Agent-VLM against \textbf{EcoVid (Ours)} across \textsc{Bench60}.} Green subscripts indicate relative reductions in mechanical bookkeeping and increases in visual authoring share.}
  \label{tab:supp_trajectory}
  \resizebox{\textwidth}{!}{%
  \begin{tabular}{llcccccccc}
    \toprule
    \multirow{2}{*}{\textbf{\textit{Role}}}
    & \multirow{2}{*}{\textbf{\textit{Trajectory Action Category}}}
    & \multicolumn{4}{c}{\textbf{\textit{Agent-VLM (Native VLM Baseline)}}}
    & \multicolumn{4}{c}{\cellcolor{oursrow}\textbf{\textit{EcoVid (Ours)}}} \\
    \cmidrule(lr){3-6} \cmidrule(lr){7-10}
    & & \textbf{\textit{Calls}} & \textbf{\textit{Call \%}} & \textbf{\textit{Tokens}} & \textbf{\textit{Token \%}}
      & \cellcolor{oursrow}\textbf{\textit{Calls}} & \cellcolor{oursrow}\textbf{\textit{Call \%}} & \cellcolor{oursrow}\textbf{\textit{Tokens}} & \cellcolor{oursrow}\textbf{\textit{Token \%}} \\
    \midrule
    \multirow{5}{*}{\makecell[l]{\textbf{Mechanical}\\\textbf{Bookkeeping}}}
    & Pixel-Level VLM Screenshot Inspection
      & 7.10 & 23.9\% & 258.2K & 32.0\%
      & \cellcolor{oursrow}\textbf{0.00}\downgood{100\%} & \cellcolor{oursrow}\textbf{0.0\%} & \cellcolor{oursrow}\textbf{0.0K}\downgood{100\%} & \cellcolor{oursrow}\textbf{0.0\%} \\
    & Frame Probing \& Layout Verification
      & 6.32 & 21.3\% & 184.7K & 22.9\%
      & \cellcolor{oursrow}\textbf{3.48}\downgood{44.9\%} & \cellcolor{oursrow}26.2\% & \cellcolor{oursrow}\textbf{63.7K}\downgood{65.5\%} & \cellcolor{oursrow}26.2\% \\
    & Manual Timestamp Parsing \& Frame-Math Scripting
      & 4.75 & 16.0\% & 46.7K & 5.8\%
      & \cellcolor{oursrow}\textbf{0.00}\downgood{100\%} & \cellcolor{oursrow}\textbf{0.0\%} & \cellcolor{oursrow}\textbf{0.0K}\downgood{100\%} & \cellcolor{oursrow}\textbf{0.0\%} \\
    & Intermediate Rendering, Speech \& File Exploration
      & 4.23 & 14.2\% & 114.7K & 14.2\%
      & \cellcolor{oursrow}\textbf{1.32}\downgood{68.8\%} & \cellcolor{oursrow}9.9\% & \cellcolor{oursrow}\textbf{31.0K}\downgood{73.0\%} & \cellcolor{oursrow}12.8\% \\
    \cmidrule(lr){2-10}
    & \cellcolor{secgray}\textbf{\textit{Subtotal: Mechanical Bookkeeping}}
      & \cellcolor{secgray}\textbf{22.40} & \cellcolor{secgray}\textbf{75.4\%} & \cellcolor{secgray}\textbf{604.3K} & \cellcolor{secgray}\textbf{75.0\%}
      & \cellcolor{oursrow}\textbf{4.80}\downgood{78.6\%} & \cellcolor{oursrow}\textbf{36.1\%} & \cellcolor{oursrow}\textbf{94.7K}\downgood{84.3\%} & \cellcolor{oursrow}\textbf{39.0\%} \\
    \midrule
    \multirow{3}{*}{\makecell[l]{\textbf{High-Level}\\\textbf{Visual Authoring}}}
    & Animation Code Generation \& Refinement
      & 4.55 & 15.3\% & 148.9K & 18.5\%
      & \cellcolor{oursrow}\textbf{6.20} & \cellcolor{oursrow}\textbf{46.6\%}\upgood{31.3\%} & \cellcolor{oursrow}\textbf{118.8K} & \cellcolor{oursrow}\textbf{48.9\%}\upgood{30.4\%} \\
    & Pedagogical Narration Scripting \& Summary
      & 2.75 & 9.3\% & 52.8K & 6.5\%
      & \cellcolor{oursrow}\textbf{2.30} & \cellcolor{oursrow}\textbf{17.3\%}\upgood{8.0\%} & \cellcolor{oursrow}\textbf{29.5K} & \cellcolor{oursrow}\textbf{12.1\%}\upgood{5.6\%} \\
    \cmidrule(lr){2-10}
    & \cellcolor{secgray}\textbf{\textit{Subtotal: High-Level Visual Authoring}}
      & \cellcolor{secgray}\textbf{7.30} & \cellcolor{secgray}\textbf{24.6\%} & \cellcolor{secgray}\textbf{201.7K} & \cellcolor{secgray}\textbf{25.0\%}
      & \cellcolor{oursrow}\textbf{8.50} & \cellcolor{oursrow}\textbf{63.9\%}\upgood{39.3\%} & \cellcolor{oursrow}\textbf{148.3K} & \cellcolor{oursrow}\textbf{61.0\%}\upgood{36.0\%} \\
    \midrule
    \multicolumn{2}{l}{\textbf{Full Trajectory Total (Per Video Mean)}}
      & \textbf{29.70} & \textbf{100.0\%} & \textbf{806.0K} & \textbf{100.0\%}
      & \cellcolor{oursrow}\textbf{13.30}\downgood{55.2\%} & \cellcolor{oursrow}\textbf{100.0\%} & \cellcolor{oursrow}\textbf{243.0K}\downgood{69.9\%} & \cellcolor{oursrow}\textbf{100.0\%} \\
    \bottomrule
  \end{tabular}%
  }
\end{table*}

\paragraph{Trajectory Attribution Between Mechanical Bookkeeping and Visual Authoring.}
To examine how iterative coding agents allocate their interaction budget, \Cref{tab:supp_trajectory} categorizes all model calls and tokens across \textsc{Bench60} on Remotion into mechanical bookkeeping and high-level visual authoring.
In the Agent-VLM baseline, three-quarters of both model calls and total tokens are spent on mechanical bookkeeping—primarily screenshot inspection, ad-hoc frame probing, intermediate rendering, and manual timestamp arithmetic—leaving only a small fraction of the budget for writing or refining animation code.
With \textbf{EcoVid}, symbolic TTS timing binding and geometry-aware feedback eliminate standalone screenshot inspection and manual timestamp calculations altogether. As a result, mechanical bookkeeping tokens drop by over $84\%$, allowing the majority of model calls and tokens to focus directly on high-level visual authoring and narration design.

\section{Domain Sensitivity and Bilingual Evaluation}
\label{sec:supp_categories}

\begin{table*}[t]
  \centering
  \scriptsize
  \renewcommand{\arraystretch}{0.96}
  \setlength{\tabcolsep}{5.4pt}
  \caption{\textbf{Per-category performance breakdown across the six subject domains on \textsc{Bench60}.}}
  \label{tab:supp_categories}
  \begin{tabular}{lcccccccccc}
    \toprule
    \multirow{2}{*}{\textbf{\textit{Subject Category}}}
    & \multicolumn{5}{c}{\textbf{\textit{Quality, Synchronization \& Pedagogical Evaluation}}}
    & \multicolumn{5}{c}{\textbf{\textit{Sync Diagnostics \& Cost Evaluation}}} \\
    \cmidrule(lr){2-6} \cmidrule(lr){7-11}
    & \textbf{\textit{Def.$\downarrow$}} & \textbf{\textit{VS$\uparrow$}} & \textbf{\textit{Vis-Rel$\uparrow$}} & \textbf{\textit{M-KC$\uparrow$}} & \textbf{\textit{AES$\uparrow$}}
    & \textbf{\textit{Lag$\downarrow$}} & \textbf{\textit{Lead$\downarrow$}} & \textbf{\textit{Tokens$\downarrow$}} & \textbf{\textit{Time (m)$\downarrow$}} & \textbf{\textit{Len.\ (s)}} \\
    \midrule
    \rowcolor{oursrow}\multicolumn{11}{c}{\textbf{EcoVid on Remotion (Ours)}} \\
    \midrule
    Science \& Computing & 3.4\% & 0.778 & 4.75 & 0.837 & 96.2 & 10.6\% & 10.0\% & \textbf{182.4K} & \textbf{9.7} & 59.7 \\
    Everyday Life & 11.1\% & 0.848 & 4.60 & 0.805 & 95.6 & \textbf{5.1\%} & 6.3\% & 299.3K & 12.3 & 58.6 \\
    Astronomy & 6.7\% & 0.755 & 4.64 & 0.831 & 97.3 & 12.0\% & 8.1\% & 260.6K & 12.2 & 60.5 \\
    Geography & 12.5\% & \textbf{0.861} & 4.51 & 0.796 & 94.4 & 6.3\% & \textbf{2.5\%} & 313.9K & 14.6 & 62.2 \\
    Health \& Physiology & \textbf{0.0\%} & 0.847 & \textbf{4.81} & \textbf{0.856} & \textbf{100.0} & 7.0\% & 8.4\% & 268.3K & 12.0 & 56.5 \\
    Sports Rules & 4.7\% & 0.828 & 4.72 & 0.773 & 98.1 & 9.4\% & 3.2\% & 221.1K & 11.6 & 61.1 \\
    \midrule
    \rowcolor{secgray}\multicolumn{11}{c}{\textit{DataMagic~\cite{xie2026datamagic} on Remotion}} \\
    \midrule
    Science \& Computing & 13.7\% & \textbf{0.697} & 4.29 & 0.772 & 94.2 & 17.8\% & \textbf{7.5\%} & 658.3K & 13.0 & 55.3 \\
    Everyday Life & 12.5\% & 0.613 & 4.41 & 0.792 & 94.4 & \textbf{5.6\%} & 20.4\% & 631.6K & 12.7 & 53.9 \\
    Astronomy & 15.1\% & 0.564 & 4.29 & 0.775 & 94.7 & 20.7\% & 8.7\% & 688.3K & 14.4 & 54.0 \\
    Geography & 2.3\% & 0.494 & \textbf{4.81} & \textbf{0.836} & 98.8 & 24.3\% & 14.7\% & 851.8K & 15.2 & 56.8 \\
    Health \& Physiology & 11.2\% & 0.681 & 4.42 & 0.801 & 94.9 & 9.5\% & 13.1\% & \textbf{501.2K} & 11.9 & 53.9 \\
    Sports Rules & \textbf{0.0\%} & 0.599 & 4.80 & 0.814 & \textbf{99.9} & 15.4\% & 14.8\% & 514.1K & \textbf{11.6} & 54.7 \\
    \midrule
    \rowcolor{secgray}\multicolumn{11}{c}{\textit{Reels-RSI~\cite{lin2026reelsrsi} on HyperFrames}} \\
    \midrule
    Science \& Computing & 30.4\% & 0.679 & \textbf{3.40} & 0.711 & 87.6 & 16.0\% & 8.9\% & 1,290.9K & 26.6 & 53.9 \\
    Everyday Life & 34.6\% & 0.728 & 3.06 & \textbf{0.742} & 86.1 & 13.9\% & 6.9\% & \textbf{887.3K} & \textbf{20.0} & 53.6 \\
    Astronomy & 46.2\% & 0.639 & 2.81 & 0.691 & 81.3 & 17.8\% & 5.3\% & 2,076.2K & 29.1 & 53.8 \\
    Geography & 34.2\% & 0.669 & 3.26 & 0.712 & 85.0 & 22.1\% & 4.6\% & 3,630.1K & 34.5 & 54.8 \\
    Health \& Physiology & 46.3\% & \textbf{0.784} & 3.27 & 0.608 & 81.5 & 14.2\% & \textbf{2.5\%} & 1,088.0K & 23.9 & 52.1 \\
    Sports Rules & \textbf{25.9\%} & 0.648 & 3.39 & 0.692 & \textbf{88.6} & \textbf{13.0\%} & 13.4\% & 1,031.5K & 25.6 & 53.1 \\
    \midrule
    \rowcolor{secgray}\multicolumn{11}{c}{\textit{TheoremExplainAgent~\cite{ku2025theoremexplainagent} on Manim}} \\
    \midrule
    Science \& Computing & 9.5\% & 0.605 & 4.04 & 0.721 & 93.2 & 20.1\% & 8.7\% & \textbf{114.8K} & \textbf{13.7} & 43.6 \\
    Everyday Life & 14.5\% & 0.541 & 3.83 & 0.687 & 92.9 & 18.4\% & 15.7\% & 121.0K & 14.8 & 42.0 \\
    Astronomy & \textbf{3.1\%} & 0.504 & \textbf{4.34} & 0.707 & \textbf{97.7} & 20.4\% & 13.8\% & 132.1K & 16.2 & 43.9 \\
    Geography & 15.1\% & 0.485 & 3.85 & 0.666 & 92.0 & 17.6\% & 17.1\% & 135.1K & 16.0 & 42.1 \\
    Health \& Physiology & 4.0\% & \textbf{0.722} & 4.22 & \textbf{0.767} & 96.6 & \textbf{9.3\%} & \textbf{6.4\%} & 133.8K & 15.8 & 42.6 \\
    Sports Rules & 8.0\% & 0.532 & 4.24 & 0.727 & 95.3 & 17.0\% & 14.0\% & 133.5K & 16.2 & 38.6 \\
    \midrule
    \rowcolor{secgray}\multicolumn{11}{c}{\textit{video-shotcraft~\cite{wei2026shotcraft} on Remotion}} \\
    \midrule
    Science \& Computing & 40.3\% & 0.789 & \textbf{3.55} & 0.635 & 83.3 & \textbf{6.4\%} & 8.6\% & \textbf{10,077.7K} & \textbf{75.0} & 56.8 \\
    Everyday Life & 39.7\% & 0.689 & 3.44 & 0.673 & 82.6 & 13.4\% & 11.2\% & 12,902.4K & 89.7 & 56.0 \\
    Astronomy & 72.2\% & 0.649 & 3.11 & 0.462 & 69.5 & 12.1\% & 11.4\% & 14,311.9K & 96.3 & 55.7 \\
    Geography & 56.0\% & \textbf{0.801} & \textbf{3.55} & 0.571 & 76.6 & 9.3\% & \textbf{5.6\%} & 15,417.0K & 95.6 & 58.1 \\
    Health \& Physiology & \textbf{33.1\%} & 0.731 & 3.29 & \textbf{0.702} & \textbf{85.2} & 12.9\% & 9.2\% & 12,908.2K & 85.3 & 58.6 \\
    Sports Rules & 42.7\% & 0.764 & 3.23 & 0.589 & 81.1 & 9.6\% & 12.2\% & 14,320.5K & 102.1 & 58.1 \\
    \midrule
    \rowcolor{secgray}\multicolumn{11}{c}{\textit{Agent-VLM on Remotion}} \\
    \midrule
    Science \& Computing & \textbf{15.3\%} & 0.763 & 4.45 & 0.725 & 90.5 & 13.5\% & \textbf{3.9\%} & \textbf{704.8K} & 13.3 & 56.5 \\
    Everyday Life & 40.7\% & \textbf{0.769} & 3.74 & 0.685 & 86.5 & 13.2\% & 5.0\% & 840.3K & 15.6 & 55.2 \\
    Astronomy & 19.5\% & 0.686 & \textbf{4.79} & \textbf{0.838} & \textbf{100.0} & \textbf{9.9\%} & 10.9\% & 726.9K & 15.4 & 57.7 \\
    Geography & 50.8\% & 0.684 & 3.54 & 0.534 & 78.7 & 18.5\% & 6.4\% & 859.1K & \textbf{13.2} & 57.0 \\
    Health \& Physiology & 32.9\% & 0.738 & 3.96 & 0.563 & 85.2 & 10.2\% & 9.0\% & 850.1K & 14.1 & 56.0 \\
    Sports Rules & 33.3\% & 0.725 & 3.86 & 0.575 & 86.0 & 13.5\% & 4.1\% & 1,006.8K & 17.1 & 56.7 \\
    \midrule
    \rowcolor{secgray}\multicolumn{11}{c}{\textit{Code2Video~\cite{chen2026code2video} on Manim}} \\
    \midrule
    Science \& Computing & 13.4\% & 0.597 & 3.46 & 0.730 & 93.3 & 13.8\% & 19.1\% & 123.5K & 16.0 & 55.0 \\
    Everyday Life & 17.5\% & 0.597 & 3.00 & 0.673 & 91.6 & 10.7\% & 17.9\% & 128.3K & 16.9 & 53.7 \\
    Astronomy & 8.0\% & 0.458 & 3.52 & 0.713 & \textbf{95.5} & 13.1\% & 26.5\% & 170.0K & 17.7 & 59.1 \\
    Geography & 15.8\% & \textbf{0.711} & 3.44 & 0.661 & 93.2 & \textbf{9.6\%} & \textbf{11.7\%} & 183.0K & 16.1 & 51.7 \\
    Health \& Physiology & 19.4\% & 0.584 & 3.17 & \textbf{0.735} & 91.4 & 13.7\% & 14.5\% & 103.4K & \textbf{14.5} & 51.0 \\
    Sports Rules & \textbf{5.7\%} & 0.472 & \textbf{3.74} & 0.650 & 95.0 & 16.6\% & 25.8\% & \textbf{100.8K} & 14.9 & 55.7 \\
    \bottomrule
  \end{tabular}
\end{table*}

\begin{table}[t]
  \centering
  \scriptsize
  \renewcommand{\arraystretch}{1.04}
  \setlength{\tabcolsep}{2.2pt}
  \caption{\textbf{Bilingual performance breakdown across English and Chinese topic subsets on \textsc{Bench60}.}}
  \label{tab:supp_bilingual}
  \resizebox{\columnwidth}{!}{%
  \begin{tabular}{lllcccccc}
    \toprule
    \textbf{\textit{Carrier}} & \textbf{\textit{Method}} & \textbf{\textit{Split}} & \textbf{\textit{Tok.$\downarrow$}} & \textbf{\textit{Def.$\downarrow$}} & \textbf{\textit{VS$\uparrow$}} & \textbf{\textit{Vis-Rel$\uparrow$}} & \textbf{\textit{M-KC$\uparrow$}} & \textbf{\textit{AES$\uparrow$}} \\
    \midrule
    \multirow{8}{*}{Remotion}
    & \multirow{2}{*}{anything2explainer~\cite{wei2026anything2explainer}} & EN & 10,648K & 23.1\% & \textbf{0.844} & 3.68 & 0.774 & 90.8 \\
    & & ZH & 9,867K & 19.6\% & 0.797 & 4.04 & 0.778 & 92.2 \\
    \cmidrule(lr){2-9}
    & \multirow{2}{*}{video-shotcraft~\cite{wei2026shotcraft}} & EN & 13,554K & 50.3\% & 0.825 & 3.14 & 0.609 & 79.0 \\
    & & ZH & 11,794K & 41.4\% & 0.671 & 3.65 & 0.613 & 82.4 \\
    \cmidrule(lr){2-9}
    & \multirow{2}{*}{DataMagic~\cite{xie2026datamagic}} & EN & 608K & \textbf{3.5\%} & 0.627 & 4.55 & 0.816 & 95.4 \\
    & & ZH & 681K & 16.6\% & 0.624 & 4.37 & 0.769 & 96.2 \\
    \cmidrule(lr){2-9}
    & \cellcolor{oursrow}\textbf{EcoVid (Ours)} & \cellcolor{oursrow}EN & \cellcolor{oursrow}\textbf{258K} & \cellcolor{oursrow}5.1\% & \cellcolor{oursrow}0.807 & \cellcolor{oursrow}\textbf{4.73} & \cellcolor{oursrow}\textbf{0.829} & \cellcolor{oursrow}\textbf{96.3} \\
    & \cellcolor{oursrow}\textbf{EcoVid (Ours)} & \cellcolor{oursrow}ZH & \cellcolor{oursrow}\textbf{227K} & \cellcolor{oursrow}\textbf{6.5\%} & \cellcolor{oursrow}\textbf{0.815} & \cellcolor{oursrow}\textbf{4.64} & \cellcolor{oursrow}\textbf{0.811} & \cellcolor{oursrow}\textbf{97.2} \\
    \midrule
    \multirow{4}{*}{HyperFrames}
    & \multirow{2}{*}{Reels-RSI~\cite{lin2026reelsrsi}} & EN & 1,657K & 32.9\% & 0.630 & 3.26 & 0.731 & 86.0 \\
    & & ZH & 1,527K & 37.3\% & 0.747 & 3.23 & 0.661 & 85.1 \\
    \cmidrule(lr){2-9}
    & \cellcolor{oursrow}\textbf{EcoVid (Ours)} & \cellcolor{oursrow}EN & \cellcolor{oursrow}\textbf{249K} & \cellcolor{oursrow}\textbf{4.5\%} & \cellcolor{oursrow}\textbf{0.755} & \cellcolor{oursrow}\textbf{4.66} & \cellcolor{oursrow}\textbf{0.831} & \cellcolor{oursrow}\textbf{96.0} \\
    & \cellcolor{oursrow}\textbf{EcoVid (Ours)} & \cellcolor{oursrow}ZH & \cellcolor{oursrow}\textbf{172K} & \cellcolor{oursrow}\textbf{9.0\%} & \cellcolor{oursrow}\textbf{0.783} & \cellcolor{oursrow}\textbf{4.53} & \cellcolor{oursrow}\textbf{0.791} & \cellcolor{oursrow}\textbf{96.2} \\
    \midrule
    \multirow{6}{*}{Manim}
    & \multirow{2}{*}{TheoremExplainAgent~\cite{ku2025theoremexplainagent}} & EN & \textbf{128K} & 7.4\% & 0.515 & 4.14 & 0.720 & 93.7 \\
    & & ZH & \textbf{123K} & 10.8\% & 0.630 & 4.01 & 0.708 & 94.9 \\
    \cmidrule(lr){2-9}
    & \multirow{2}{*}{Code2Video~\cite{chen2026code2video}} & EN & 134K & 15.3\% & 0.609 & 3.29 & 0.708 & 92.8 \\
    & & ZH & 131K & 11.3\% & 0.542 & 3.52 & 0.694 & 93.9 \\
    \cmidrule(lr){2-9}
    & \cellcolor{oursrow}\textbf{EcoVid (Ours)} & \cellcolor{oursrow}EN & \cellcolor{oursrow}202K & \cellcolor{oursrow}\textbf{3.2\%} & \cellcolor{oursrow}\textbf{0.854} & \cellcolor{oursrow}\textbf{4.58} & \cellcolor{oursrow}\textbf{0.830} & \cellcolor{oursrow}\textbf{98.5} \\
    & \cellcolor{oursrow}\textbf{EcoVid (Ours)} & \cellcolor{oursrow}ZH & \cellcolor{oursrow}163K & \cellcolor{oursrow}\textbf{2.3\%} & \cellcolor{oursrow}\textbf{0.826} & \cellcolor{oursrow}\textbf{4.70} & \cellcolor{oursrow}\textbf{0.827} & \cellcolor{oursrow}\textbf{98.8} \\
    \bottomrule
  \end{tabular}%
  }
\end{table}

\paragraph{Performance Across Subject Domains.}
Comparing performance across the six subject domains of \textsc{Bench60} in \Cref{tab:supp_categories} shows that different baseline paradigms struggle with distinct types of instructional content:
\begin{itemize}
  \item \textbf{Diagram-Dense Domains:}
  Subjects such as geography, astronomy, and everyday life mechanisms require multi-layered spatial diagrams with directional arrows, region labels, and callout boxes. On these visually dense topics, methods that rely on VLM screenshot inspection exhibit high layout defect rates, because vision models struggle to infer precise coordinate adjustments when multiple labels cluster near diagram boundaries. By querying exact element bounding boxes from the structured scene layout, \textbf{EcoVid} resolves spatial collisions reliably across complex schematics.

  \item \textbf{Derivation- and Rule-Intensive Domains:}
  Conversely, declarative compilation pipelines maintain relatively clean static layouts on structured card layouts, but suffer substantial timing lag and lead errors on mathematical derivations and multi-step sports rules. In these domains, spoken sentences vary widely in length—from short symbolic identities to multi-clause explanations—causing uniform or heuristic scene duration estimates to drift out of sync with the voiceover. Binding every sentence directly to its measured speech interval enables \textbf{EcoVid} to maintain consistent audio-visual synchronization across all six domains.
\end{itemize}

\paragraph{Cross-Lingual Consistency Across English and Chinese.}
Comparing the English and Simplified Chinese subsets in \Cref{tab:supp_bilingual} reveals two common cross-lingual challenges in prior systems.
First, Chinese narration conveys different information density per second than English; pipelines calibrated for English speaking rates often misestimate Chinese beat durations and suffer reduced visual synchronization.
Second, Chinese characters have uniform square aspect ratios and require larger minimum font sizes for legibility than Latin text, causing layout templates tuned on English strings to overflow their containers when rendering Chinese labels.
Because \textbf{EcoVid} adapts script length to language-specific speaking rates, binds animations to actual synthesized speech timestamps, and verifies rendered text bounding boxes directly in the target engine, it achieves consistent synchronization and low defect rates across both languages.

\begin{figure*}[t]
  \centering
  \includegraphics[width=0.95\linewidth,keepaspectratio]{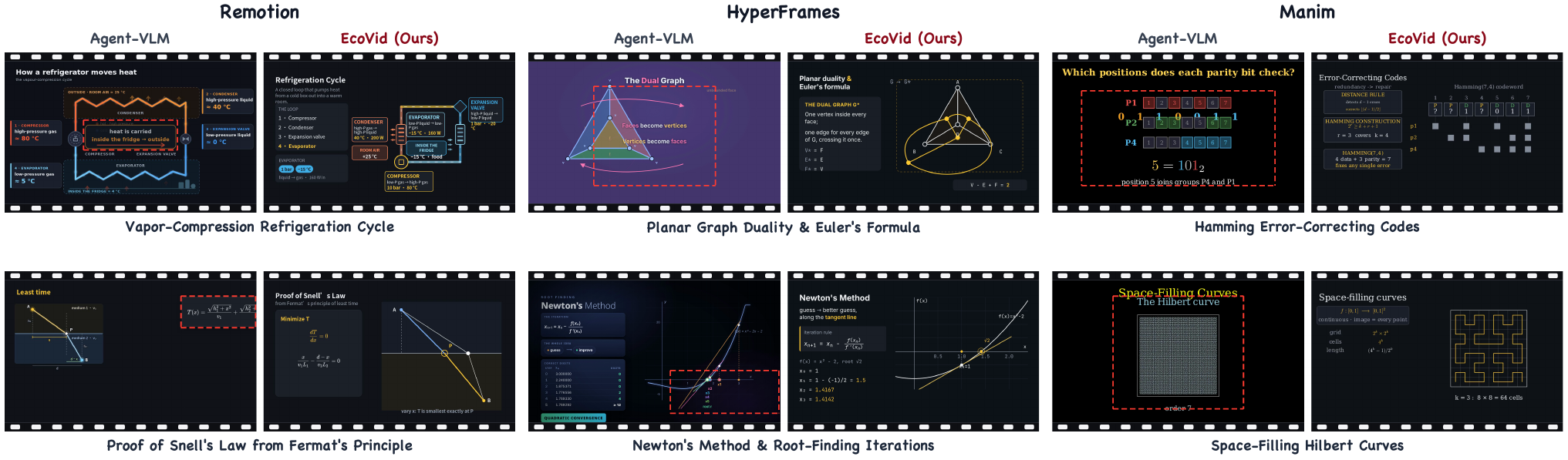}
  \caption{\textbf{Controlled $3 \times 2$ rest-frame comparison between Agent-VLM and EcoVid under identical visual themes across Remotion, HyperFrames, and Manim.}
  Across all six benchmark topics, screenshot-based inspection in Agent-VLM (left of each pair; red dashed boxes mark layout defects) leaves text overlaps, badge overflows, and occluded equations, whereas \textbf{EcoVid} (right of each pair) resolves spatial defects via geometry-aware feedback.}
  \label{fig:suppl_agentvlm_comp}
\end{figure*}

\section{Model-Agnostic Generalization Across Open and Closed Backbones}
\label{sec:supp_small_model}

\begin{table*}[t]
  \centering
  \scriptsize
  \renewcommand{\arraystretch}{1.02}
  \setlength{\tabcolsep}{2.6pt}
  \caption{\textbf{Model-agnostic evaluation across open and closed backbones on Remotion, HyperFrames, and Manim.} Green subscripts denote improvements of \textbf{EcoVid} over Agent-VLM under the same backbone.}
  \label{tab:supp_model_agnostic}
  \label{tab:supp_qwen27b}
  \resizebox{\textwidth}{!}{%
  \begin{tabular}{lllllccccc|ccccc}
    \toprule
    \multirow{2}{*}{\textbf{\textit{Carrier}}}
    & \multirow{2}{*}{\textbf{\textit{Backbone Model}}}
    & \multirow{2}{*}{\textbf{\textit{Category}}}
    & \multirow{2}{*}{\textbf{\textit{Thinking}}}
    & \multirow{2}{*}{\textbf{\textit{Method}}}
    & \multicolumn{5}{c|}{\textbf{\textit{Efficiency \& Prefix KV-Cache Breakdown}}}
    & \multicolumn{5}{c}{\textbf{\textit{Video Quality, Synchronization \& Pedagogy}}} \\
    \cmidrule(lr){6-10} \cmidrule(lr){11-15}
    & & & & & \textbf{\textit{Calls$\downarrow$}} & \textbf{\textit{Tokens$\downarrow$}} & \textbf{\textit{Uncach.$\downarrow$}} & \textbf{\textit{Hit\%$\uparrow$}} & \textbf{\textit{Time (m)$\downarrow$}}
    & \textbf{\textit{Def.$\downarrow$}} & \textbf{\textit{VS$\uparrow$}} & \textbf{\textit{Vis-Rel$\uparrow$}} & \textbf{\textit{M-KC$\uparrow$}} & \textbf{\textit{AES$\uparrow$}} \\
    \midrule
    \multirow{16}{*}{Remotion}
    & \multirow{2}{*}{Qwen3.8-27B}
    & \multirow{2}{*}{Open}
    & \multirow{2}{*}{xhigh}
    & Agent-VLM & 18.6 & 214.5K & 31.2K & \textbf{83.6\%} & 8.9 & 29.0\% & 0.687 & 3.88 & 0.642 & 85.6 \\
    & & & & \cellcolor{oursrow}\textbf{EcoVid (Ours)} & \cellcolor{oursrow}\textbf{10.4}\downgood{8.2} & \cellcolor{oursrow}\textbf{118.2K}\downgood{96.3K} & \cellcolor{oursrow}\textbf{24.8K}\downgood{6.4K} & \cellcolor{oursrow}74.5\% & \cellcolor{oursrow}\textbf{7.4}\downgood{1.5} & \cellcolor{oursrow}\textbf{6.9\%}\downgood{22.1} & \cellcolor{oursrow}\textbf{0.779}\upgood{0.092} & \cellcolor{oursrow}\textbf{4.54}\upgood{0.66} & \cellcolor{oursrow}\textbf{0.798}\upgood{0.156} & \cellcolor{oursrow}\textbf{94.9}\upgood{9.3} \\
    \cmidrule(lr){2-15}
    & \multirow{2}{*}{GLM-5.3}
    & \multirow{2}{*}{Open}
    & \multirow{2}{*}{medium}
    & Agent-VLM & 26.2 & 266.4K & 50.6K & \textbf{80.1\%} & 8.2 & 30.4\% & 0.697 & 4.07 & 0.686 & 87.6 \\
    & & & & \cellcolor{oursrow}\textbf{EcoVid (Ours)} & \cellcolor{oursrow}\textbf{13.4}\downgood{12.8} & \cellcolor{oursrow}\textbf{154.3K}\downgood{112.1K} & \cellcolor{oursrow}\textbf{39.3K}\downgood{11.3K} & \cellcolor{oursrow}72.2\% & \cellcolor{oursrow}\textbf{7.2}\downgood{1.0} & \cellcolor{oursrow}\textbf{6.1\%}\downgood{24.3} & \cellcolor{oursrow}\textbf{0.806}\upgood{0.109} & \cellcolor{oursrow}\textbf{4.66}\upgood{0.59} & \cellcolor{oursrow}\textbf{0.816}\upgood{0.130} & \cellcolor{oursrow}\textbf{96.8}\upgood{9.2} \\
    \cmidrule(lr){2-15}
    & \multirow{2}{*}{DeepSeek-V4.1}
    & \multirow{2}{*}{Open}
    & \multirow{2}{*}{medium}
    & Agent-VLM & 33.1 & 1,629.3K & 83.5K & \textbf{94.7\%} & 18.7 & 26.8\% & 0.729 & 4.19 & 0.711 & 89.9 \\
    & & & & \cellcolor{oursrow}\textbf{EcoVid (Ours)} & \cellcolor{oursrow}\textbf{7.2}\downgood{25.9} & \cellcolor{oursrow}\textbf{210.1K}\downgood{1,419.2K} & \cellcolor{oursrow}\textbf{43.1K}\downgood{40.4K} & \cellcolor{oursrow}75.4\% & \cellcolor{oursrow}\textbf{6.4}\downgood{12.3} & \cellcolor{oursrow}\textbf{5.2\%}\downgood{21.6} & \cellcolor{oursrow}\textbf{0.813}\upgood{0.084} & \cellcolor{oursrow}\textbf{4.70}\upgood{0.51} & \cellcolor{oursrow}\textbf{0.819}\upgood{0.108} & \cellcolor{oursrow}\textbf{97.1}\upgood{7.2} \\
    \cmidrule(lr){2-15}
    & \multirow{2}{*}{Kimi-K3.1-Pro}
    & \multirow{2}{*}{Open}
    & \multirow{2}{*}{medium}
    & Agent-VLM & 21.1 & 305.2K & 108.9K & \textbf{60.0\%} & 14.2 & 48.2\% & 0.615 & 4.00 & 0.562 & 81.9 \\
    & & & & \cellcolor{oursrow}\textbf{EcoVid (Ours)} & \cellcolor{oursrow}\textbf{9.5}\downgood{11.6} & \cellcolor{oursrow}\textbf{127.4K}\downgood{177.8K} & \cellcolor{oursrow}\textbf{64.2K}\downgood{44.7K} & \cellcolor{oursrow}41.6\% & \cellcolor{oursrow}\textbf{10.6}\downgood{3.6} & \cellcolor{oursrow}\textbf{0.9\%}\downgood{47.3} & \cellcolor{oursrow}\textbf{0.866}\upgood{0.251} & \cellcolor{oursrow}\textbf{4.89}\upgood{0.89} & \cellcolor{oursrow}\textbf{0.837}\upgood{0.275} & \cellcolor{oursrow}\textbf{98.9}\upgood{17.0} \\
    \cmidrule(lr){2-15}
    & \multirow{2}{*}{Qwen3.8-Max}
    & \multirow{2}{*}{Closed}
    & \multirow{2}{*}{medium}
    & Agent-VLM & 29.7 & 806.0K & 73.0K & \textbf{90.7\%} & 14.4 & 28.7\% & 0.735 & 4.13 & 0.668 & 88.4 \\
    & & & & \cellcolor{oursrow}\textbf{EcoVid (Ours)} & \cellcolor{oursrow}\textbf{13.3}\downgood{16.4} & \cellcolor{oursrow}\textbf{243.0K}\downgood{563.0K} & \cellcolor{oursrow}\textbf{56.0K}\downgood{17.0K} & \cellcolor{oursrow}74.1\% & \cellcolor{oursrow}\textbf{11.6}\downgood{2.8} & \cellcolor{oursrow}\textbf{5.8\%}\downgood{22.9} & \cellcolor{oursrow}\textbf{0.811}\upgood{0.076} & \cellcolor{oursrow}\textbf{4.69}\upgood{0.56} & \cellcolor{oursrow}\textbf{0.820}\upgood{0.152} & \cellcolor{oursrow}\textbf{96.8}\upgood{8.4} \\
    \cmidrule(lr){2-15}
    & \multirow{2}{*}{Gemini-4-Argon}
    & \multirow{2}{*}{Closed}
    & \multirow{2}{*}{medium}
    & Agent-VLM & 12.7 & 126.6K & 76.1K & \textbf{32.6\%} & 10.6 & 28.7\% & 0.736 & 4.13 & 0.707 & 89.4 \\
    & & & & \cellcolor{oursrow}\textbf{EcoVid (Ours)} & \cellcolor{oursrow}\textbf{4.1}\downgood{8.6} & \cellcolor{oursrow}\textbf{57.8K}\downgood{68.8K} & \cellcolor{oursrow}\textbf{31.9K}\downgood{44.2K} & \cellcolor{oursrow}24.6\% & \cellcolor{oursrow}\textbf{6.9}\downgood{3.7} & \cellcolor{oursrow}\textbf{5.2\%}\downgood{23.5} & \cellcolor{oursrow}\textbf{0.812}\upgood{0.076} & \cellcolor{oursrow}\textbf{4.70}\upgood{0.57} & \cellcolor{oursrow}\textbf{0.820}\upgood{0.113} & \cellcolor{oursrow}\textbf{97.1}\upgood{7.7} \\
    \cmidrule(lr){2-15}
    & \multirow{2}{*}{GPT-5.6}
    & \multirow{2}{*}{Closed}
    & \multirow{2}{*}{medium}
    & Agent-VLM & 21.6 & 245.0K & 121.0K & 49.1\% & 9.8 & 28.9\% & 0.686 & 4.11 & 0.677 & 87.1 \\
    & & & & \cellcolor{oursrow}\textbf{EcoVid (Ours)} & \cellcolor{oursrow}\textbf{10.3}\downgood{11.3} & \cellcolor{oursrow}\textbf{81.5K}\downgood{163.5K} & \cellcolor{oursrow}\textbf{31.3K}\downgood{89.7K} & \cellcolor{oursrow}\textbf{58.3\%}\upgood{9.2} & \cellcolor{oursrow}\textbf{3.5}\downgood{6.3} & \cellcolor{oursrow}\textbf{6.3\%}\downgood{22.6} & \cellcolor{oursrow}\textbf{0.815}\upgood{0.129} & \cellcolor{oursrow}\textbf{4.66}\upgood{0.55} & \cellcolor{oursrow}\textbf{0.818}\upgood{0.141} & \cellcolor{oursrow}\textbf{96.2}\upgood{9.1} \\
    \cmidrule(lr){2-15}
    & \multirow{2}{*}{Claude-Opus-5.5}
    & \multirow{2}{*}{Closed}
    & \multirow{2}{*}{medium}
    & Agent-VLM & 20.8 & 790.4K & 93.5K & \textbf{74.7\%} & 14.8 & 22.3\% & 0.688 & 4.19 & 0.662 & 88.2 \\
    & & & & \cellcolor{oursrow}\textbf{EcoVid (Ours)} & \cellcolor{oursrow}\textbf{5.4}\downgood{15.4} & \cellcolor{oursrow}\textbf{107.1K}\downgood{683.3K} & \cellcolor{oursrow}\textbf{42.4K}\downgood{51.1K} & \cellcolor{oursrow}46.4\% & \cellcolor{oursrow}\textbf{7.3}\downgood{7.5} & \cellcolor{oursrow}\textbf{0.2\%}\downgood{22.1} & \cellcolor{oursrow}\textbf{0.788}\upgood{0.100} & \cellcolor{oursrow}\textbf{4.93}\upgood{0.74} & \cellcolor{oursrow}\textbf{0.841}\upgood{0.179} & \cellcolor{oursrow}\textbf{99.2}\upgood{11.0} \\
    \midrule
    \multirow{4}{*}{HyperFrames}
    & \multirow{2}{*}{Qwen3.8-27B}
    & \multirow{2}{*}{Open}
    & \multirow{2}{*}{xhigh}
    & Agent-VLM & 17.8 & 226.4K & 33.4K & \textbf{83.3\%} & 9.6 & 22.4\% & 0.694 & 3.88 & 0.685 & 88.4 \\
    & & & & \cellcolor{oursrow}\textbf{EcoVid (Ours)} & \cellcolor{oursrow}\textbf{9.8}\downgood{8.0} & \cellcolor{oursrow}\textbf{109.6K}\downgood{116.8K} & \cellcolor{oursrow}\textbf{23.2K}\downgood{10.2K} & \cellcolor{oursrow}74.4\% & \cellcolor{oursrow}\textbf{7.1}\downgood{2.5} & \cellcolor{oursrow}\textbf{7.8\%}\downgood{14.6} & \cellcolor{oursrow}\textbf{0.748}\upgood{0.054} & \cellcolor{oursrow}\textbf{4.44}\upgood{0.56} & \cellcolor{oursrow}\textbf{0.790}\upgood{0.105} & \cellcolor{oursrow}\textbf{94.1}\upgood{5.7} \\
    \cmidrule(lr){2-15}
    & \multirow{2}{*}{Qwen3.8-Max}
    & \multirow{2}{*}{Closed}
    & \multirow{2}{*}{medium}
    & Agent-VLM & 30.4 & 1,146.0K & 98.0K & \textbf{91.2\%} & 16.5 & 18.2\% & 0.746 & 4.20 & 0.723 & 91.8 \\
    & & & & \cellcolor{oursrow}\textbf{EcoVid (Ours)} & \cellcolor{oursrow}\textbf{11.4}\downgood{19.0} & \cellcolor{oursrow}\textbf{210.0K}\downgood{936.0K} & \cellcolor{oursrow}\textbf{50.0K}\downgood{48.0K} & \cellcolor{oursrow}73.3\% & \cellcolor{oursrow}\textbf{10.8}\downgood{5.7} & \cellcolor{oursrow}\textbf{6.8\%}\downgood{11.4} & \cellcolor{oursrow}\textbf{0.769}\upgood{0.023} & \cellcolor{oursrow}\textbf{4.60}\upgood{0.40} & \cellcolor{oursrow}\textbf{0.811}\upgood{0.088} & \cellcolor{oursrow}\textbf{96.1}\upgood{4.3} \\
    \midrule
    \multirow{4}{*}{Manim}
    & \multirow{2}{*}{Qwen3.8-27B}
    & \multirow{2}{*}{Open}
    & \multirow{2}{*}{xhigh}
    & Agent-VLM & 21.4 & 248.2K & 29.8K & \textbf{86.5\%} & 10.2 & 19.8\% & 0.692 & 3.84 & 0.672 & 89.0 \\
    & & & & \cellcolor{oursrow}\textbf{EcoVid (Ours)} & \cellcolor{oursrow}\textbf{11.2}\downgood{10.2} & \cellcolor{oursrow}\textbf{104.8K}\downgood{143.4K} & \cellcolor{oursrow}\textbf{20.6K}\downgood{9.2K} & \cellcolor{oursrow}76.3\% & \cellcolor{oursrow}\textbf{6.8}\downgood{3.4} & \cellcolor{oursrow}\textbf{3.6\%}\downgood{16.2} & \cellcolor{oursrow}\textbf{0.809}\upgood{0.117} & \cellcolor{oursrow}\textbf{4.50}\upgood{0.66} & \cellcolor{oursrow}\textbf{0.810}\upgood{0.138} & \cellcolor{oursrow}\textbf{96.8}\upgood{7.8} \\
    \cmidrule(lr){2-15}
    & \multirow{2}{*}{Qwen3.8-Max}
    & \multirow{2}{*}{Closed}
    & \multirow{2}{*}{medium}
    & Agent-VLM & 36.4 & 1,135.0K & 83.0K & \textbf{92.5\%} & 19.9 & 15.4\% & 0.746 & 4.18 & 0.704 & 92.6 \\
    & & & & \cellcolor{oursrow}\textbf{EcoVid (Ours)} & \cellcolor{oursrow}\textbf{12.8}\downgood{23.6} & \cellcolor{oursrow}\textbf{183.0K}\downgood{952.0K} & \cellcolor{oursrow}\textbf{39.0K}\downgood{44.0K} & \cellcolor{oursrow}76.0\% & \cellcolor{oursrow}\textbf{10.3}\downgood{9.6} & \cellcolor{oursrow}\textbf{2.8\%}\downgood{12.6} & \cellcolor{oursrow}\textbf{0.840}\upgood{0.094} & \cellcolor{oursrow}\textbf{4.64}\upgood{0.46} & \cellcolor{oursrow}\textbf{0.829}\upgood{0.125} & \cellcolor{oursrow}\textbf{98.6}\upgood{6.0} \\
    \bottomrule
  \end{tabular}%
  }
\end{table*}

To verify that our harness is model-agnostic rather than tailored to a single LLM family, \Cref{tab:supp_model_agnostic} evaluates \textbf{EcoVid} alongside the native Agent-VLM baseline across diverse open-weight and closed frontier backbones on Remotion, HyperFrames, and Manim. The results highlight three consistent trends:
\begin{enumerate}
  \item \textbf{Consistent Efficiency and Quality Gains Across Model Families:}
  Across both open-weight and closed frontier backbones in \Cref{tab:supp_model_agnostic}, running the model as a standard native VLM agent requires numerous interaction turns and high token consumption while still leaving frequent layout collisions on stationary frames. Equipping the same backbones with \textbf{EcoVid} consistently cuts model calls and total token usage by more than half, substantially reduces uncached input tokens, lowers static-frame defect rates to a small single-digit percentage, and improves visual synchronization, visual relevance, multimodal knowledge coverage, and aesthetic scores across every model family.

  \item \textbf{Offloading Mechanical Bookkeeping Benefits Smaller Open Models:}
  When the locally deployed open-weight model operates as a native VLM agent without our harness, it expends considerable reasoning effort computing frame offsets from word timestamps and interpreting crowded screenshots, resulting in high layout defect rates across Remotion, HyperFrames, and Manim. Once \textbf{EcoVid} offloads timing alignment to symbolic sentence anchors and spatial verification to geometry-aware feedback, the model can focus its capacity on pedagogical structure and animation design—cutting token usage roughly in half and eliminating the vast majority of layout collisions across all three rendering engines.

  \item \textbf{Open Models with EcoVid Surpass Closed Frontier Models Without the Harness:}
  Comparing open-weight models equipped with \textbf{EcoVid} against closed frontier models running as standard native VLM agents in \Cref{tab:supp_model_agnostic} reveals a striking crossover: every open-weight backbone paired with \textbf{EcoVid} achieves lower layout defect rates, higher visual synchronization, and stronger multimodal knowledge coverage and aesthetic scores than even the strongest closed frontier models without the harness, while consuming only a fraction of the tokens. This confirms that the primary bottleneck in programmatic video generation lies in mechanical bookkeeping rather than raw backbone scale.
\end{enumerate}

\begin{figure*}[t]
  \centering
  \includegraphics[width=0.95\linewidth,height=0.58\textheight,keepaspectratio]{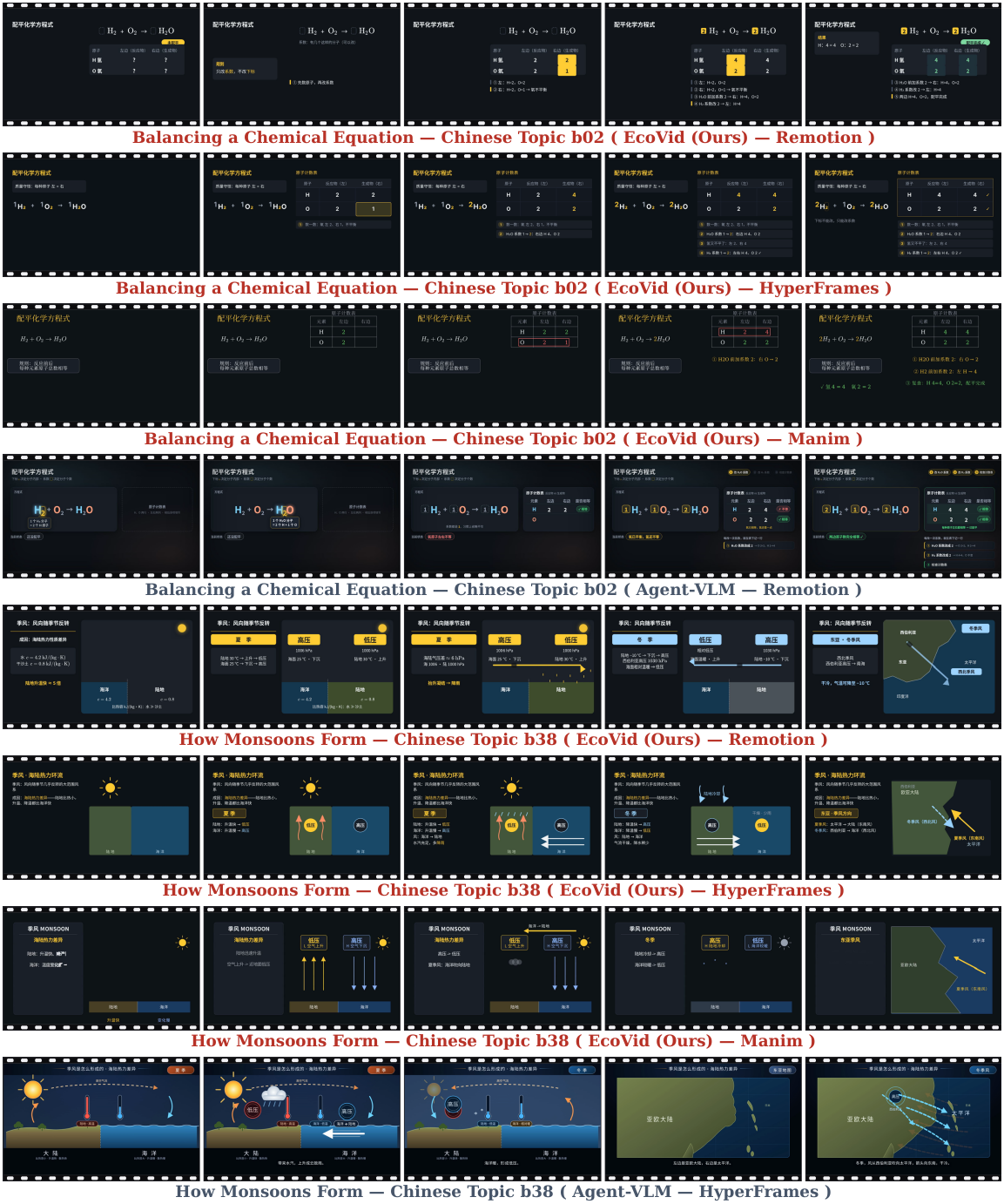}
  \caption{\textbf{Qualitative 5-frame filmstrip comparison on Simplified Chinese topics from \textsc{Bench60} (\emph{Balancing a Chemical Equation} and \emph{How Monsoons Form}).}
  \textbf{EcoVid} (red captions) renders clean Chinese typography, step-by-step stoichiometry tables, and animated meteorological diagrams across Remotion, HyperFrames, and Manim.}
  \label{fig:supp_zh}
\end{figure*}

\section{Qualitative Comparisons}
\label{sec:supp_filmstrips}

\begin{figure*}[t]
  \centering
  \includegraphics[width=0.62\linewidth]{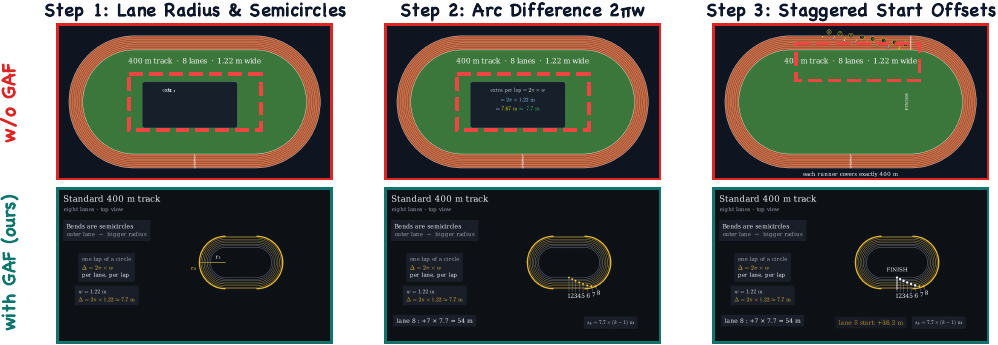}
  \caption{\textbf{Qualitative ablation of Geometry-Aware Feedback (GAF) on Manim (\emph{Why 400\,m Start Lines Are Staggered}).} Red dashed boxes mark track occlusion and lane-marker collisions without GAF.}
  \label{fig:suppl_gaf_manim}
\end{figure*}

\begin{figure*}[t]
  \centering
  \includegraphics[width=0.95\linewidth,height=0.58\textheight,keepaspectratio]{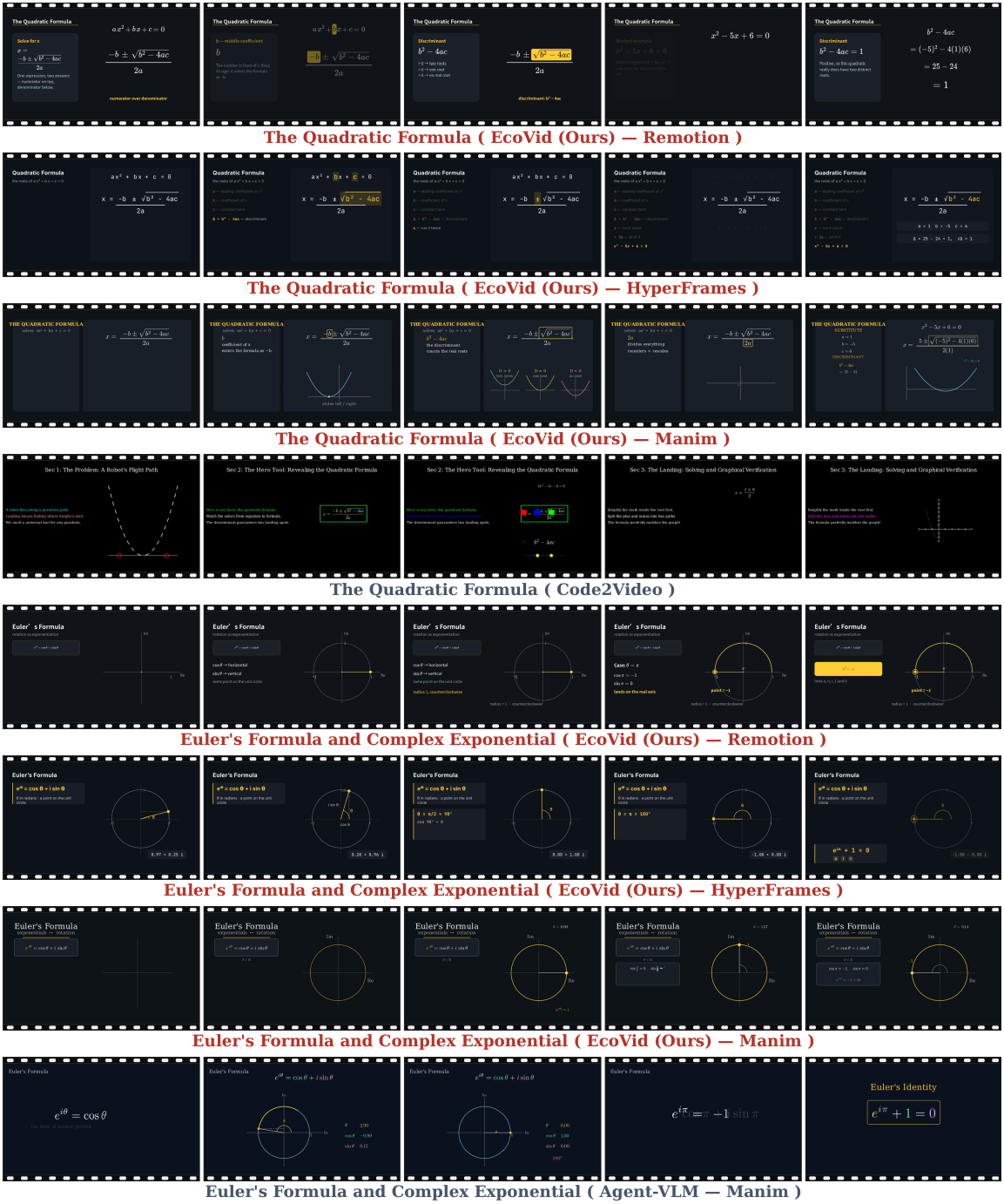}
  \caption{\textbf{Qualitative 5-frame filmstrip comparison on Mathematics and Science topics (\emph{The Quadratic Formula} and \emph{Euler's Formula $e^{\pi i} = -1$}).}
  Across all three rendering engines, \textbf{EcoVid} (red captions) synchronizes symbolic derivations with geometric visualizations.}
  \label{fig:supp_sci}
\end{figure*}

\begin{figure*}[t]
  \centering
  \includegraphics[width=0.95\linewidth,height=0.56\textheight,keepaspectratio]{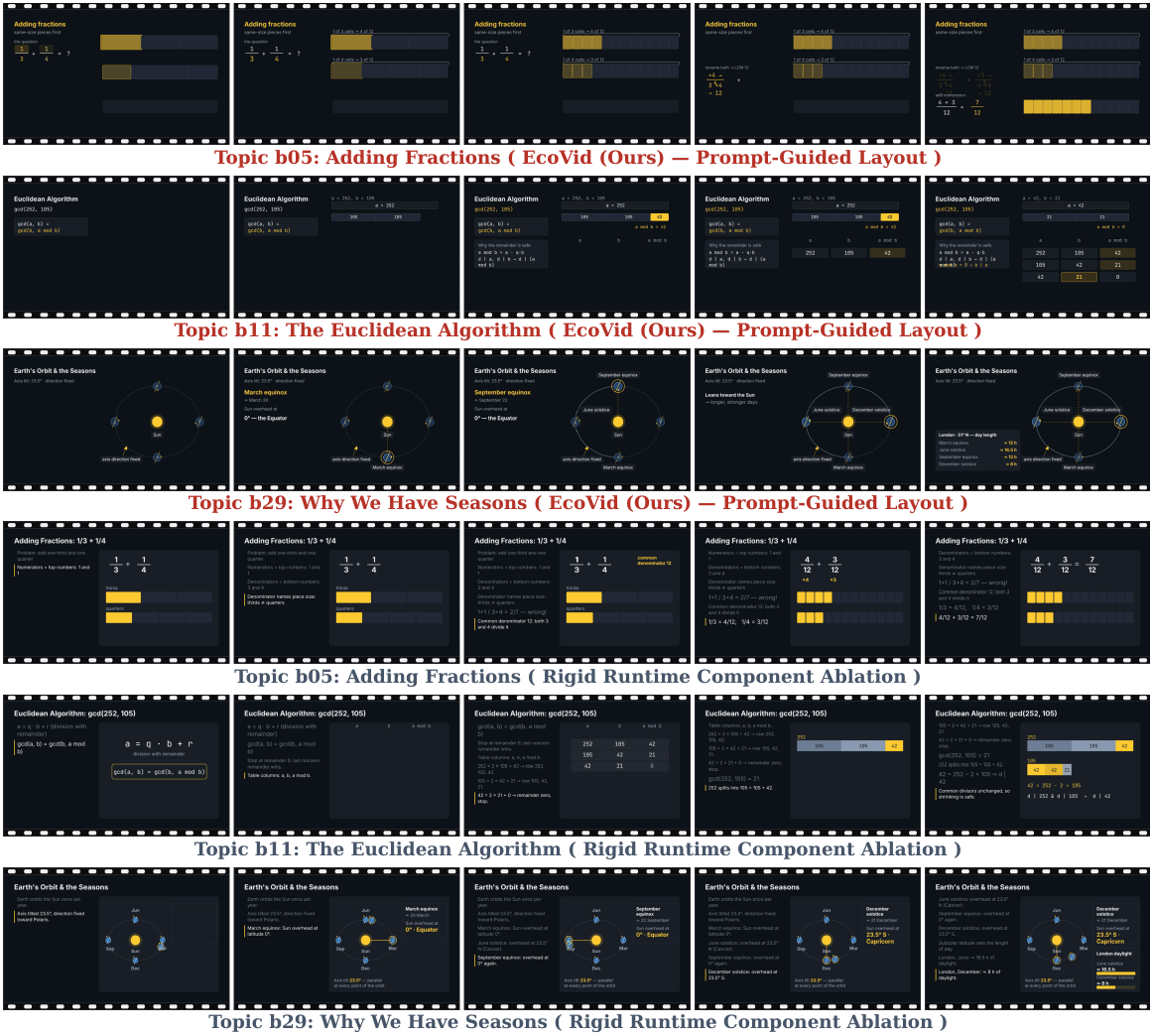}
  \caption{\textbf{Visual comparison between prompt-level layout guidance and a fixed runtime layout template across three distinct topics.}
  Top three rows (\textbf{EcoVid}, red captions): prompt-level layout guidance allows the model to tailor visual structure to each subject.
  Bottom three rows (\textbf{Fixed Runtime Layout Template}, gray captions): locking notes into a fixed template forces a uniform left-column card layout across all topics.}
  \label{fig:supp_abl}
\end{figure*}

\begin{figure*}[t]
  \centering
  \includegraphics[width=0.95\linewidth,height=0.50\textheight,keepaspectratio]{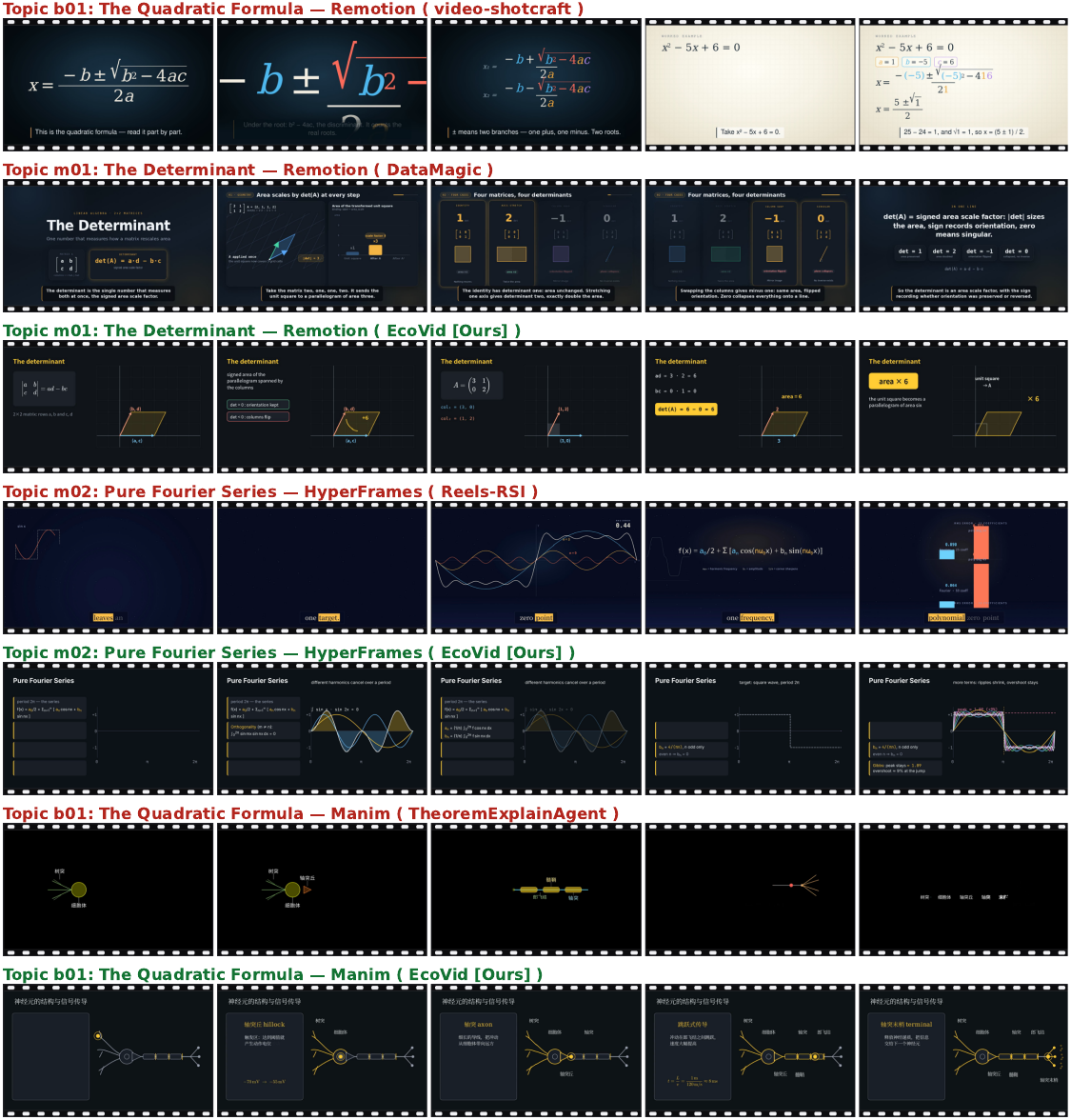}
  \caption{\textbf{Cross-engine 5-frame filmstrip comparison against community skills and academic pipelines on Remotion, HyperFrames, and Manim.}
  Across all three engines and both English and Chinese topics, \textbf{EcoVid} (red captions) produces clean layouts and synchronized step-by-step diagrams compared to the baselines (gray captions).}
  \label{fig:supp_ext}
\end{figure*}

To complement the quantitative evaluations in \Cref{sec:supp_evolution,sec:supp_token_cost,sec:supp_categories,sec:supp_small_model}, \Cref{fig:suppl_agentvlm_comp,fig:supp_zh,fig:suppl_gaf_manim,fig:supp_sci,fig:supp_abl,fig:supp_ext} present visual comparisons across rendering engines, languages, and baseline methods.

\paragraph{Comparison Against Screenshot-Based Inspection (\Cref{fig:suppl_agentvlm_comp}).}
To isolate the effect of geometry-aware feedback from visual styling, \Cref{fig:suppl_agentvlm_comp} compares Agent-VLM and \textbf{EcoVid} under an identical dark-gold color theme across six topics on Remotion, HyperFrames, and Manim.
Across all six pairs, the Agent-VLM baseline exhibits persistent layout collisions even after multiple rounds of screenshot inspection: multi-line captions overlap step badges, formula cards extend past the right canvas edge, and algebraic labels collide with coordinate axes. Because a vision model inspecting a downsampled screenshot can notice visual crowding but cannot read exact coordinate clearances, its code edits often overshoot or shift collisions to adjacent elements. By providing exact bounding-box intersection coordinates from the structured scene layout, \textbf{EcoVid} enables the coding agent to resolve these spatial defects within one to two text-only repair turns.

\paragraph{Bilingual Typography and Multi-Panel Coordination (\Cref{fig:supp_zh}).}
\Cref{fig:supp_zh} shows five-stage progressions on two Simplified Chinese topics: \emph{Balancing a Chemical Equation} and \emph{How Monsoons Form}.
On the chemistry topic, \textbf{EcoVid} coordinates left-panel conservation principles and reaction steps with right-panel atomic balance tables across all three rendering engines, whereas Agent-VLM suffers line-wrapping overflow where the equation banner overlaps the atom-counting table.
Similarly, on the meteorology topic, \textbf{EcoVid} animates land-ocean heating contrasts, pressure-gradient wind arrows, and seasonal circulation shifts in sync with the Chinese narration, whereas the Agent-VLM baseline leaves overlapping callout boxes and static arrows.

\paragraph{Qualitative Ablation of Geometry-Aware Feedback on Manim (\Cref{fig:suppl_gaf_manim}).}
Complementing the Remotion ablation in the main paper, \Cref{fig:suppl_gaf_manim} compares three aligned pedagogical steps on Manim for \emph{Why 400\,m Start Lines Are Staggered} with and without Geometry-Aware Feedback (\textbf{GAF}).
Without \textbf{GAF}, an opaque annotation box is placed directly over the center of the stadium oval in the first two steps, occluding the inner field and lane arcs, while the numbered start-line markers in the third step crowd together and overlap along the top bend.
With \textbf{GAF}, our structured scene layout verifier detects both the shape occlusion and the marker bounding-box overlaps prior to rendering, guiding the agent to separate the derivation notes on the left from the concentric track diagram and staggered lane markers on the right.

\paragraph{Synchronizing Symbolic Derivations With Geometric Diagrams (\Cref{fig:supp_sci}).}
\Cref{fig:supp_sci} illustrates multi-stage mathematical derivations on \emph{The Quadratic Formula} and \emph{Euler's Formula and Complex Exponential}.
In mathematical explainers, clear comprehension requires aligning algebraic steps with corresponding geometric updates, such as completing the square while shifting the parabola's axis of symmetry, or expanding Euler's formula while tracing the phase rotation on the complex unit circle.
Across Remotion, HyperFrames, and Manim, \textbf{EcoVid} binds each algebraic step and highlight to its measured speech interval while keeping the derivation panel on the left cleanly separated from the geometric plot on the right.
By contrast, Code2Video places centered multi-line text directly over the parabola axes without clearing earlier equations, while Agent-VLM on Manim exhibits both text-on-circle occlusion and cumulative timing lag where the final identity appears after the voiceover has ended.

\paragraph{Visual Diversity Under Prompt-Level Layout Guidance (\Cref{fig:supp_abl}).}
\Cref{fig:supp_abl} illustrates the layout homogenization effect analyzed in \Cref{tab:hf_ablations} and \cref{fig:supp_i5_bar} across \emph{Adding Fractions with Different Denominators}, \emph{The Euclidean Algorithm for GCD}, and \emph{Why We Have Seasons}.
With prompt-level layout guidance and geometry-aware feedback, the coding agent creates visual layouts tailored to each subject: horizontal fraction-bar grids for fraction addition, step-by-step division tables for the Euclidean algorithm, and orbital diagrams illustrating Earth's axial tilt and seasonal sunlight.
When constrained by a fixed runtime layout template, the agent instead defaults to placing a uniform vertical card stack on the left and a minimal static diagram on the right across all three topics, reducing visual expressiveness.

\paragraph{Qualitative Comparison With External Baselines (\Cref{fig:supp_ext}).}
\Cref{fig:supp_ext} compares external baselines against \textbf{EcoVid} across Remotion, HyperFrames, and Manim under identical backbone and voiceover settings:
\begin{itemize}
  \item \textbf{Open-Source Skills and Declarative Pipelines on Remotion:} Open-source skills often generate oversized title cards that clip past the canvas boundary and overlap algebraic formulas, while declarative compilation pipelines produce clean static cards but divide scene duration evenly across sentences of unequal length, causing geometric transformations to lag behind the voiceover. In contrast, \textbf{EcoVid} aligns each visual transformation step with its corresponding spoken sentence while keeping text and diagrams cleanly separated.
  \item \textbf{Iterative VLM-Guided Workflows on HyperFrames:} Iterative workflows that rely on screenshot-based VLM feedback frequently suffer text-on-diagram collisions where multi-line equations overlap animated waveforms, whereas \textbf{EcoVid} maintains clear spatial separation between the formula panel and the animated diagram.
  \item \textbf{Multi-Agent Scene Pipelines on Manim:} Stateless scene pipelines often produce text-heavy slides with overlapping labels and coarse scene-level timing on complex schematics, whereas \textbf{EcoVid} renders high-contrast diagrams synchronized sentence by sentence with the narration.
\end{itemize}

\clearpage\twocolumn
\section{System Prompts and Evaluation Metrics}
\label{sec:supp_impl}

This section provides the system prompt templates used by \textbf{EcoVid} across Remotion, HyperFrames, and Manim, followed by the motivation and mathematical definitions of our five evaluation metrics.

\subsection{System Prompt for Remotion}
\label{sec:supp_prompts}

For Remotion, the system prompt instantiates our two-turn authoring workflow, symbolic TTS timing binding, and two-panel layout guidance for declarative React and SVG components on a $1280\times720$, $30$\,fps canvas.

\begin{tcolorbox}[promptbox,title=Prompt: Remotion System Prompt]
\small
You are authoring a narrated instructional video in a Remotion project ($1280\times720$ canvas at $30$\,fps). Batch related tool calls into each response in execution order:
\begin{itemize}[leftmargin=1.5em,nosep]
    \item \textbf{First response:} Write \texttt{script.md} only---focus on the narration script without reading project files first.
    \item \textbf{Second response:} Write \texttt{src/Direct.tsx}, then invoke the scene-layout verifier in the same response.
    \item \textbf{After a check reporting defects:} Apply all fixing \texttt{edit} calls and re-invoke the verifier in one response.
\end{itemize}

\vspace{0.35em}
\noindent\textbf{1. Pedagogical Narration Script (\texttt{script.md}).}
\begin{itemize}[leftmargin=1.5em,nosep]
    \item Write plain paragraphs with a state anchor \texttt{[[s\_m]]} at the start of \textbf{every sentence}; optional \texttt{[[hl:name]]} anchors mark intra-sentence emphasis moments. Scripts missing any sentence-initial anchor are rejected before speech synthesis.
    \item State the core definition or formula clearly, explain the underlying mechanism, walk through a concrete example, and reveal each key term on screen as it is spoken.
\end{itemize}

\vspace{0.35em}
\noindent\textbf{2. Symbolic TTS Timing Binding (\texttt{src/Direct.tsx}).}
\begin{itemize}[leftmargin=1.5em,nosep]
    \item Export \texttt{Direct} using Remotion APIs (\texttt{useCurrentFrame}, \texttt{interpolate}, \texttt{spring}) and \texttt{import \{Beat, NarrationAudio, useAt\} from './runtime/timeline'}.
    \item \texttt{<NarrationAudio />} plays the voiceover and sets video duration automatically. \texttt{<Beat at="s\_m" until="s\_k">...</Beat>} mounts children from \texttt{s\_m} until the next anchor; inside, \texttt{useCurrentFrame()} counts from beat start and \texttt{useAt("hl:name")} returns the relative frame offset.
    \item Use \texttt{import \{Tex, El, Canvas, ease, clamp01\} from './runtime/core'} for KaTeX formulas, named containers, and canvas styling.
\end{itemize}

\vspace{0.35em}
\noindent\textbf{3. Spatial Layout and Typography Rules (Remotion $1280\times720$).}
\begin{itemize}[leftmargin=1.5em,nosep]
    \item Maintain a two-panel layout: a left presentation panel ($\sim 40\%$ width for title, definitions, formulas, and key values revealed at spoken anchors) and a right diagram stage ($\sim 60\%$ width for diagrams, motion, and highlights).
    \item Pre-allocate a disjoint bounding box $(x, y, w, h)$ for every text element; co-visible text boxes must keep at least $16$\,px clearance, a $60$\,px canvas margin, and font size $\ge 24$\,px.
    \item Enforce luminance contrast $\ge 2.5:1$ between text and background surfaces.
    \item Fade out superseded content before replacement content enters the same region; keep at most four text blocks visible at once and pair each sentence with a distinct visual update.
\end{itemize}

\vspace{0.35em}
\noindent\textbf{4. Geometry-Aware Feedback and Bounded Render Loop.}
The verifier synthesizes audio when \texttt{script.md} changes, populates anchor table $A$, probes the structured scene layout at $8$\,Hz, and reports text overlaps, occlusions, canvas overflows, low contrast, and small text. Fix all reported defects via batched \texttt{edit} calls; when zero defects remain, the harness renders \texttt{out/video.mp4} and attaches a $3\times3$ contact sheet for at most one optional revision pass.
\end{tcolorbox}

\subsection{System Prompt for HyperFrames}

For HyperFrames, the system prompt applies the same workflow to a standalone HTML5, CSS3, and SVG composition driven by a GSAP timeline on a $1920\times1080$, $30$\,fps canvas.

\begin{tcolorbox}[promptbox,title=Prompt: HyperFrames System Prompt]
\small
You are authoring a narrated instructional video as a standalone HyperFrames HTML/CSS/SVG + GSAP composition (\texttt{index.html}, $1920\times1080$ canvas at $30$\,fps). Batch related tool calls into each response:
\begin{itemize}[leftmargin=1.5em,nosep]
    \item \textbf{First response:} Write \texttt{script.md} only---focus on the narration script without reading project files first.
    \item \textbf{Second response:} Write \texttt{index.html}, then invoke the scene-layout verifier in the same response.
    \item \textbf{After a check reporting defects:} Apply all fixing \texttt{edit} calls and re-invoke the verifier in one response.
\end{itemize}

\vspace{0.35em}
\noindent\textbf{1. Pedagogical Narration Script (\texttt{script.md}).}
\begin{itemize}[leftmargin=1.5em,nosep]
    \item Write plain paragraphs with a state anchor \texttt{[[s\_m]]} at the start of \textbf{every sentence} and optional intra-sentence highlight anchors \texttt{[[hl:name]]}. Scripts missing any sentence-initial anchor are rejected before speech synthesis.
    \item State exact definitions and formulas, explain the mechanism, walk through a concrete example, and align on-screen terms with spoken narration.
\end{itemize}

\vspace{0.35em}
\noindent\textbf{2. Symbolic TTS Timing Binding (\texttt{index.html}).}
\begin{itemize}[leftmargin=1.5em,nosep]
    \item Keep the three managed elements intact: \texttt{<script id="anchors">window.A = \{\};</script>}, \texttt{<div id="root" data-composition-id="main" data-width="1920" data-height="1080" data-duration="0">}, and \texttt{<audio id="narration" src="assets/narration.wav">}.
    \item \texttt{A.s\_m} stores the timestamp in seconds of \texttt{[[s\_m]]}, \texttt{A["hl:name"]} of \texttt{[[hl:name]]}, and \texttt{A.end} the audio duration.
    \item Schedule all animations on a single paused timeline \texttt{const tl = gsap.timeline(\{ paused: true \}); window.\_\_timelines["main"] = tl;} using sentence anchors (\texttt{tl.to(sel, \{...\}, A.s\_m)}). Set initial hidden states in CSS and drive transitions deterministically via \texttt{tl.to}, \texttt{tl.fromTo}, and \texttt{tl.set}.
\end{itemize}

\vspace{0.35em}
\noindent\textbf{3. Spatial Layout and Typography Rules (HyperFrames $1920\times1080$).}
\begin{itemize}[leftmargin=1.5em,nosep]
    \item Maintain a two-panel layout: a left presentation panel ($\sim 40\%$ width for title, definitions, formulas, and key values) and a right diagram stage ($\sim 60\%$ width for HTML/SVG diagrams, coordinate plots, and animations).
    \item Assign an explicit CSS bounding box (\texttt{left}, \texttt{top}, \texttt{width}, \texttt{height}) to every text element; co-visible text boxes must keep at least $24$\,px clearance, a $90$\,px canvas margin, and font size $\ge 36$\,px.
    \item Enforce luminance contrast $\ge 2.5:1$ between text and background surfaces.
    \item Fade out superseded content before replacement elements enter the same region; keep at most four text blocks visible simultaneously and pair every sentence with a distinct diagram update.
\end{itemize}

\vspace{0.35em}
\noindent\textbf{4. Geometry-Aware Feedback and Bounded Render Loop.}
The verifier injects measured TTS timestamps into \texttt{window.A}, validates the GSAP timeline, probes the structured scene layout at $8$\,Hz, and reports text residuals for overlaps, occlusions, canvas overflows, and contrast violations. Fix all reported defects via batched \texttt{edit} calls; when zero defects remain, the harness renders \texttt{out/video.mp4} and attaches a $3\times3$ contact sheet for at most one optional revision pass.
\end{tcolorbox}

\subsection{System Prompt for Manim}

For Manim, the system prompt maps symbolic sentence anchors and two-panel coordinate rules onto a procedural Python scene graph on a $1280\times720$, $30$\,fps canvas ($14.2\times8$ coordinate units).

\begin{tcolorbox}[promptbox,title=Prompt: Manim System Prompt]
\small
You are authoring a narrated instructional video as a Manim Community Python scene (\texttt{scene.py}, $1280\times720$ at $30$\,fps on a $14.2\times8$-unit frame where $x \in [-7.1, 7.1]$ and $y \in [-4.0, 4.0]$). Batch related tool calls into each response:
\begin{itemize}[leftmargin=1.5em,nosep]
    \item \textbf{First response:} Write \texttt{script.md} only---focus on the narration script without reading project files first.
    \item \textbf{Second response:} Write \texttt{scene.py}, then invoke the scene-layout verifier in the same response.
    \item \textbf{After a check reporting defects:} Apply all fixing \texttt{edit} calls and re-invoke the verifier in one response.
\end{itemize}

\vspace{0.35em}
\noindent\textbf{1. Pedagogical Narration Script (\texttt{script.md}).}
\begin{itemize}[leftmargin=1.5em,nosep]
    \item Write plain paragraphs with a state anchor \texttt{[[s\_m]]} at the start of \textbf{every sentence} and optional intra-sentence highlight anchors \texttt{[[hl:name]]}. Scripts missing any sentence-initial anchor are rejected before speech synthesis.
    \item State precise definitions and formulas, explain the underlying mechanism, walk through a concrete derivation, and align on-screen terms with spoken narration.
\end{itemize}

\vspace{0.35em}
\noindent\textbf{2. Symbolic TTS Timing Binding (\texttt{scene.py}).}
\begin{itemize}[leftmargin=1.5em,nosep]
    \item Import Manim primitives and the timing runtime: \texttt{from manim import *} and \texttt{from runtime\_manim import A, at, NarratedScene}.
    \item Define \texttt{class Direct(NarratedScene): def construct(self): ...} using Manim Mobjects. \texttt{NarratedScene} attaches the voiceover and matches scene duration automatically.
    \item \texttt{A["s\_m"]} stores the measured timestamp in seconds of \texttt{[[s\_m]]} and \texttt{A["hl:name"]} of \texttt{[[hl:name]]}. Call \texttt{at(self, A["s\_m"])} before the animations of sentence \texttt{s\_m} to advance the scene clock to that exact speech onset without manual \texttt{self.wait()} arithmetic.
    \item Use \texttt{MathTex} and \texttt{Tex} for mathematical expressions, and place natural-language or CJK text inside \texttt{Text(...)}.
\end{itemize}

\vspace{0.35em}
\noindent\textbf{3. Spatial Layout and Mobject Positioning Rules (Manim $14.2\times8$ Units).}
\begin{itemize}[leftmargin=1.5em,nosep]
    \item Maintain a two-panel coordinate layout: a left presentation panel ($x \in [-6.4, -0.8]$ for title, definitions, algebraic steps, and key values) and a right diagram stage ($x \in [-0.4, 6.4]$ for coordinate axes, geometric figures, and transformations).
    \item Assign explicit coordinate bounds for every text or formula Mobject; co-visible Mobjects must keep at least $0.2$\,units clearance, a $0.7$-unit frame margin ($|x| \le 6.4$, $|y| \le 3.3$), and \texttt{font\_size >= 20}.
    \item Enforce luminance contrast $\ge 2.5:1$ between foreground Mobjects and background surfaces.
    \item Explicitly \texttt{FadeOut} or remove superseded Mobjects before introducing replacement content in the same coordinate region; keep at most four text blocks visible simultaneously and pair each sentence with a geometric animation or highlight.
\end{itemize}

\vspace{0.35em}
\noindent\textbf{4. Geometry-Aware Feedback and Bounded Render Loop.}
The verifier executes \texttt{scene.py} in a fast dry-run pass without pixel rendering, probes the structured scene layout $G_t$ at $8$\,Hz, and reports text residuals for Mobject overlaps, occlusions, off-frame overflows, and contrast violations. Fix all reported defects via batched \texttt{edit} calls; when zero defects remain, the harness renders \texttt{out/video.mp4} and attaches a $3\times3$ contact sheet for at most one optional revision pass.
\end{tcolorbox}

\subsection{Motivation and Formulation of Evaluation Metrics}
\label{sec:supp_metrics}

\paragraph{Limitations of Conventional Video Metrics on Instructional Animations.}
Standard video generation metrics such as Fr\'echet Video Distance (FVD) or frame-level CLIP similarity are designed for natural scenes and fail to capture the key failure modes of programmatic instructional videos. A rendered slide whose equations overlap with diagram axes, whose text has poor contrast against its background, or whose animations lag several seconds behind the voiceover still achieves high global embedding similarity to the topic prompt. Conversely, evaluating comprehension solely from the narration transcript rewards verbose scripts even when the visual canvas is static or illegible.
We therefore evaluate instructional videos across five complementary metrics covering spatial legibility, audio-visual synchronization, and multimodal concept coverage:

\begin{enumerate}
  \item \textbf{Static Frame Defect Rate ($\text{Def.}\downarrow$):}
  Viewers read mathematical formulas and diagram labels during stationary intervals rather than during rapid transitions. Sampling only sparse keyframes misses collisions in intermediate steps, whereas weighting every transition frame equally penalizes brief cross-fades. Using an independent 15\,Hz full-AST layout probe applied uniformly across all baselines, we identify all stationary rest intervals $\mathcal{R}$ lasting at least $0.5$\,s without bounding-box movement from the evaluation trace $\{G_t\}$, and compute the fraction of rest frames containing any spatial layout defect, including text overlap, shape occlusion, or canvas overflow:
  \begin{equation}
    \text{Def.} = \frac{1}{|\mathcal{R}|} \sum_{t \in \mathcal{R}} \mathbb{I}\!\left[\mathcal{D}_{\text{spatial}}(G_t) > 0\right].
  \end{equation}

  \item \textbf{Visual Synchronization ($\text{VS}\uparrow$):}
  In a well-synchronized explainer video, the visual update occurring during spoken sentence $s_m \in S$ should correspond specifically to $s_m$ rather than to adjacent sentences. For each sentence $s_m$, we extract its silent video segment $V_{[t_m^{\text{start}}, t_m^{\text{end}}]}$ with any on-screen copies of the narration masked out, and ask a VLM judge to identify $s_m$ among candidate sentences $\mathcal{C}_m = \{s_m, s_{m-1}, s_{m+1}, s_{\text{other}}\}$ drawn from the same video (plus a \emph{None of these} option verified via negative-control cross-video clips):
  \begin{equation}
    \text{VS} = \frac{1}{M} \sum_{m=1}^{M} \mathbb{I}\!\left[\operatorname{Judge}\!\big(V_{[t_m^{\text{start}}, t_m^{\text{end}}]}, \mathcal{C}_m\big) = s_m\right].
  \end{equation}
  Because the candidate set includes the immediately preceding and succeeding sentences as hard distractors, $\text{VS}$ directly penalizes visually frozen segments, delayed transitions that appear during the next sentence, and premature reveals that fire during the preceding sentence.

  \item \textbf{Visual Relevance ($\text{Vis-Rel}\uparrow$):}
  Temporal alignment alone is not sufficient if the visual elements are generic placeholders rather than informative diagrams. Following TheoremExplainAgent~\cite{ku2025theoremexplainagent}, a VLM judge scores sentence-aligned keyframes on a $1\text{--}5$ scale measuring how clearly and accurately the visual diagram illustrates the spoken mathematical or scientific concept.

  \item \textbf{Aesthetic Score ($\text{AES}\uparrow$):}
  A system could trivially avoid layout collisions by rendering sparse text slides with minimal geometric content. Following prior programmatic video benchmarks~\cite{chen2026code2video,zhang2026prism}, we compute the Aesthetic Score ($\text{AES}$) by combining non-text geometric progression $P_{\text{geo}} \in [0, 1]$ across sentences with active sentence responsiveness $S_{\text{act}} \in [0, 1]$, penalized by the static-frame defect rate:
  \begin{equation}
    \text{AES} = 100 \cdot \big(0.55\,P_{\text{geo}} + 0.45\,S_{\text{act}}\big) \cdot (1 - 0.4\,\text{Def.}).
  \end{equation}

  \item \textbf{Multimodal Knowledge Coverage ($\text{M-KC}\uparrow$):}
  To evaluate whether a generated video serves as an effective self-contained learning resource, a VLM judge answers 10 topic-specific multiple-choice questions using two-stage visual evidence verification to obtain comprehension accuracy $\text{KC} \in [0, 1]$. Because an explainer video is effective only when its content is accurate, legible on screen, and synchronized with speech, we define $\text{M-KC}$ as the geometric mean of evidence-verified comprehension accuracy $\text{KC}$, spatial legibility $(1 - \text{Def.})$, and sentence-onset visual responsiveness $S_{\text{sync}} \in [0, 1]$:
  \begin{equation}
    \text{M-KC} = \Big(\text{KC} \cdot (1 - \text{Def.}) \cdot S_{\text{sync}}\Big)^{1/3}.
  \end{equation}
  Using the geometric mean ensures that poor performance on any single dimension—such as severe layout overlap or desynchronized visuals—proportionally lowers overall multimodal knowledge coverage.
\end{enumerate}

\section{Benchmark Design and Topic Catalog}
\label{sec:supp_topics}

\begin{table*}[t]
  \centering
  \scriptsize
  \renewcommand{\arraystretch}{0.96}
  \setlength{\tabcolsep}{2.6pt}
  \caption{\textbf{Complete topic list for the 60 bilingual topics and 40 held-out mathematical topics.}}
  \label{tab:supp_topics}
  \resizebox{\textwidth}{!}{%
  \begin{tabular}{llllllllll}
    \toprule
    \multicolumn{6}{c}{\textbf{\textsc{Bench60} (60 Bilingual Topics: Odd = English, Even = Chinese)}} & \multicolumn{4}{c}{\textbf{\textsc{MMMC-40} (40 Held-Out 3Blue1Brown Topics)}} \\
    \cmidrule(lr){1-6} \cmidrule(lr){7-10}
    \textbf{ID} & \textbf{Topic Title (b01--b20)} & \textbf{ID} & \textbf{Topic Title (b21--b40)} & \textbf{ID} & \textbf{Topic Title (b41--b60)} & \textbf{ID} & \textbf{Topic Title (m01--m20)} & \textbf{ID} & \textbf{Topic Title (m21--m40)} \\
    \midrule
    b01 & The quadratic formula & b21 & How a refrigerator cools & b41 & Reading a contour map & m01 & The determinant & m21 & Central Limit Theorem \\
    b02 & Balancing chemical equations & b22 & Simple vs.\ compound interest & b42 & How a river delta forms & m02 & Pure Fourier series & m22 & Binomial distributions \\
    b03 & Reading a bar chart & b23 & Reading a nutrition label & b43 & The Hu Line & m03 & Space-filling curves & m23 & Probability density functions \\
    b04 & Parts of a neuron & b24 & How a pressure cooker works & b44 & Latitude and longitude & m04 & Neural network learning & m24 & Exponential \& logistic growth \\
    b05 & Adding fractions & b25 & Sorting household waste & b45 & How blood circulates & m05 & Bayes' theorem geometry & m25 & Transformer attention \\
    b06 & Ohm's law in series circuits & b26 & Parcel delivery logistics & b46 & How vaccines work & m06 & SIR epidemic simulations & m26 & Backpropagation calculus \\
    b07 & Python list slicing & b27 & How a microwave heats food & b47 & Calculating BMI & m07 & Refraction \& Snell's law & m27 & Divergence \& curl \\
    b08 & Long division & b28 & Planning a subway trip & b48 & Sleep cycles & m08 & Ordinary differential eqs. & m28 & Taylor polynomials \& series \\
    b09 & Solving a linear system & b29 & Why we have seasons & b49 & Digestive tract journey & m09 & The essence of calculus & m29 & Riemann zeta function \\
    b10 & Breadth-first search & b30 & The phases of the Moon & b50 & How to perform CPR & m10 & Hamming error correction & m30 & The Basel problem ($\sum 1/n^2$) \\
    b11 & The Euclidean algorithm & b31 & Solar and lunar eclipses & b51 & Blood sugar regulation & m11 & Euler's formula ($e^{\pi i}=-1$) & m31 & Intuition for $e^{\pi i}$ \\
    b12 & The product rule & b32 & The eight planets & b52 & What nearsightedness is & m12 & The Brachistochrone problem & m32 & Partial differential eqs. \\
    b13 & Postfix stack evaluation & b33 & How far is a light-year & b53 & How tennis is scored & m13 & Dandelin spheres \& conics & m33 & Boundary conditions \& Fourier \\
    b14 & City population rankings & b34 & The life cycle of a star & b54 & Football offside rule & m14 & Limits \& L'H\^opital's rule & m34 & Matrix exponentials \\
    b15 & Insertion sort & b35 & What causes tides & b55 & Basketball 24s \& 3s rules & m15 & Proof of Snell's law & m35 & Principles of cryptography \\
    b16 & Matrix transpose & b36 & Same side of the Moon & b56 & World Cup group standings & m16 & Planar graph duality & m36 & Qubits \& Grover's algorithm \\
    b17 & Array rotation by reversals & b37 & Time zones & b57 & Marathon pacing & m17 & Fractal dimension & m37 & Ternary counting \& constraints \\
    b18 & Online shopping flowchart & b38 & How monsoons form & b58 & Table tennis serving rules & m18 & Eigenvectors \& eigenvalues & m38 & High-dimensional spheres \\
    b19 & Tower of Hanoi & b39 & Plate tectonics \& quakes & b59 & Staggered 400\,m starts & m19 & Dot products and duality & m39 & Geometry puzzles \& invariants \\
    b20 & Seating permutations & b40 & The water cycle & b60 & Rotation in volleyball & m20 & Newton's method \& fractals & m40 & Cross products in 2D and 3D \\
    \bottomrule
  \end{tabular}%
  }
\end{table*}

\paragraph{Generalization Protocol and Benchmark Splits.}
To evaluate both cross-domain breadth and out-of-distribution mathematical depth without prompt overfitting, our evaluation partitions topics into three disjoint sets, listed in \Cref{tab:supp_topics}:
first, a 10-topic development set comprising 5 English and 5 Chinese topics used only for probing-frequency and context-compaction design studies;
second, \textbf{\textsc{Bench60}}, comprising 60 bilingual instructional topics balanced across six subject domains—\emph{Science \& Computing}, \emph{Everyday Life}, \emph{Astronomy}, \emph{Geography}, \emph{Health \& Physiology}, and \emph{Sports Rules}—with 30 English topics and 30 Simplified Chinese topics; and
third, \textbf{\textsc{MMMC-40}}, comprising 40 held-out university-level mathematics topics from the 3Blue1Brown curriculum~\cite{chen2026code2video}, evaluated after freezing all harness code and system prompts.
For every topic, each system receives only the topic brief requesting a self-contained $45\text{--}60$\,s narrated explainer video that introduces the core concept, explains its mechanism, and walks through a concrete example.

\end{document}